\documentclass{article}
\usepackage{arxiv}
\usepackage{cite}
\usepackage{xcolor}
\usepackage{url,multirow,cancel,textcomp,tfrupee}
\usepackage{pifont}
\usepackage[nointegrals]{wasysym}
\usepackage{amsmath,amssymb,amsfonts}
\usepackage{algorithmic}
\usepackage{graphicx}
\usepackage{textcomp}
\usepackage{caption}
\usepackage{subcaption}
\usepackage{setspace}
\usepackage{fancyhdr}

\title{Preimplantation Blastomere Boundary Identification in HMC Microscopic Images of Early Stage Human Embryos}

\author{
  Shakiba~Kheradmand \\
  Laboratory for Robotic Vision\\
  School of Engineering Science\\
Simon Fraser University\\
 Burnaby, BC, Canada\\
  \texttt{shakiba.kheradmand@gmail.com} \\
   \And
 Parvaneh~Saeedi \\
 Laboratory for Robotic Vision\\
 School of Engineering Science\\
Simon Fraser University\\
 Burnaby, BC, Canada\\
  \texttt{psaeedi@sfu.ca} \\
   \And
 Jason Au, John Havelock \\
  Pacific Centre for Reproductive Medicine\\
  Burnaby, BC, Canada\\
  \texttt{{jau, jhavelock}@pacificfertility.ca} \\
}
    
\begin{document}
\maketitle
\begin{abstract}
We present a novel method for identification of the boundary of embryonic cells (blastomeres) in Hoffman Modulation Contrast (HMC) microscopic images that are taken between day one to day three. Identification of boundaries of blastomeres is a challenging task, especially in the cases containing four or more cells. This is because these cells are bundled up tightly inside an embryo's membrane and any 2D image projection of such 3D embryo includes cell overlaps, occlusions, and projection ambiguities. Moreover, human embryos include fragmentation, which does not conform to any specific patterns or shape. Here we developed a model-based iterative approach, in which blastomeres are modeled as ellipses that conform to the local image features, such as edges and normals. In an iterative process, each image feature contributes only to one candidate and is removed upon being associated to a model candidate. We have tested the proposed algorithm on an image dataset comprising of 468 human embryos obtained from different sources. An overall \textit{Precision}, \textit{Sensitivity} and \textit{Overall Quality} (\textit{OQ}) of 92\%, 88\% and 83\% are achieved.
\end{abstract}


\maketitle

\section{Introduction}
\label{sec:introduction}
It is estimated that one out of every six couples in Canada suffers from infertility\footnote{https://www.canada.ca/en/public-health/services/fertility/fertility.html [accessed 2019-06-08]}. Assisted Reproductive Technology (ART) involves medical procedures to overcome infertility using procedures that handle eggs and embryos outside the women's body. Among the most common fertility treatments, In Vitro Fertilization (IVF) has shown tremendous success \cite{gardner2017handbook}. Dating back to the 1970's, in IVF, multiple hyper-stimulated ovarian follicles are retrieved from the female. The oocytes are extracted from the follicles and then fertilized and cultured for as long as five days, during which they are regularly monitored and assessed morphologically. The assessment is aimed to identify embryos with high potentials for implantation. Usually, multiple embryos will be selected and transferred to the female's uterus. The successful birth of a baby using just one embryo, which is what is the core of ART, requires accurate selection procedure~\cite{ElderKay}. The selection process is manually undertaken by experienced embryologists who inspect the development pattern of the embryos and their growth rate. The selection criteria, the precision of the selection, and the level of expertise are crucial factors in the outcome of IVF procedures. Automating part of the selection process will reduce the time spent by experts, hence reducing the high cost and rendering IVF more widely accessible.

When a fertilized egg divides into two cells, it enters the cleavage stage of its development. The cells of a normal two-cell embryo will later divide and create a four-cell embryo. Each cell in the four-cell embryo will divide again to form an eight-cell embryo. Embryologists have investigated some attributes at this stage that may indicate which embryos will advance further. These attributes include similarity in size, little or no fragmentation, and division time. Healthy embryos have a fairly strict rate of progression starting at the time of their fertilization. 
Several morphological criteria are considered in embryo evaluation, with cell size and symmetry as one of the most important factors~\cite{gardner2016assessment}. The existence of cells with different sizes reduces the chance of a successful pregnancy. 

It is highly crucial to curate effective morphological metrics in order to evaluate embryos in terms of their implant-ability potentials. Having the number of blastomeres, we present a novel algorithm to measure various parameters related to the normal growth of embryos including their size and shape symmetry. The algorithm generates ellipsoidal models using local image features. More specifically, all potential cell candidates are estimated initially. The qualities of these candidates are assessed using their geometric and algebraic primitives. Our objective is to automatically and robustly identify blastomere boundaries. Such identification 
can be used to measure the size of blastomeres in time-lapse IVF images. 
%
 Contributions made through this work include:
\begin{itemize}
\item Identification of boundaries of up to eight blastomeres in microscopic embryo images of single focus plane including occlusion and fragmentation,
\item Creation of a dataset of over 450 embryo images of actual patients with the ground truth generated by the expert embryologists that is available to the researchers in this field upon request, and 
\item Introduction of a system that can perform boundary identification automatically (with known cell number) on images from various sources. 
\end{itemize}
\vspace{-3 mm}
\section{Related Work} 
In IVF, embryologists monitor embryos during the development in successive growth stages from different morphological aspects including size, shape, fragmentation and development rate. While some semi-automatic tools are available for computer-assisted boundary detection, IVF analysis including identification and association largely requires direct human intervention. Fully-automatic systems are yet to be developed \cite{singh2014automatic}.

3D modeling of blastomeres has been attempted in multiple studies. For example, Pedersen~\textit{et al.}~\cite{IT-C:564} applied the variational level set approach to develop 3D models of blastomere systems. Their approach mainly relied on ``representing the geometrical conditions with level sets''. Analysis of 3D and 3D-like images, such as those produced with a side-lit, has also attracted attention. Giusti~\textit{et al.}~\cite{Giusti2010} introduced a graph-based algorithm to segment 3D images obtained from HMC microscopy. The algorithm is shown to be robust with respect to noise, clutter, and other artifacts. This work followed an earlier study by \cite {giusti2009lighting} who used a graph-based algorithm to segment a single embryo cell. More recently, Khan~\textit{et al.}~\cite{khan2016segmentation} developed a graph-based algorithm for segmenting the overall embryo boundaries. 

Probabilistic approaches are attempted in the literature. For instance, Wong~\textit{et al.}~\cite{Wong2010} leveraged sequential Monte Carlo simulations to automatically track cell divisions. In their approach, embryos were modeled as a collection of ellipses with position, orientation and overlap indices. Limited visual results pertaining to 14 embryo sequences were presented. However, no quantitative results were reported. Later, Khan~\textit{et al.}~\cite{khan2015linear} provided methods based on a linear chain Markov model to estimate the number of cells and their locations. Their method finds the most likely sequence of hypotheses over all time in a sequence of time-lapse microscopy images up to four cells and relates this to the actual cell numbers and locations. They reported an overall \textit{IoU} (Intersection over Union) value of $84\%$.

In recent years, methods based on Conditional Random Field (CRF) are proposed. Using time-lapse microscopic images, Moussavi~\textit{et al.} designed a CRF-based mitosis detection algorithm \cite{moussavi2014unified}. Khan~\textit{et al.}~\cite{khan2015automated} combined a CRF approach with supervised machine learning to predict the number of embryonic cells. Using this method, the author reported predicting up to five cells in the time-lapse images with 94\% accuracy, outperforming their own previous work\cite{khan2015linear}. The studies reviewed above mainly concern extracting the number of cells, and do not segment the boundaries of the blastomeres. It is also worth mentioning that the promising performance of deep learning algorithms in extracting the number of embryonic cells has not unfortunately been repeated in blastomere segmentation. This is primarily because the deep learning algorithms are data-hungry, and do not perform well when data is scarce. With the current datasets of a few hundred images, other image analysis techniques are deemed more practical.

As for segmentation, Grushnikov\textit{et al}~\cite{grushnikov20183d} introduced a level-set algorithm, designed to work with two different sets of microscopy images, each focused at different depth. One set is focused on the cell nuclei, and the other set is focused on the membranes. They used a dataset comprising of 20 mouse embryos and detected the inner regions of blastomeres and the volume of the membrane. While this study shed some light on how level-set algorithms can be used on multiple sets of images, the performance of level set methods and region growing techniques are quickly deteriorated by extreme occlusion and cell overlaps in blastomere images, along with extreme human embryo fragments. Other segmentation techniques used for nucleus/cell images (such as ~\cite{irshad2014methods, xing2016robust}) fail in blastomere boundary detection for the same reasons.

Model-based detection and segmentation systems have become popular in recent years. Singh~\textit{et al.}~\cite{singh2014automatic} described a model-based system for the detection of up to four blastomeres with a detection accuracy of $80\%$ using 40 images. More recently, Moradi rad et.al~\cite{rad2018hybrid} proposed an automated hybrid approach to segment embryos with up to eight cells with different shapes and sizes. Their model identifies and captures local and global features, which then allows it to obtain the location of each blastomere. They ran their algorithm on 271 embryo images and have reported an average precision of 85.9\%, while the recall is 85.3\%.

In this paper, we present a novel method to detect up to 8 blastomere boundaries in human embryos. More specifically, our algorithm informs of the size and shapes of blastomeres. Information about the size and the shapes of the blastomere is extremely useful for the embryologist to determine the quality of embryos as they develop~\cite{gardner2016assessment}. In the following section, we explain the segmentation algorithm and showcase its performance against 468 embryo images. 
\vspace{-3 mm}

\section{Proposed method}
Here, we propose an ellipsoidal model-based approach for detecting blastomere cells in embryos of day 1 to day 3. Before detailing different components of our algorithm, it is necessary to justify why an ellipsoidal model is chosen for modeling blastomeres. Different parts of our algorithms in the order that are applied to the input images are explained later. 

\subsection{Ellipsoidal Model-Based Approach Justification}\label{fid_res}
Visually, the boundary profile of blastomeres in the image of an embryo follows an ellipse for most of the part and most of the time. 
To quantify how well an ellipse can represent the blastomere shape, we compare the Ground Truth (GT) (more details in Section~\ref{GT_sec}) with the best-fitted ellipse representing the GT contour. 
We then computed a validation metric for spatial overlap index called the Dice Similarity Coefficient or $DSC$~\cite{zou04} to measure the error between the area of each blastomere in GT and the best-fitted ellipse. For two regions of $A$ and $B$, $DSC$ is computed by $DSC(A,B)= 2(A\cap B)/(A\cup B)$ at pixel levels. Table~\ref{tab:fid} depicts the average $DSC$ measures according to the number of blastomeres in each embryo image for all of our input images. The average error in the worst and best-case scenarios vary from $3\%$ to $1\%$. These values indicate that the ellipsoidal model assumption is indeed a reasonable one for presenting the general shape profile of the blastomeres and lead only to a maximum of $3\%$ overall error. 

\begin{table}[h]
\caption{\small{Mean DSC values for 1-8 cell embryos.}}
\tabcolsep=0.11cm
\centering
 \begin{tabular}{ |c|c|c|c|c|c|c|c|c| }
    \hline
     &\small{1-cell} & \small{2-cell} & \small{3-cell} & \small{4-cell} &\small{5-cell} & \small{6-cell} & \small{7-cell} & \small{8-cell} \\ \hline\hline
 \small{DSC} & \small{0.989} &\small{0.979} &\small{0.971} &\small{0.979} &\small{0.978} &\small{0.978} &\small{0.977} &\small{0.977} \\ 
    \hline
 \end{tabular}
 \label{tab:fid}
 \end{table}

Several previous works exist for fitting ellipsoidal models in images of different types in the literature~\cite{prasad2012edge,hahn2008new,chia2011split}. While these methods seem to work well for their specific applications, they all failed to properly detect blastomere profiles in our HMC microscopic images. Several reasons could be the source of such performance inconsistencies including: large amount of discontinuity due to full/partial occlusion caused by the neighbouring blastomeres, cell fragmentation phenomenon, embryo's motion, and blastomere's surface texture that leads to a large number of short and fine edge features that do not correspond to the boundary of the blastomeres. 

For all the above reasons, here we had to develop a new approach for detecting ellipsoidal shape blastomeres.
\subsection{Edge Detection}\label{sec:edgeDetection}
Human embryo images, by nature, include fine details that may cause traditional edge 
detectors to detect some unnecessary or out of interest details. 
On the other hand, sometimes two or more blastomeres fully or partially 
occlude each other or another blastomere (usually hidden under) and in such cases, 
the boundaries of the hidden blastomere are faint or invisible. Clearly, increasing 
the sensitivity of any traditional edge detector can cause an imbalance 
between the two above features. To address this problem
as well as to identify the boundaries of the cells more accurately, we use Frangi Vesselness filter~\cite{Frangi98multiscalevessel}  to utilize 
multi-scale second-order local structures of the image. Frangi filter measures ``vessel-likeliness'' through calculating and analyzing the eigenvalues of the Hessian matrix, which is shown to have great performance on embryo images with vessel-like boundaries of blastomeres. 

The filter can be described by the second-order Taylor expansion of image $I(x,y)$, in a local neighbourhood of point ($x_0$, $y_0$):
\begin{equation}
I(x_0,y_0) \approx  I(x_0+\delta x_0, y_0+\delta y_0) + \delta X^T\nabla I|_{(x_0,y_0)} \\\nonumber
+\frac{1}{2}(\delta X^T H|_{(x_0, y_0)}  \delta X)
\end{equation}

here $\nabla$ is the gradient vector, $H$ is the Hessian matrix computed at 
($x_0$, $y_0$), and and $\delta X$ is ($\delta x_0$,$\delta y_0$). The elements of the Hessian matrix  contain second derivatives, and therefore inform of the shape characteristics, both qualitatively and quantitatively. 
The eigenvalues of the Hessian matrix are denoted by$\lambda_1$ and $\lambda_2$, where $| \lambda_1| \le | \lambda_2|$. Following \cite{Frangi98multiscalevessel}, we define: "blobness measure" ($R_B$),  "second-order structureness" ($S$) and "vessel likeliness" ($V_0(\sigma)$):
\begin{subequations}\begin{align}\nonumber
&R_B=\lambda_1/\lambda_2, \\ \nonumber
&S=\parallel H \parallel_F=\sqrt{{\lambda_1}^2+{\lambda_2}^2},
\end{align} \end{subequations}
\begin{equation} \nonumber
V_0(\sigma)= 
\begin{cases}
  \qquad   \qquad 0 & \text{if } \lambda_2 > 0,\\
\text{exp}(\frac{-{R_B}^2}{2{\alpha}^2})(1-\text{exp}(-\frac{S^2}{2\beta^2})) & \text{otherwise}  \end{cases}, 
\end{equation}
where $\alpha$ and $\beta$ are hyper parameters that tunes the sensitivity of the filter to  $R_B$ and $S$.
As it is indicated in \cite{Frangi98multiscalevessel}, the vessel likeliness at any point is given by 
\begin{equation}
V_0=\underset{\sigma_{\text{min}} \le \sigma \le \sigma_{\text{max}}}{\text{max}} V_0(\sigma)
\end{equation}
Notice that the maximizing operation takes place in $ [\sigma_{\text{min}}, \sigma_{\text{max}}]$ which corresponds to the size of ridge \cite{Frangi98multiscalevessel}. Upon computing $V_0$ over the entire image, non-maxima suppression and 
hysteresis thresholding are performed in the same manner as Canny edge detector~\cite{Canny:1986}
to generate an image edge map.  
This is followed by a cleaning process in which small segments are removed from the edge image.
\vspace{-1mm}
\subsection{Edge Processing}
In this section, we take advantage of the elliptical profile of embryos and blastomeres to perform a pre-processing step in which redundant edges are removed and edges of low-contrast condition are enhanced. To establish some coherence between edges corresponding to a physical object,  we form edge clusters. 
\subsubsection{Piecewise-linear Approximation of Edges to Edge Clusters}\label{plae}
Following~\cite{Dunham86}, we use a piecewise linear line segment fitting algorithm in order to approximate curves. This happens by minimizing the number of line segments within a 
uniform error $\epsilon$ (=2 pixels in this work) with fixed initial and final points. This algorithm incorporates the edge map created in Section~\ref{sec:edgeDetection}. 

\begin{figure}[!htb]
\begin{center}
   \includegraphics[width=0.6\linewidth]{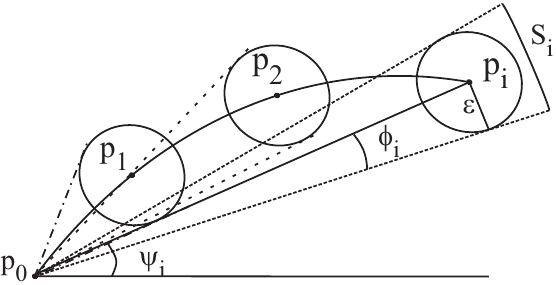}
\end{center}
  \caption{\small{Piecewise linear line segment approximation process.}}
\label{piecewise}
\end{figure}
\begin{figure}
\centering
   \includegraphics[width=3.00in]{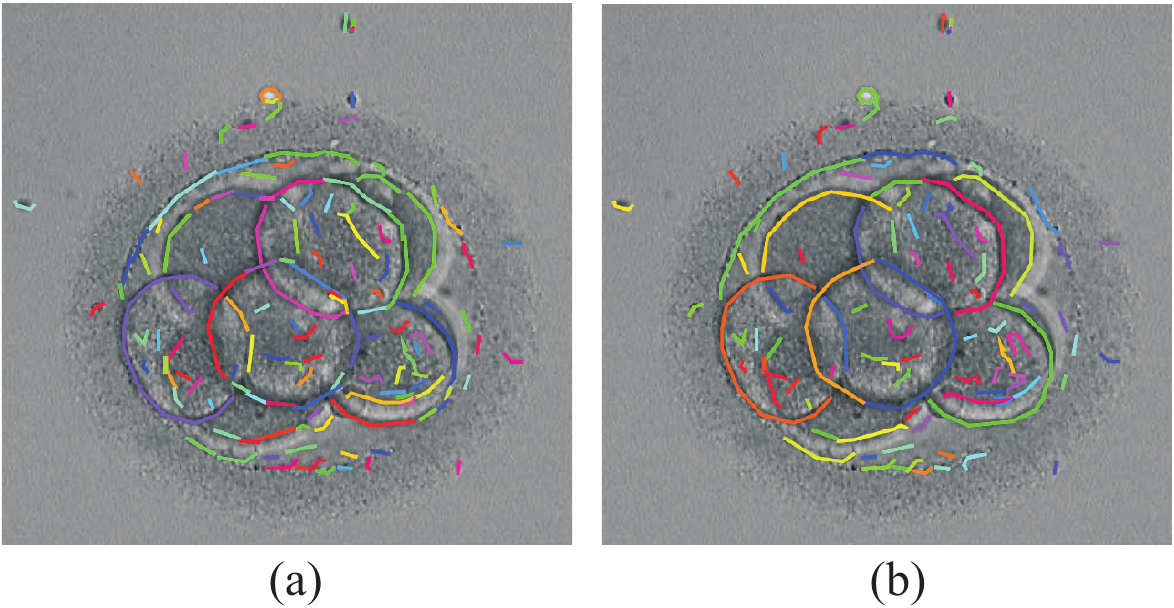}
  \caption{\small{(a) Edge clusters. (b) Edge clusters after co-association.}}
\label{lineEdges}
\end{figure}

If $p_0$ is the origin of a curve segment in Fig.~\ref{piecewise}, for point $p_1$ a circle of radius 
$\epsilon$ represents all lines through $p_0$ which pass within distance $\epsilon$ of $p_1$ (set $S_1$). 
Similarly at point $p_2$  a circle of radius $\epsilon$ represents all lines through $p_0$ which pass 
within distance $\epsilon$ of $p_2$ (set $S_2$). Then the set $T_2 = S_1 \cap S_2$ represents all lines through $p_0$  which pass within $\epsilon$ of both 
$p_1$  and $p_2$. In general we can say:
\begin{equation}
\psi_i =\arctan(\frac{y_0-y_i}{x_0-x_i})
\end{equation}
\begin{equation}
\phi_i=\arcsin(\frac{\epsilon}{(x_0-x_i)^2+(y_0-y_i)^2})
\end{equation}
\begin{equation}
S_i=\{\theta | \psi_i - \phi_i \le   \theta \le  \psi_i + \phi_i \}, \quad T_i=\bigcap_{j=1}^i  S_j
\end{equation}
%
$T_i$ can be recursively computed as follows. 
\begin{equation}
T_i=[\theta_{\min}(i), \theta_{\max}(i)]
\end{equation}
\begin{equation}
\theta_{\min}(i+1)=\min\Big((\max(\theta_{\min}(i), \\\nonumber
\psi_{i+1} - \phi_{i+1})), \theta_{\max}(i)+\delta\Big)
\end{equation}
\begin{equation}
\theta_{\max}(i+1)=\max\Big((\min(\theta_{\max}(i), \\\nonumber \psi_{i+1} - \phi_{i+1})), \theta_{\min}(i)+\delta\Big)
\end{equation}
If $T_i$ is empty, there can be no line passing within $\epsilon$ of
points $p_0$, $p_1$, $\cdots$, $p_i$ and hence no more points in $V(p_0)$. 
Here  $V(p_0)$ denotes all the points of $p_i$ 
where $e(p_i, p_0) \le \epsilon$. $e$ represents the error over a segment between any two points of $p_{i}$ and $p_{j}$:
\begin{equation}
e(p_{i}, p_{j}) = {\underset{p_i \le k \le p_{j}}{\max}}e(p_{i}, k)
\end{equation}
If $T_i$ is not empty, then if $\phi_i \in T_i$, all the points $p_0$, $p_1$, 
$\cdots$, $p_i$ are potentially within $\epsilon$  of the line $\overline{p_0p_i}$. 
If however $\phi_i \notin T_i$, one or more points in $p_0$, $p_1$, $\cdots$, $p_i$ are greater than
$\epsilon$ away from the line $\overline{p_0 p_i}$  and therefore $p_i \notin V(p_0)$. Fig.~\ref{lineEdges}(a) shows edge 
line segments after piecewise linear approximation. 
In this work, we refer to these segments by edge clusters.
\subsubsection{Edge Clusters Properties}
Here, we utilize several properties of each edge cluster to further characterize it more specifically. 
It is important to note that microscopic embryo images may include noise and cell fragments and they may have been taken under non-ideal illumination conditions. These conditions could easily hinder the correct identification of blastomeres. Moreover, the profile of blastomeres may not necessarily follow the curvature of an ellipse. 
The edge cluster properties are: 
\begin{itemize}
\item \textbf{Arch Centroid:}
Assuming that the curve of boundaries of each embryo is circular, we find its centroid. For this, the perpendicular bisectors to all line segments in a curve are calculated. Given the two endpoints of a line $(x_1, y_1)$ and $(x_2, y_2)$, the perpendicular bisector line will be:
\begin{equation}
	\left(y_1-y_2\right)y + \left(x_1-x_2\right)x = \frac{1}{2}\left(x_1^2-x_2^2+y_1^2-y_2^2\right)
\end{equation}
For an ideal circular profile, all perpendicular bisectors meet at one point (the centroid). However, given the non-ideal condition, the point with minimum distances from all the perpendicular bisectors is found. Assuming $N$ bisectors with equations of $y=m_i x+b_i$, where $i=1,2,3,...,N$, the goal is to find point $P(P_x,P_y)$ such that sum of distances of $P$ from bisector lines is minimum. The distance of point $P$ from the $i^{th}$ line is given by:
\begin{equation}
D_i^2 = \frac{\left(m_iP_x-P_y+b_i\right)^2}{m_i^2+1}
\end{equation} 
therefore, point $P$ is found such that $f$ is minimized:
\begin{equation} 
f(P_x,P_y) = \sum_{i=1}^N D_i^2
\end{equation}
\item \textbf{Radius:} 
We define radius as the average distances between the two endpoints and the center. 

\item \textbf{Arch Length:}
Arch length is simply the sum of the all line segment lengths in each edge cluster. 

\item \textbf{Arch Angle:}
The angle of a curve is approximated by the angle between the lines that connects the two endpoints of the curve's edge cluster to its center. More specifically: 
\end{itemize}
\begin{equation}
    \mu = \arctan{\frac{(x_c-x_1)(y_c-y_2)-(y_c-y_1)(x_c-x_2)}{(x_c-x_1)(x_c-x_2)+(y_c-y_1)(y_c-y_2)}}
\end{equation}

\subsubsection{Inner ZP Boundary Extraction}\label{sec:Inner}
\begin{figure}
	  \centering
	  	 \includegraphics[totalheight=0.14\textheight]{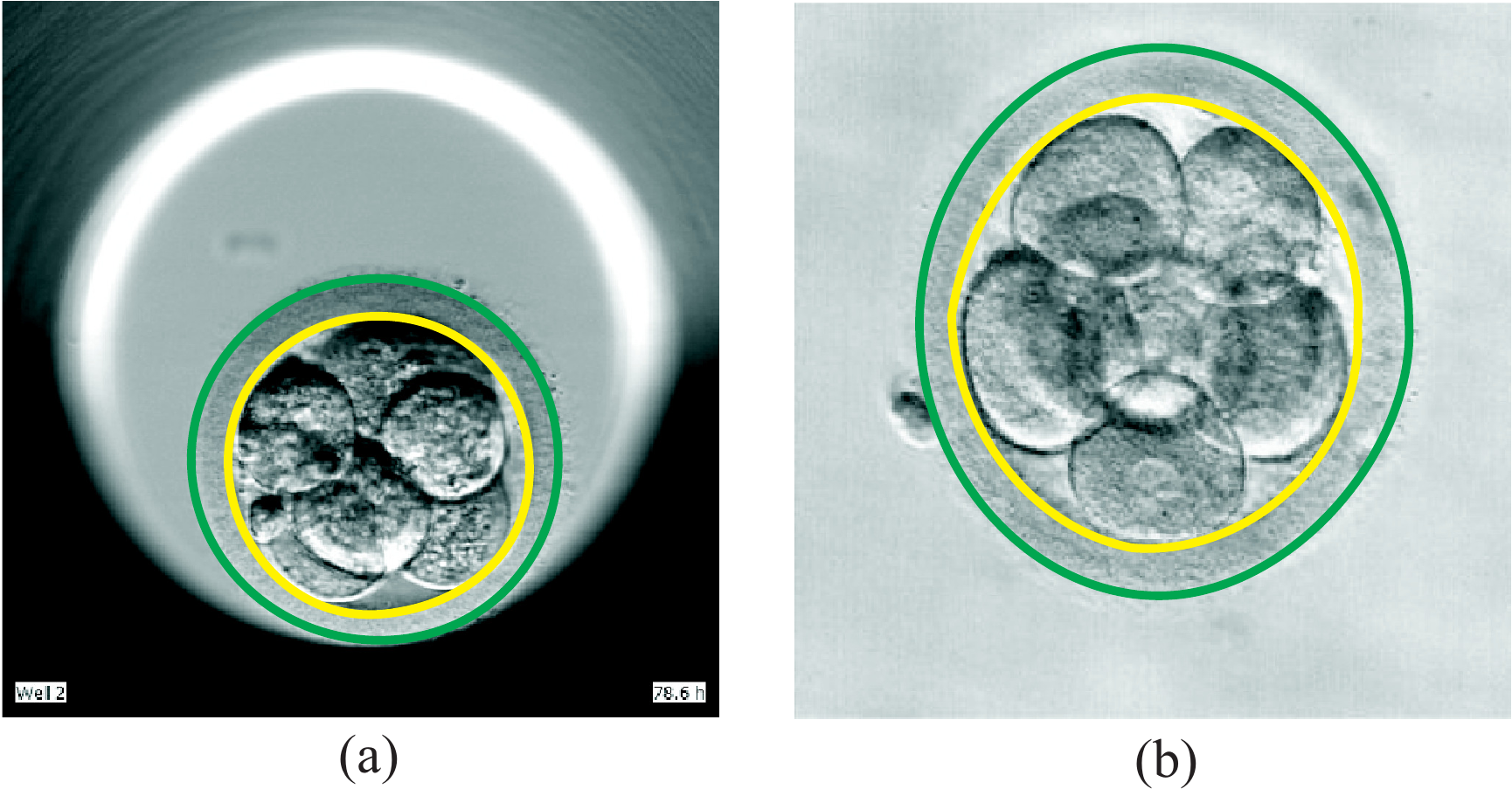}
  		\caption{\small{Sample embryo images with their inner (yellow) and outer (green) ZPs identified.}}
  		\label{fig:Inner}
\end{figure}

Embryo cells are protected by a membrane layer called Zona Pellucida (ZP). In Fig. \ref{fig:Inner}, we have highlighted the inner and outer boundaries of ZP with yellow and green colors, respectively. 
Detection of inner ZP boundary in this paper is important because: 
\begin{enumerate}
\item Determining the region inside the inner ZP confines the region of interest, and therefore reduces the computational complexity. 
\item The volume contained by the inner ZP provides estimates for potential ranges of blastomere sizes (see Section \ref{predSize}). 
\item  No blastomere candidate can physically cross ZP inner boundary. 
\item It will allow us to differentiate between the edge clusters of blastomeres and the edge clusters of the inner ZP boundary (Section~\ref{subsec:removeInner}). 
\end{enumerate}

Detection of inner ZP is carried out using an algorithm developed in one of our previous works \cite{DYee2013}. However, for images acquired by the embryoscope, we first use the well property to automatically remove edges corresponding to the well. 
%
%
\subsubsection{Edge Co-Association}
Human embryos, as well as their internal blastomere cells, are 3D spherical structures that are projected onto 2D image planes when photographed. 
Blastomeres are also semi-transparent structures and often (after the one-cell stage) they partially cover each other at different planes.  When imaged at one focal plane, the edges corresponding to the boundaries of one-cell might not look as sharp as they should especially if they are partially/fully occluded by other cells. This phenomenon as well as the loss of dimensionality, the imperfections associated with the imaging system, and the motion of blastomeres lead to image discontinuity of physical edges at various points.  To determine if close edge clusters belong
to the same blastomere, the following
metrics are calculated for each edge cluster and all of its neighbours:

\begin{itemize}
\item \textbf{Slope Variation:} The slope variation for each edge cluster before and after connecting to its neighbouring clusters are calculated. 
 A variation smaller than $\pi/8$ is allowed.
\item \textbf{Centroid Location:} A cluster's centroid after the neighbouring cluster is connected to it must remain close to the original centroid. A maximum distance deviation of 25\% (in pixel) is allowed in the new centroid's location after connecting the two. 
\item \textbf{Concavity Consistency:} The concavity before and after the connection
should not change. Concavity is measured by passing a parabola through every three consequent points. More specifically, for three points $(x_1, y_1)$, $(x_2, y_2)$ and $(x_3, y_3)$ the concavity is determined by the sign of the quadratic term ($a$) of a parabola that passes through these three points. Ensuing some algebra, we can show:
\begin{equation}\begin{split}
	a=&\frac{2y_1}{\left(x_1 - x_2\right)\left(x_1 - x_3\right)} + \frac{2y_2}{\left(x_2 - x_1\right)\left(x_2 - x_3\right)}\\+&\frac{2y_3}{\left(x_3 - x_1\right)\left(x_3 - x_2\right)}
\end{split}\end{equation} 
\end{itemize}
If two neighbouring clusters fulfill these three conditions, they will be connected to form one larger cluster. 
Fig.~\ref{lineEdges}.(b) depicts the output edge clusters after this process for a sample case. 
\subsubsection{Removal of Inner ZP's Edge Clusters}\label{subsec:removeInner}
\begin{figure}[h]
	  \centering
	  	 \includegraphics[width=3.in]{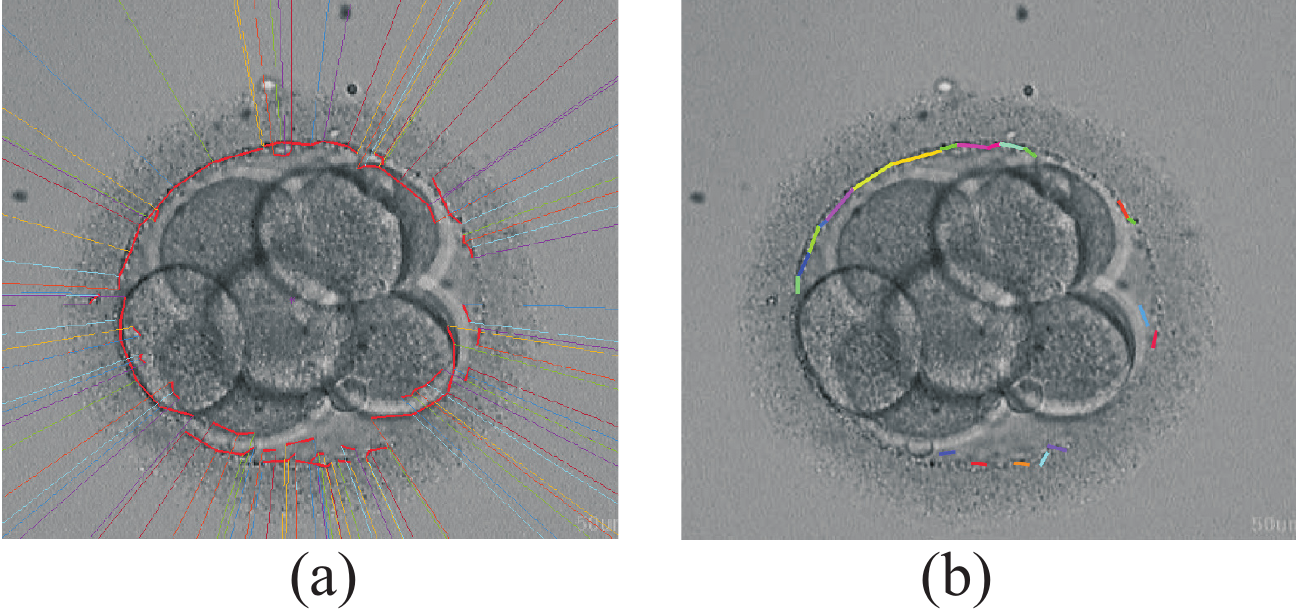}
	  \caption{\small{(a) Beams are radiated out to find the utmost edge clusters. (b) Edge clusters corresponding to the
inner ZP's boundary.}}
  		\label{fig:bound}
\end{figure}

Earlier in Section~\ref{sec:Inner}, we identified the inner ZP  boundary.  Our objective in this section is to identify all those edge clusters corresponding only to the inner ZP boundary and remove them. Two conditions (based on the parameter-tuning dataset) are checked to identify such edge clusters:
\begin{itemize}
\item distance of the edge cluster's all vertices from the inner ZP (maximum 2\% of inner ZP radius pixels), and
\item distance of the edge cluster's centroid from the inner ZP's centroid (maximum 10\% location variation allowed).
\end{itemize}
Figs.~\ref{fig:bound}.(a) and (b) show all clusters close to the inner ZP and those clusters corresponding to the inner ZP, respectively. 

\subsection{Blastomere Hypotheses Generation}
\subsubsection{Admissible Ellipse Sizes}\label{predSize}

Here, we try ellipses of different sizes and calculate their fitness scores. Detecting correct ellipses depends on choosing a proper range for sizes of the trial ellipses and such range must be decided according to the overall size of each embryo individually. 

\begin{figure}
	  \centering
	    \includegraphics[width=3.25in]{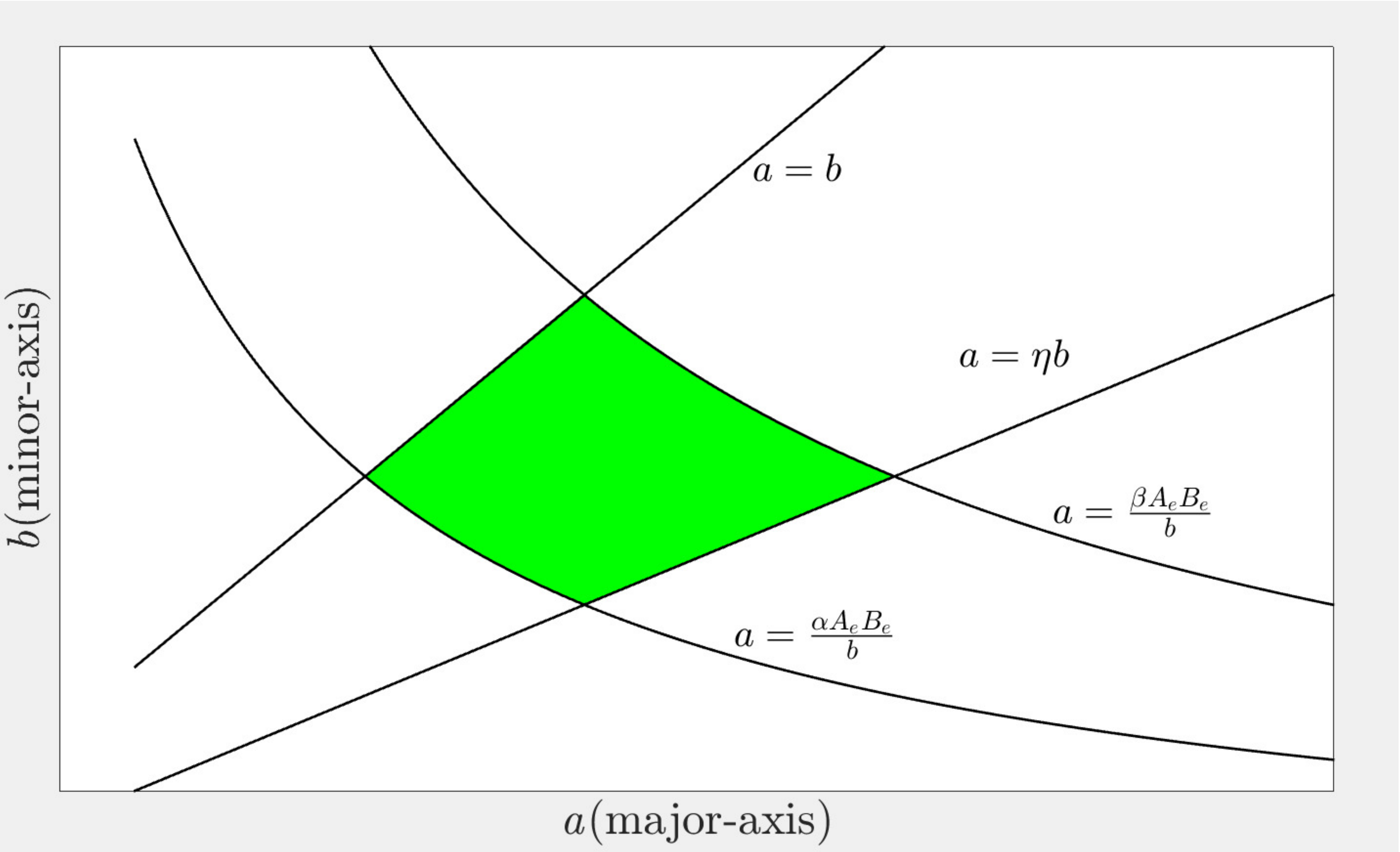}
  \caption{\small{Valid range of semi-major and semi-minor axes.}}
  \label{fig:ab}
\end{figure}

The inner ZP boundary ellipsoidal model, found in Section~\ref{sec:Inner}, guides us to the appropriate blastomere size range at different stages of growth. For example, when there is one cell only, the size is similar to the inner ZP. Once this cell cleaves into two cells, the size of each cell is somewhat closer to half the size of the inner ZP. 
If the semi-major and semi-minor axes of the inner ZP model are $A_e$ and $B_e$, we have:
\begin{equation}
  \alpha A_e B_e < a_i b_i < \beta A_e B_e
\end{equation}
where $a_i$ and $b_i$ are the candidate ellipse semi-major and semi-minor axes. The values for $\alpha$ and $\beta$ have a direct relationship with the number of blastomeres in the embryo. For an embryo with $n$ blastomeres they are defined as:  
\begin{equation}
\alpha = \frac{A}{n} ,~~~\beta= \min\left(\frac{B}{n} +C, 1\right)
\end{equation}
In this paper, $A$, $B$ and $C$ have been set to 0.7, 1, 0.15 (based on parameter-tuning dataset). 
With additional cleavages, the blastomeres tend to have more circular profiles. Also in case of surrounding forces, they may be squeezed, but the possibility of a highly squeezed ellipse is slim, and therefore we set the following condition:
\begin{equation}\label{eq:eta}
b_i < a_i < \eta  b_i,~~~~~
\eta > 1
\end{equation}
Equation (\ref{eq:eta}) limits the eccentricity of the candidate ellipses. (based on the parameter-tuning dataset, we) found $\eta =1.6$ to be more suitable for embryos with less than 6 blastomeres and 1.3 otherwise. 
Fig.~\ref{fig:ab} shows the range of eligible values for semi-major and semi-minor axes for generating appropriate candidates. Using this region, we find the minimum and maximum values of the semi-major and -minor axes for ellipse hypotheses to be generated before testing the fit.

\subsubsection{Hypothesis Generation}
The combination of the values for the semi-minor and semi-major axes values (from the previous section) represents a set of ellipses. Each ellipse can have different locations and orientations. The location, however, is confined within the Inner ZP layer. Therefore, all the boundary points of the candidate ellipse hypotheses must completely stay within the inner ZP boundary. To find the location and
orientation for each candidate ellipse with the highest correlation with 
the edge map, each candidate is rotated 18 times (incremental angle of  $10^\circ$) to cover a range of [$0^\circ$ $180^\circ$]; each time, we correlate the rotated model with the edge map. The correlation is performed in the Fourier transform domain and therefore is fast. The pixel with the maximum correlation shows the center of the best candidate ellipse. We keep the best candidate as well as its correlation score. 
We then remove edges corresponding to that candidate and continue the process of fitting. At the end of this process, we will have several potential candidates with different orientations and sizes and correlation scores.  
\subsection{Blastomere Detection}
Here, the compliance of the found candidates with image normals is verified. We calculate the normal vector direction at each boundary point of the candidate hypothesis and check whether they conform with the local image normals at those points. We then rate each candidate ellipse accordingly. For an ellipse, the general equation is:

\begin{equation}
ax^2+bxy+cy^2+dx+ey+f = 0
\end{equation}
The direction of the normal vector of each point $A(x_1,y_1)$ on the ellipse can be calculated by: 
\begin{equation}
\theta_{(x_1, y_1)} = \arctan{\frac{bx_1+2cy_1+e}{2ax_1+by_1+d}} + \frac{\pi}{2}
\end{equation}
A window of $5\times5$ pixels is utilized to calculate the normal direction at every edge point using the gradient image. 
\begin{figure}[htp!]
	  \centering
	  	 \includegraphics[width=3.4in]{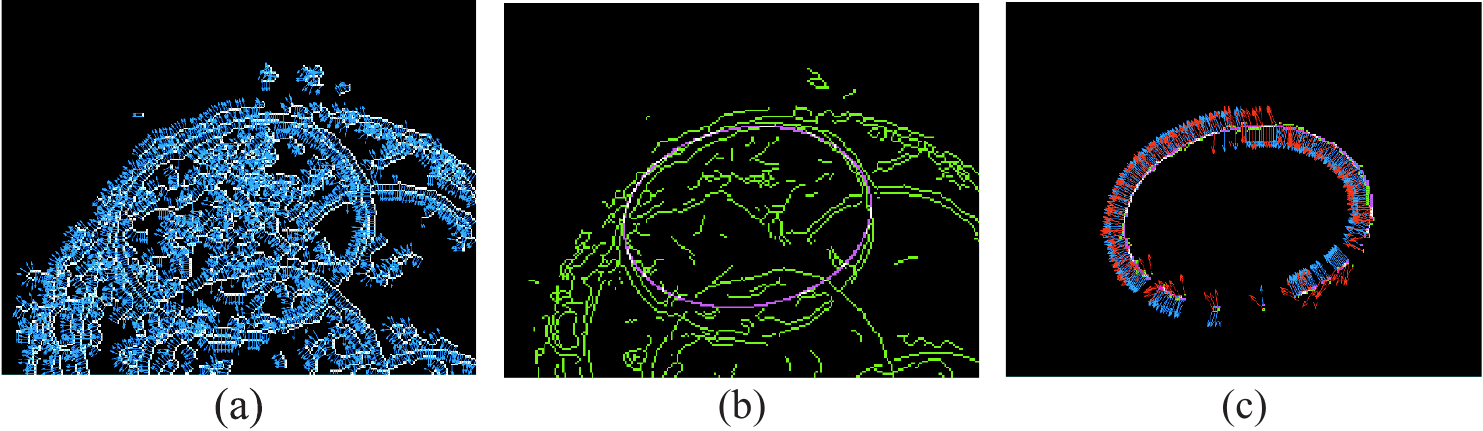}
  		\caption{\small{(a) Image normals over-imposed on the edge map. (b) A candidate ellipse over-imposed on the edge map. (c) Comparison of image normals with those of the candidate ellipse. }}
  		\label{fig:normalComparison}
\end{figure}
%
When finding edge points, the actual location of an edge point might be at some small offset from its real location. For this reason, when comparing the normal directions, if there is an edge pixel in the location of the ellipse boundary point, it is used for matching. Otherwise, we search for the closest point in the edge map, along the normal direction and its complement. A maximum distance of 5 pixels is allowed as potential displacement of the edge point from its real position. The angle difference between two points' normals must be smaller than $\pi/16$ to be considered a matching pair.

\begin{figure*}
\centering
\includegraphics[totalheight=0.23\textheight]{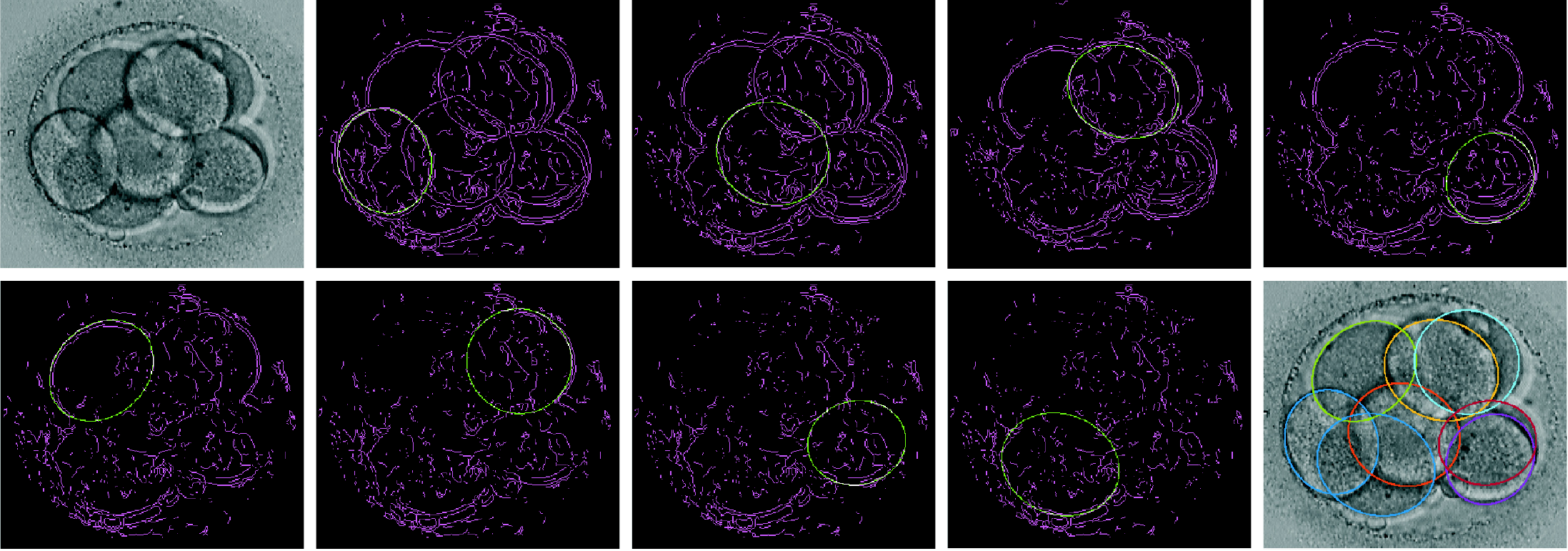}
  		\caption{\small{Step by step detection of ellipsoidal models matching the profile of the blastomeres.}}
  		\label{fig:MethodDetail}
\end{figure*}

Fig.~\ref{fig:normalComparison}-a and -b display an up-close look at the normal vectors and a candidate ellipse over-imposed on the image edge map. Fig.~\ref{fig:normalComparison}-c compares edge points with the candidate's boundary point normals. 
 A compliance metric (a value between zero to one) is calculated for each candidate hypothesis by the ratio of the number of points with similar normals to the total number of points on the boundary of the ellipse. Candidate with the highest normal score is chosen as the best candidate and all edges corresponding to the best match's boundary are removed from the edge map.  
This removal yields a cleaner edge map with more representative normals, which results in finding the occluded blastomeres. 
We repeat this process iteratively until all blastomeres are detected. A detailed visual representation of the method is depicted in Fig.~\ref{fig:MethodDetail}. 

%
\section{Image Acquisition}\label{I_AQ}
We evaluate our algorithm on a dataset of blastomere images acquired from three different sources: 
\begin{itemize}
\item[I)] Group PCRM-1: This group contains 166 images, all acquired by the Pacific Centre for Reproductive Medicine (PCRM). An Olympus IX71 inverted microscope with a Nomarski (DIC) optics and Research Instrument (RI) Cronus 3 software is used for imaging. The images are captured magnified by a factor of $1.6\times$ with a lens objective of $20\times$ and a resolution of $720 \times 479 \times 24$ Bit Per Pixel (BPP). 
\item[II)] Group PCRM-2: This group contains 217 images acquired by PCRM via Embryoscope by Vitrolife~\cite{UnisenseRef}, which includes a built-in microscope with Leica $20\times$ and 0.40 LWD Hoffman Modulation contrast objective. All the images are acquired at a resolution of $500\times 500 \times 8$ BPP.
\item[III)] Group WWW: This group contains 85 images that are obtained from the web (hence called WWW). This set especially allows us to confirm the stability of the proposed algorithm with respect to some of the used parameters.
\end{itemize}

\section{Embryo dataset and Ground Truth}\label{GT_sec}
The image dataset described above includes 468 HMC human embryo images of day 1 to day 3. PCRM-1 images are from 26 embryo sequences taken approximately 6.3 hours apart. PCRM-2 images are from 75 embryo sequences, one image per cell number (if available). WWW images are from different embryos and no similar images are considered in the dataset. To clarify the size of the dataset, we should mention that Khan \textit{et al.} \cite{khan2015linear} used 12 embryo sequences, Guitsi \textit{et al.} \cite{Giusti2010} used 53 4-cell images and Singh \textit{et al.} \cite{singh2014automatic} used 40 images in total (the number of embryos are not specified). The number of embryos is much more than the number used in other presented works.

Each image includes one complete embryo with 1 to 8 blastomeres and may contain fragmentation and/or some artifacts. 
Each embryo is inspected for all existing blastomeres (both visible and invisible to the human observer). Then the boundaries of each blastomere are manually outlined. Since the approach is model-based, we have generated two sets of Ground Truth (GT).   
In the first set, the boundary points are identified at a pixel by pixel level. The performance of the presented work is compared against this GT. We should mention that manual pixel-by-pixel extraction of GT is a time-consuming task, more than just a classification problem, and 468 images are considered a big dataset. 
In the second set, we have fitted the best ellipsoidal model that at best complies with the boundary of each blastomere. We have used this set to measure the fidelity of our model-based assumption for this work (\ref{fid_res}). 

Besides 468 images in the test dataset, a parameter-tuning dataset of 45 images are utilized to determine the value of the parameters throughout the paper. The parameter-tuning dataset is not used for evaluations.
\vspace{-3mm} 
\section{Performance}
The proposed system is implemented in MATLAB 8.5 platform and on a PC (CPU Intel Core i7 3.40-GHz with 8-GB RAM).
The input images of set PCRM-1 have a resolution of $720\times 479$ pixels. Images of PCRM-2 are $500\times500$ pixels and WWW images have various sizes usually smaller than the two other sets. The entire detection process takes an average of 180 seconds to complete.
\vspace{-3mm} 
\section{Results}
In this section, we analyze the performance of the proposed algorithm, compared to the Ground Truth as well as the state of the art algorithms in the literature. 

\begin{figure}
	  \centering
	  	 \includegraphics[width=3.4in]{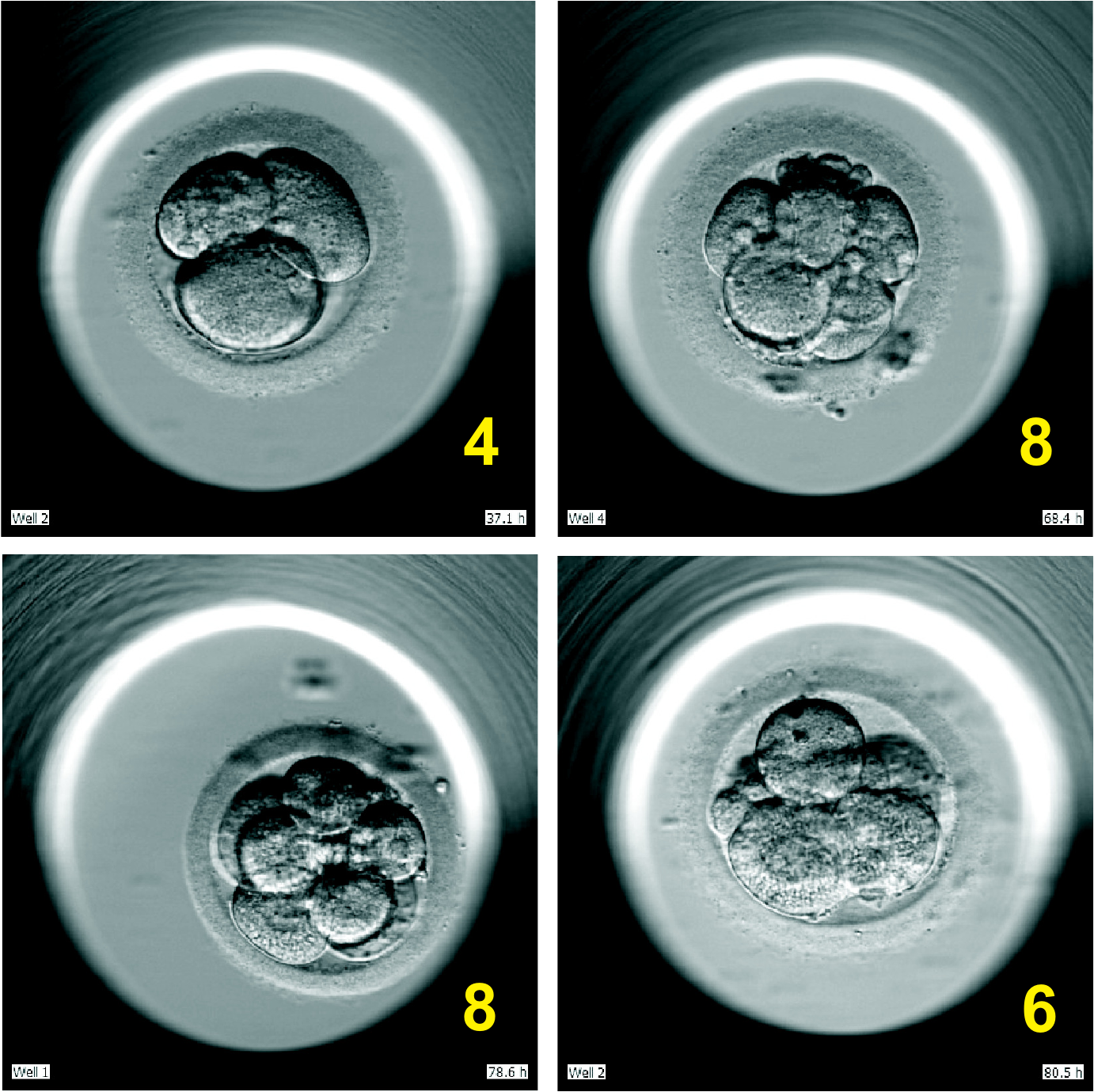}
  		\caption{\small{Embryo images with complete occlusions.}}
  		\label{fig:Diff}
\end{figure}

To highlight the inherent difficulty in the detection of blastomeres using a single microscopic image with a fixed focal length, three sample images are displayed in Fig.~\ref{fig:Diff}. The actual number of blastomeres in each image is shown with the bold yellow colour.  As can be seen, in some cases it is very difficult and sometimes impossible to see all existing blastomeres. In those cases, clearly our algorithm will be harshly penalized, even though, it might be rather unfair to expect to detect blastomeres that have no visible or obvious sign of their existence due to complete occlusions. All the results in this paper have been performed on all the images in the dataset with no pre-selection (except Table IV), no image resizing and no non-automatic pre-processing has been applied.

\subsection{Quantitative Results}
After detecting blastomeres in each image, we compare the results with that of the GT. We have utilized three metrics of: \textit{Precision}, \textit{Sensitivity} and \textit{OQ} to represent the performance quality of the proposed method. 
%
%
\begin{subequations}\label{Sensitivity}\begin{align}
Precision=&\frac{TP}{TP+FP},\\
Sensitivity=& \frac{TP}{TP+FN},\\
Overall~Quality=& \frac{TP}{TP+FN+FP},
\end{align}\end{subequations} 
where TP, FP, FN denote the True Positives (correctly identified), False Positives (incorrectly identified) and False Negatives (wrongly missed) regions, respectively. All these three metrics are defined at the pixel level and by construction, their optimal value is 1.


\begin{table}
\centering
\caption{\small{Mean Precision ($Pre.$), Sensitivity ($Sen.$) and Overall Quality ($OQ$) measures for 1 to 8-cell test cases. Results are separated to show the performance with and without the presence of artifacts. NOI stands for Number of Images and Art. for Artifact.}}
\label{tab:results}
\begin{tabular}{ | c | c | c | c | c | c | c | c | c | c | c | c | c | c | c | c | c | c | }
\hline
	&\multicolumn{2}{|c|} {\small{ 1-cell}} &  \multicolumn{2}{|c|} {\small{2-cell}} &  \multicolumn{2}{|c|} {\small{3-cell}} &   \multicolumn{2}{|c|} {\small{4-cell}} &   \multicolumn{2}{|c|} {\small{5-cell}} & \multicolumn{2}{|c|} {\small{6-cell}} &  \multicolumn{2}{|c|} {\small{7-cell}} &  \multicolumn{2}{|c|} {\small{8-cell}} & \vtop{\hbox{\strut \small{Sum}/}\hbox{\strut \small{mean}}} \\ \hline
	\small{Art.?} & \scriptsize{No} & \scriptsize{Yes} & \scriptsize{No}  & \scriptsize{Yes} &\scriptsize{No}  & \scriptsize{Yes} & \scriptsize{No}  & \scriptsize{Yes} & \scriptsize{No}  & \scriptsize{Yes} &  \scriptsize{No} & \scriptsize{Yes} & \scriptsize{No} & \scriptsize{Yes} & \scriptsize{No}  & \scriptsize{Yes} &  \\ \hline
	\footnotesize{NOI} & \scriptsize{72} & \scriptsize{60} & \scriptsize{15 }& \scriptsize{56} & \scriptsize{6} & \scriptsize{16} &  \scriptsize{33} & \scriptsize{74} & \scriptsize{4} & \scriptsize{16} & \scriptsize{5} & \scriptsize{26} & \scriptsize{3} & \scriptsize{26} &  \scriptsize{20} & \scriptsize{36} &  \footnotesize{468} \  \\ \hline
	\footnotesize{$Pre.$} & \scriptsize{0.98} &\scriptsize{0.96} & \scriptsize{0.95} & \scriptsize{0.94} & \scriptsize{0.91} & \scriptsize{0.89} & \scriptsize{0.91} & \scriptsize{0.90} & \scriptsize{0.90} & \scriptsize{0.90}  & \scriptsize{0.83}& \scriptsize{0.85} & \scriptsize{0.90} & \scriptsize{0.87}  & \scriptsize{0.89} & \scriptsize{0.84} & \footnotesize{0.92}\  \\ \hline
	\footnotesize{$Sen.$} & \scriptsize{0.97} &\scriptsize{0.96} &\scriptsize{0.93} &\scriptsize{0.91} & \scriptsize{0.90} & \scriptsize{0.82} & \scriptsize{0.85} & \scriptsize{0.83} & \scriptsize{0.76}  & \scriptsize{0.80} & \scriptsize{0.83} & \scriptsize{0.80} & \scriptsize{0.80} & \scriptsize{0.81} & \scriptsize{0.86}  & \scriptsize{0.82} & \footnotesize{0.88}\  \\ \hline
\small{$OQ$} & \scriptsize{0.95} &\scriptsize{0.92} &\scriptsize{0.88} &\scriptsize{0.85} & \scriptsize{0.85} & \scriptsize{0.77} & \scriptsize{0.80} & \scriptsize{0.78} & \scriptsize{0.73}  & \scriptsize{0.76} & \scriptsize{0.72} & \scriptsize{0.72} & \scriptsize{0.74} & \scriptsize{0.74} & \scriptsize{0.80}  & \scriptsize{0.74} & \footnotesize{0.83}\  \\ \hline
\end{tabular}
\end{table}

Table \ref{tab:results} details the average \textit{Precision}, \textit{Sensitivity} and \textit{OQ} for all images in our dataset separated according to the number of cells and the presence of artifact. We define artifact by the presence of fragmentation (\textgreater{10\%} fragmentation \cite{fragment}) inside the embryo and/or floating cells or debris on the background of the input embryo images. Fragments are small bleb pieces that are produced as a result of the breakage of cytoplasm during the embryo division. While the causes of fragmentation are not entirely understood, we know that multiple fragmentations in an embryo may lead to reduced cellular machinery, and hence underdevelopment of the embryo [\cite{stone2005embryo} needed here]. More interestingly, fragmentation is a unique feature of human embryos and makes it more difficult to identify the boundaries of the blastomeres. Among 468 total images, average \textit{Precision} of 0.92, \textit{Sensitivity} of 0.88 and \textit{OQ} of 0.83 are obtained. 

The reason for the fewer number of images in 3, 5, 6 and 7 cases, is that in the normal development of an embryo, 1, 2, 4 and 8 cell cases will happen. 3, 5, 6 and 7 seven cases are intermediate stages happening rarely when the division of the cells is not happening at almost the same time. As our images are taken a few hours apart, the number of these cases (according to the nature of embryo development) is not high.

Table~\ref{tab:batch} demonstrates the robustness of the algorithm with respect to different input image sources. It should be noted that the same parameter values are used for all images regardless of their sources in a fully automatic way. 

\begin{table}[!h]
\centering
\caption{\small{Number of images($NOI$), mean Precision ($Pre.$), Sensitivity ($Sen.$) and  Overall Quality ($OQ$) measures according to the source of input images.}}
\label{tab:batch}
\begin{tabular}{ | c | c | c | c | }
\hline
	&\small{ PCRM-1} &  \small{PCRM-2} & \small{WWW}  \\ \hline
	\small{$NOI$} & \small{166} & \small{217} & \small{85}  \\ \hline
	\small{$Pre.$} & \small{0.93} & \small{0.91} & \small{0.90}  \\ \hline
	\small{$Sen.$} & \small{0.92} & \small{0.83} & \small{0.91} \\ \hline
	\small{$OQ$} & \small{0.88} &\small{0.79} & \small{0.83}  \\ \hline

\end{tabular}
\end{table}

In this work, we consider that a blastomere is correctly detected if its \textit{OQ} value is larger than a predefined threshold when compared with the GT. 
Khan~\cite{khan2015linear} considered an OQ value of 0.7 as the threshold, whereas Guitsi~\cite{Giusti2010} used a higher value of 0.8. Table ~\ref{tab:resultsperNum} shows the percentage of detected blastomeres in each embryo image, for 1-8 cell embryo images, with \textit{OQ} threshold of 0.7. For example, for 8-cell embryos, in 58\% of images, all 8 cells in the image were correctly identified. In 25\% of images, one cell in each image were incorrectly identified (\textit{OQ} \textless 0.7) and in 17\% of images, 2 cells were incorrectly identified. We noticed that embryo images with a fewer number of cells are usually detected with higher \textit{Precision} and \textit{Sensitivity} values. As the number of blastomeres increases, it becomes harder to find blastomeres due to the occlusions that originate from an overcrowded embryo. 

\begin{table}[!h]
\centering
\caption{\small{Percentage of blastomeres detected in each embryo image according to the number of visible cells. Results are rounded to the closest integer.}}
\label{tab:resultsperNum}
\begin{tabular}{ | c | c | c | c | c | c | c | c | c | c|  }
\hline
&\multicolumn{9}{|c|} {\small{ Percentage of detected blastomeres.}}\\ \hline
        &\small{0} & \small{1} &  \small{2} &  \small{3} &  \small{4} &  \small{5} & \small{6} & \small{7} &  \small{8} \\ \hline
	\small{1-cell} &\small{1}   & \small{99}    & \small{-}  & \small{-}   & \small{-}  & \small{-}  & \small{-}  & \small{-}   & \small{-}  \\ \hline
	\small{2-cell} &\small{3}   & \small{10}    & \small{87} & \small{-}   & \small{-}  & \small{-}  & \small{-}  & \small{-}   & \small{-}  \\ \hline
	\small{3-cell} &\small{0}   & \small{14}     & \small{27} & \small{59}  & \small{-}  & \small{-}  & \small{-}  & \small{-}   & \small{-}  \\ \hline
	\small{4-cell} &\small{0}   & \small{0}     & \small{8} & \small{26}  & \small{67} & \small{-}  & \small{-}  & \small{-}   & \small{-}  \\ \hline
	\small{5-cell} &\small{0}   & \small{0}    & \small{2}  & \small{17} & \small{32} & \small{48}  & \small{-}   & \small{-}  & \small{-}\\ \hline
	\small{6-cell} &\small{0}   & \small{0}     & \small{2}  & \small{5}   & \small{5}  & \small{20} & \small{68} & \small{-}   & \small{-}  \\ \hline
	\small{7-cell} &\small{0}   & \small{0}     & \small{4}  & \small{0}   & \small{4}  & \small{8} & \small{17} & \small{67}  & \small{-}  \\ \hline
	\small{8-cell} &\small{0}   & \small{0}     & \small{0}  & \small{0}   & \small{0}  & \small{0} & \small{17} & \small{25}  & \small{58} \\ \hline
\end{tabular}
\end{table}

\begin{figure*}
	  \centering
	  	 \includegraphics[width=5.6in]{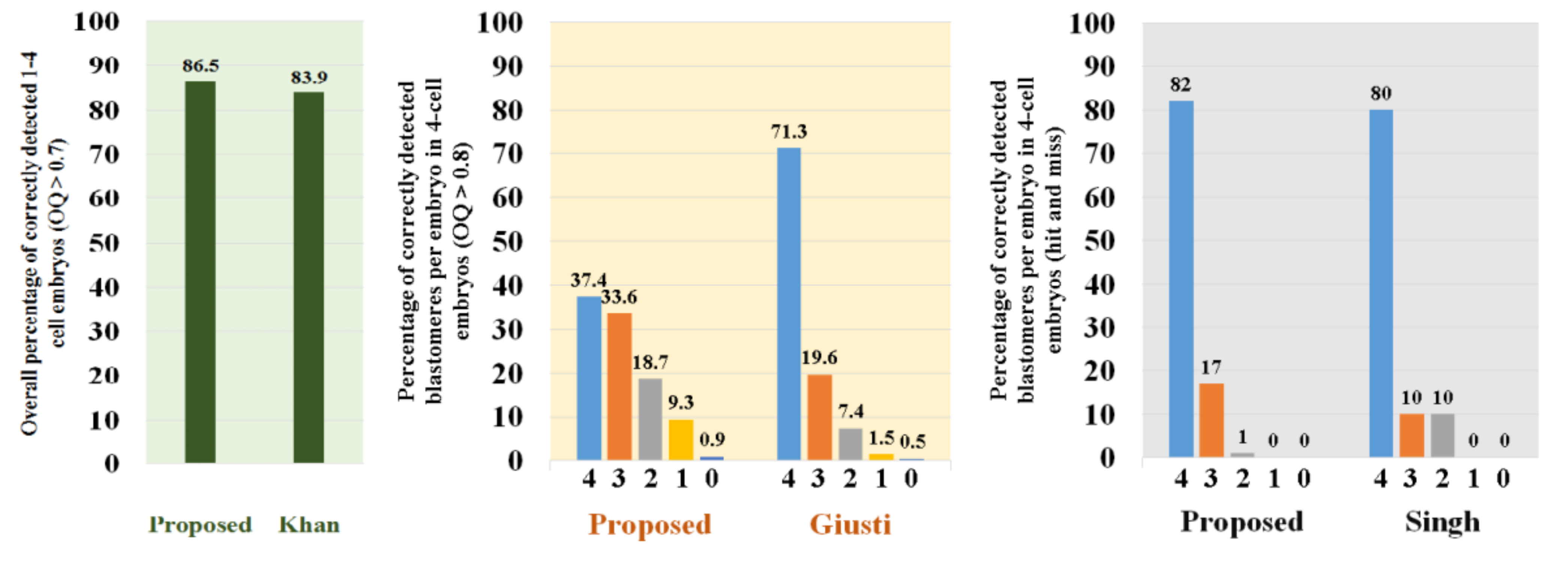}
  		\caption{\small{Comparison of the proposed method with Khan \cite{khan2015linear}, Giusti ~\cite{giusti2009lighting} and Singh ~\cite{singh2014automatic}.}}
  		\label{fig:comamgui}
\end{figure*}

The results of our work are compared against those of three known state-of-the-arts with reported quantitative results. All these works, however, are for embryos with 1 to 4 cells and therefore no comparison for embryos with 5 to 8 cells can be presented at this time. We requested Khan's group to provide us with their dataset for a more fair performance evaluation. Our request, however, was rejected. Fig.~\ref{fig:comamgui} displays such comparison. It should be noted that~\cite{singh2014automatic} and~\cite{Giusti2010} discussed their results only for images with 4 cells. Reference~\cite{khan2015linear}, however, reported the overall results for all images of 1 to 4 cells.




\begin{figure*}
	  \centering
	  	 \includegraphics[width=4.5in]{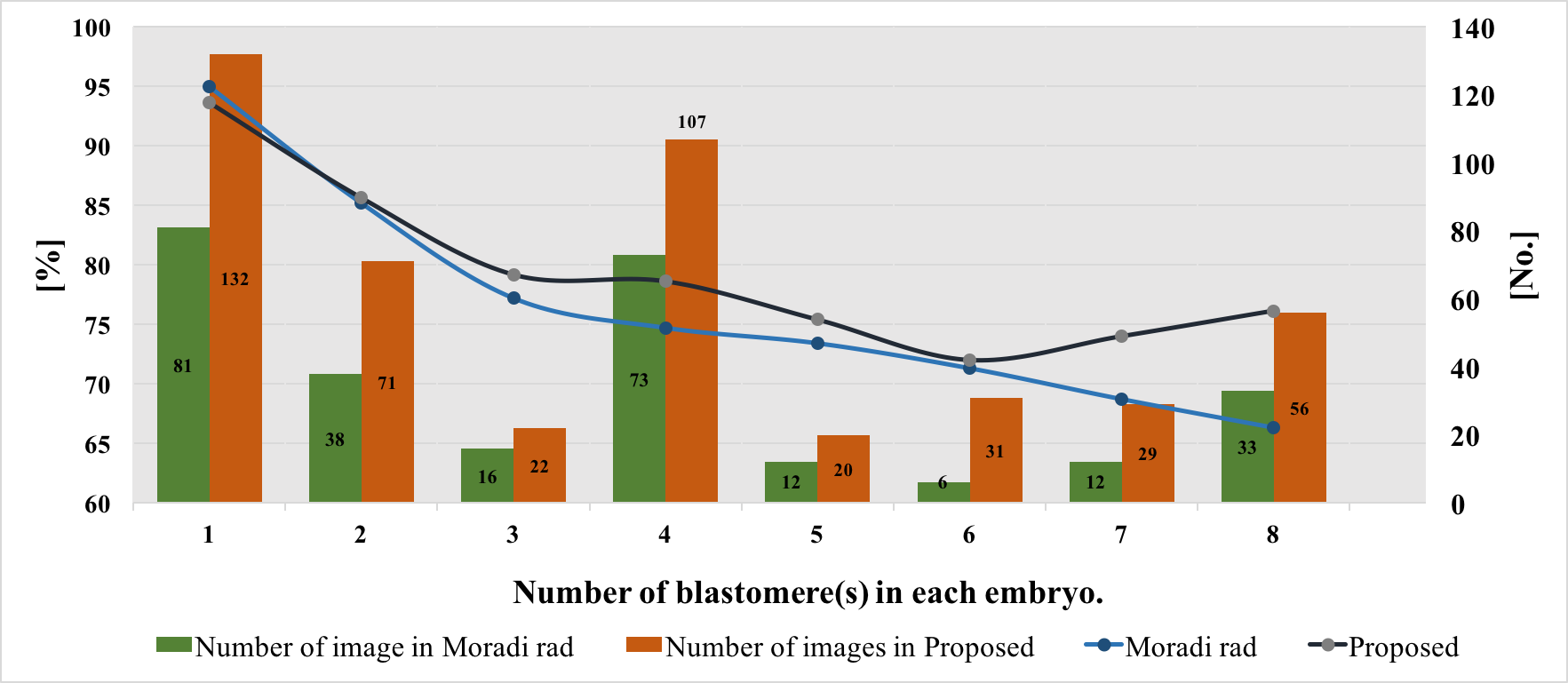}
  		\caption{\small{Comparison of the proposed method with Moradi rad \cite{rad2018hybrid} based on the number of images and the \textit{OQ} for embryos with 1 to 8 blastomeres. }}
  		\label{fig:cmpmoradi}
\end{figure*}

Khan~\textit{et al.}~\cite{khan2015linear} utilized 12 sequences of developing embryo images using dark-field microscopy, with a total number of 4122 frames taken 5 minutes apart from two patients. They used the sequence of frames and their relation to detect and localize blastomeres by finding the most likely sequence of hypotheses over all time. In comparison, our work is focused on detecting blastomeres on one frame. They report 83.9\% success rate in identifying cells with~\textit{OQ} values of higher than 0.7. As shown in Fig.~\ref{fig:comamgui}, the proposed method achieves 86.5\% success rate using the same~\textit{OQ} threshold value. 

Reference \cite{singh2014automatic} considered the detection of a blastomere as a hit/miss problem based on visual perception of the result. Using a similar hit/miss approach, we obtained 82\% for 4 cells as shown in Fig.~\ref{fig:comamgui}. The number of 4-cell images in our study is 107, whereas the overall number of all 1 to 4-cell images in~\cite{singh2014automatic} was only 40, all of which are a subset of our dataset.

Giusti~\textit{et al.} \cite{Giusti2010} used multiple images of the same embryo taken at different focus planes, making the process of finding the precise boundaries of each blastomere much more accurate. They used a threshold of 0.8 for the \textit{OQ} measure when considering whether a blastomere was detected or not. The better performance of Giusti's method is dute to the multiple images of the same embryo at different focus planes.  

Moradi rad~\textit{et al.} \cite{rad2018hybrid} utilized 271 images, a subset of our dataset, consisting of PCRM and WWW images to evaluate their method. Fig.~\ref{fig:cmpmoradi} compares the proposed method to \cite{rad2018hybrid} based on the number of blastomere cells in an embryo and \textit{OQ}. As you can see, our method performs better in all cases, except for the one-cell images.

\subsection{Qualitative Results}
Table~\ref{mynew-1-8cells} showcases a typical set of results for 24 embryo images. The boundaries of the detected blastomeres are highlighted in different colors. We have categorized the visual data by the number of cells and according to the \textit{OQ} of the produced output by our system. In each case, the GT boundaries are highlighted in white on the input image and shown on the right side of the output results. 

\begin{table*}[!ht]
\centering
\caption{\small{Output of the proposed algorithm along with the GT for sample images of 1 to 8-cells.}}
\label{mynew-1-8cells}
\begin{tabular}{|c|c|c|c|c|c|c|}
\hline 
 &\multicolumn{2}{|c|}{Good}& \multicolumn{2}{|c|}{Average}& \multicolumn{2}{|c|} {Bad} \\ 
&\multicolumn{2}{|c|} {$\ 0.8\leq$OQ$\leq 1$}&\multicolumn{2}{|c|} {$0.7 \leq$OQ$< 0.8$}& \multicolumn{2}{|c|}{$$OQ$ < 0.7$} \\\hline 
& {Our Method} & {Ground Truth}& {Our Method} & {Ground Truth}& {Our Method} & {Ground Truth} \\ \hline\hline
{\rotatebox{90}{ 1-cell}} &  \includegraphics[trim=0cm 0cm 0cm 0cm, clip=true, width=.13\textwidth, height=20mm]{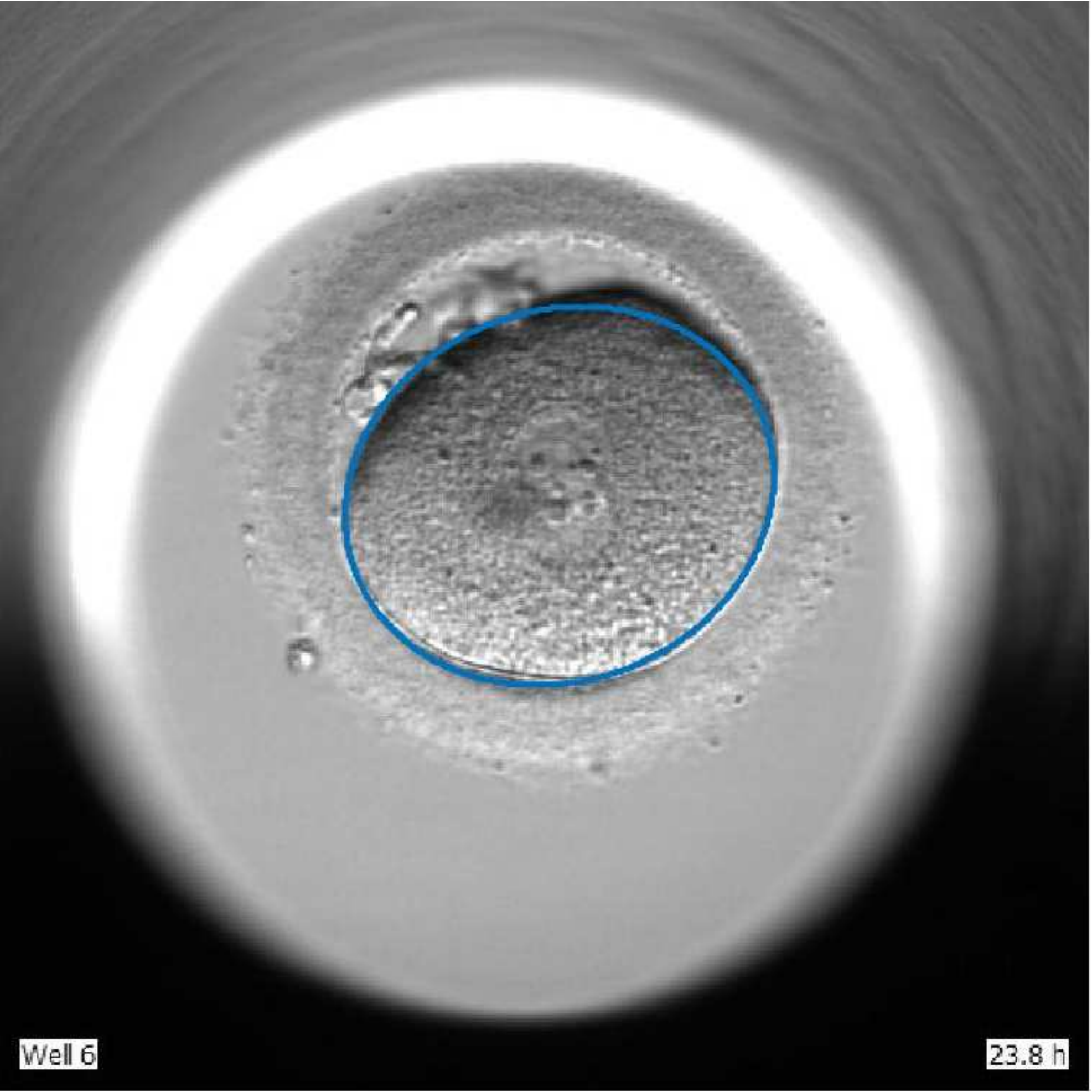} &  
\includegraphics[trim=0cm 0cm 0cm 0cm, clip=true, width=.13\textwidth, height=20mm]{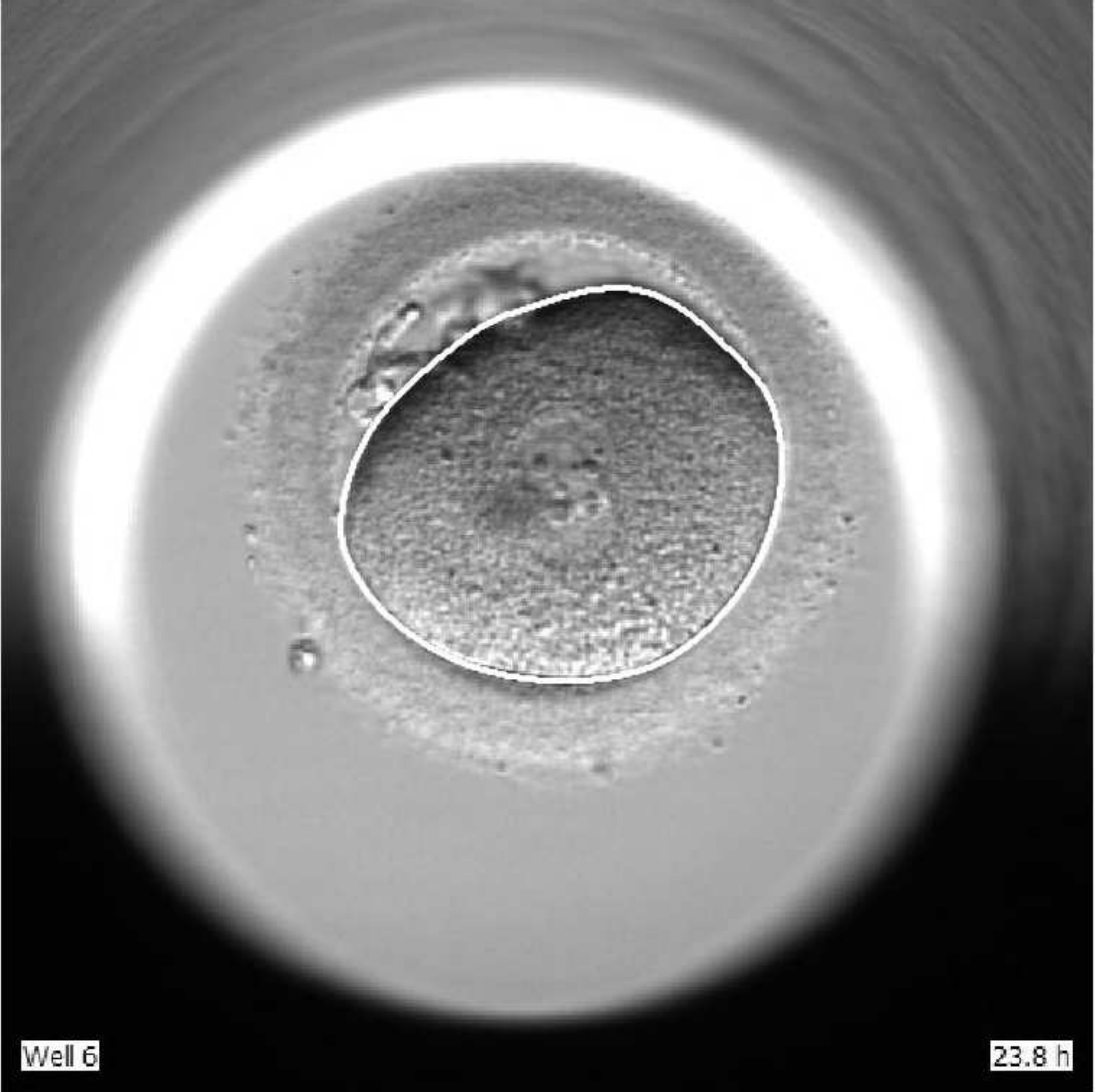} &  \includegraphics[trim=0cm 0cm 0cm 0cm, clip=true, width=.13\textwidth, height=20mm] {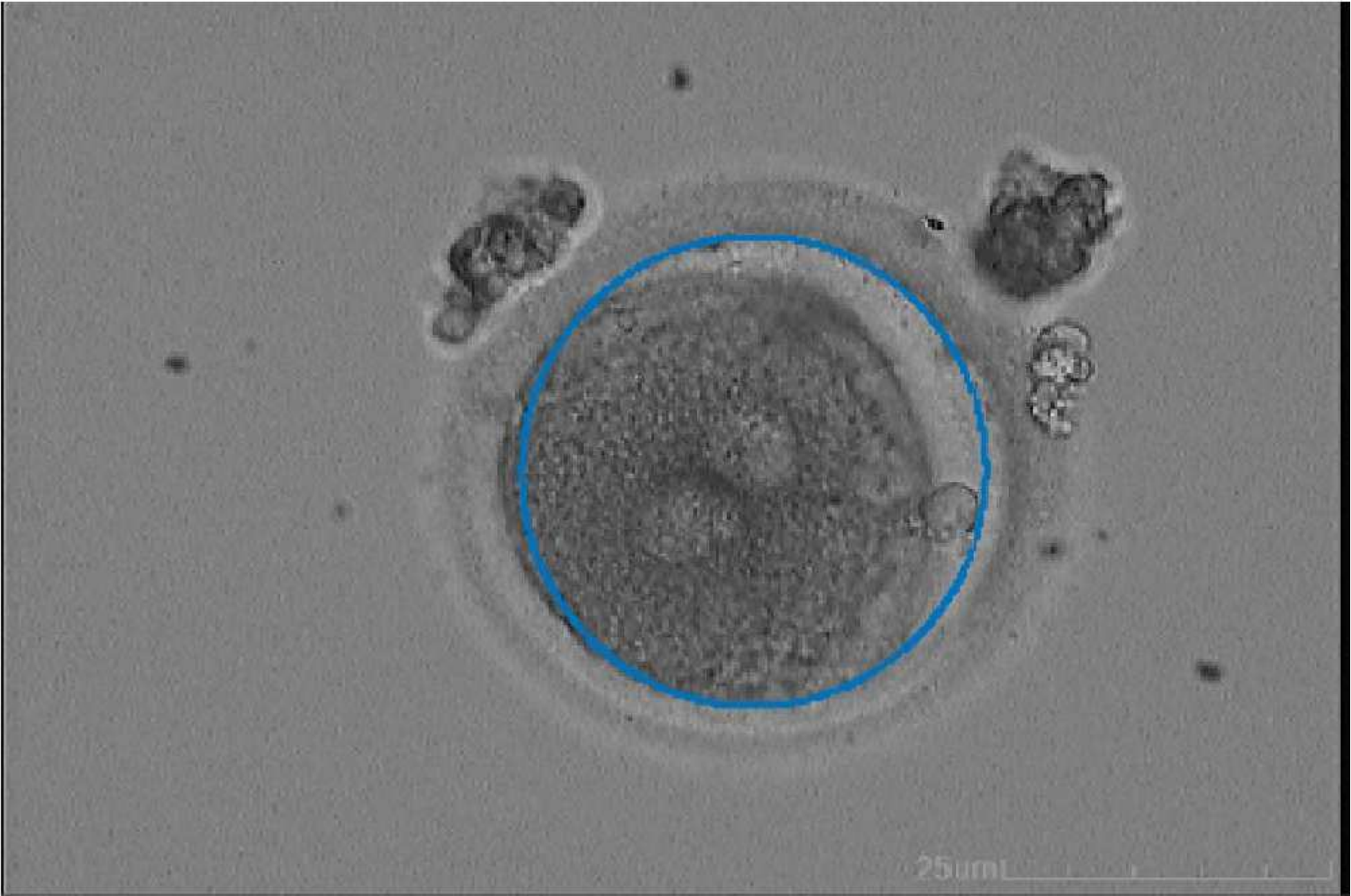}  
& \includegraphics[trim=0cm 0cm 0cm 0cm, clip=true, width=.13\textwidth, height=20mm]{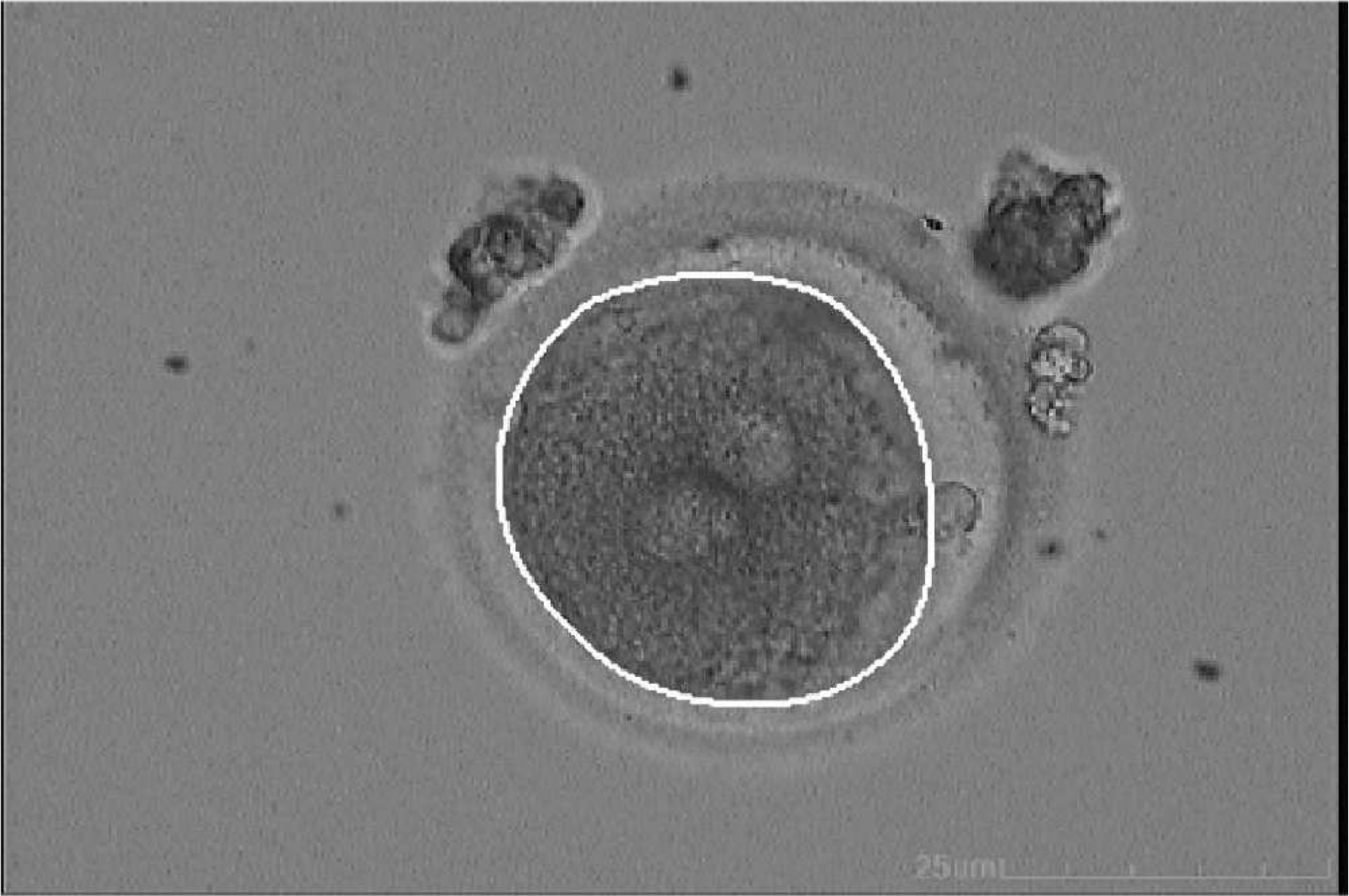} &  \includegraphics[trim=0cm 0cm 0cm 0cm, clip=true, width=.13\textwidth, height=20mm]{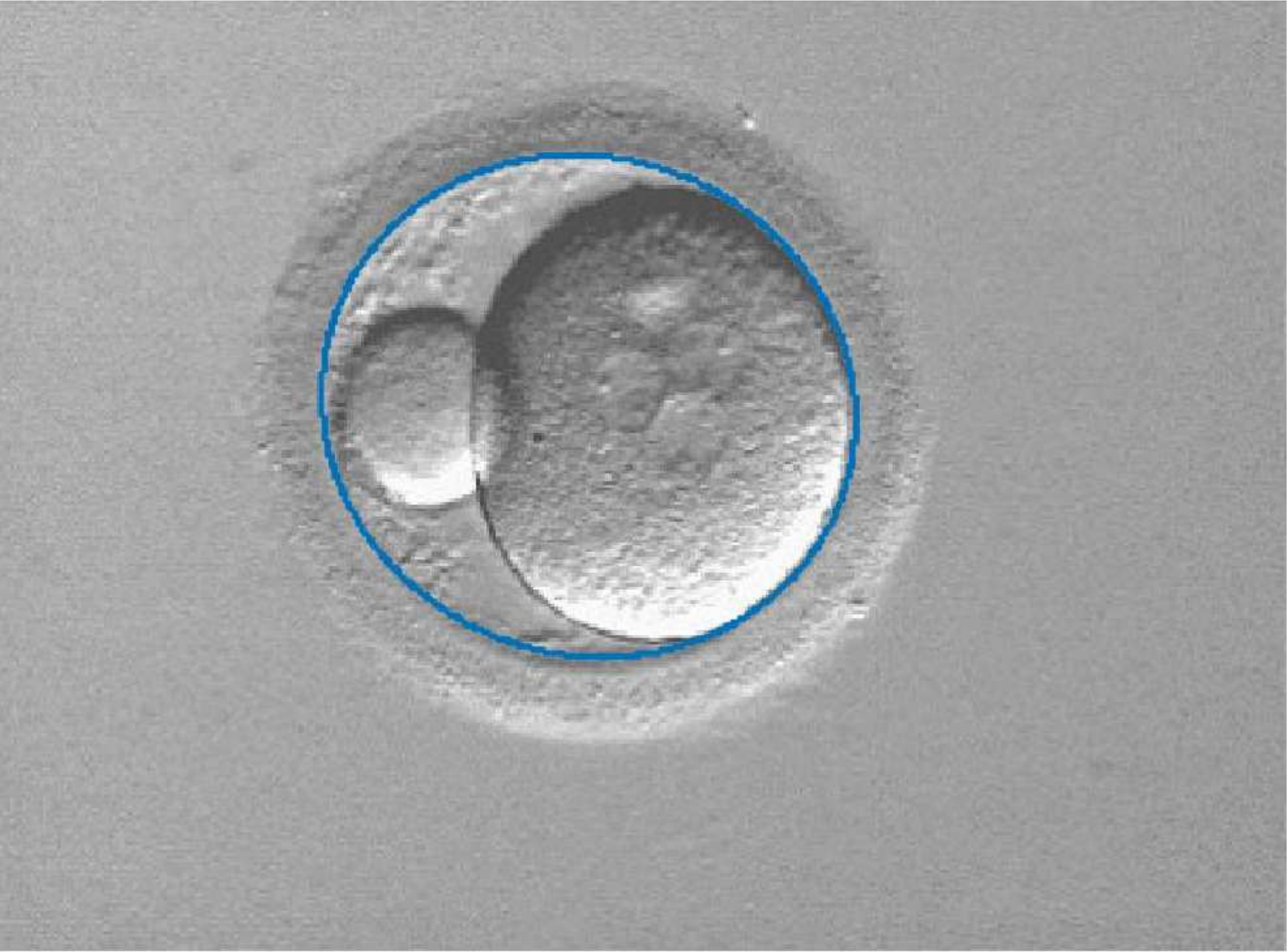} &  \includegraphics[trim=0cm 0cm 0cm 0cm, clip=true, width=.13\textwidth, height=20mm]{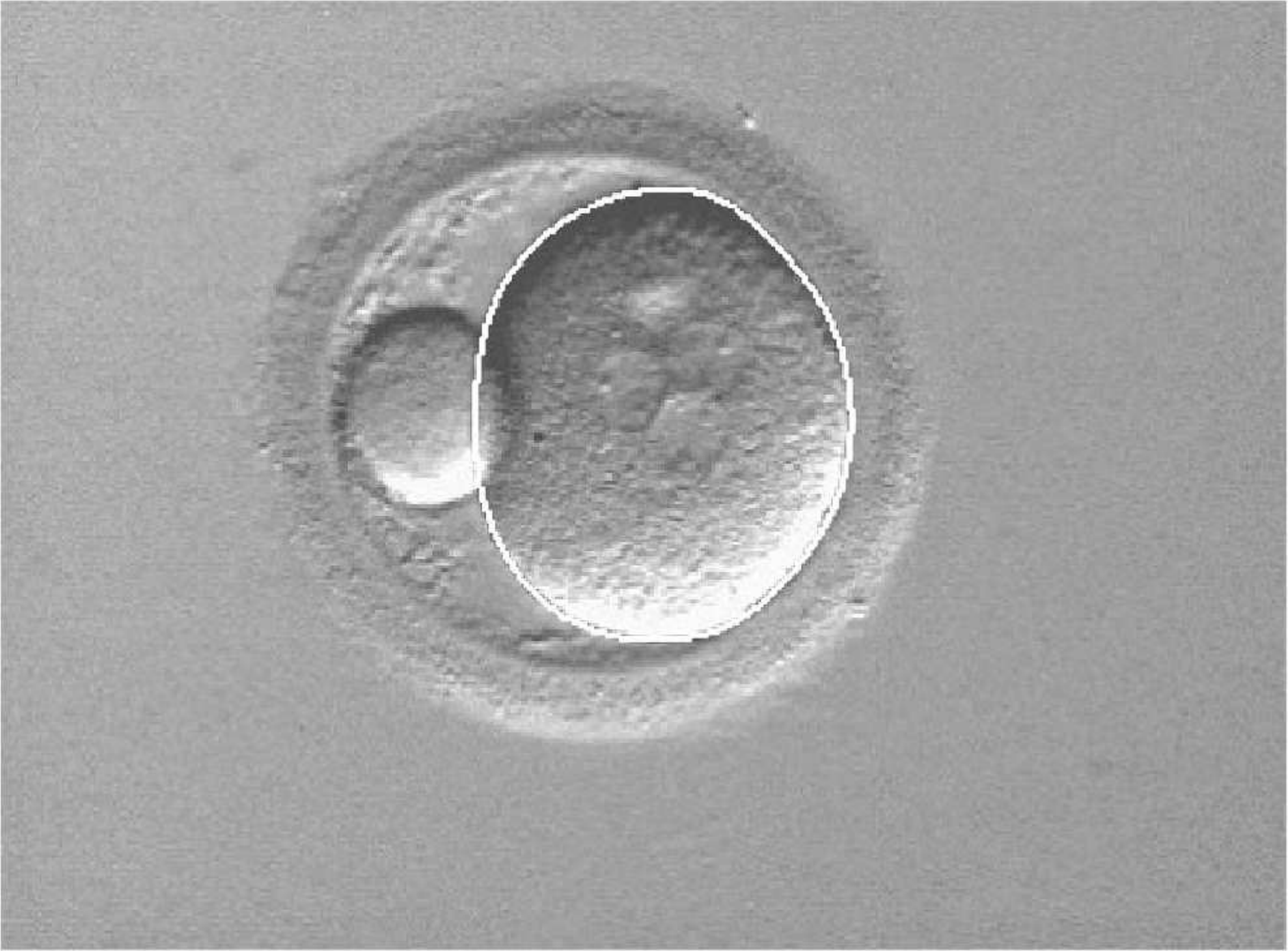} \\ \hline
{\rotatebox {90}{ 2-cell} } &  \includegraphics[trim=0cm 0cm 0cm 0cm, clip=true, width=.13\textwidth, height=20mm]{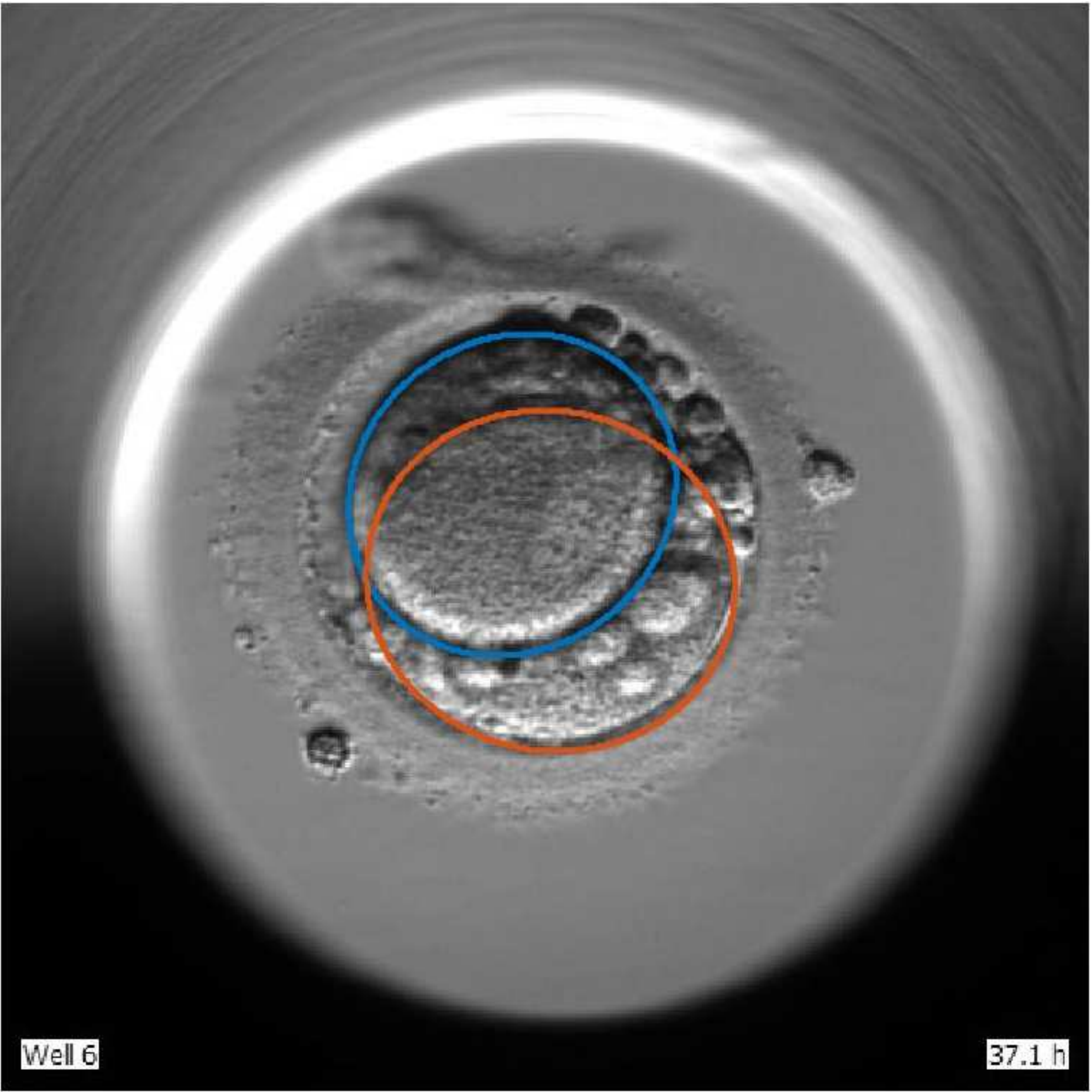} &  \includegraphics[trim=0cm 0cm 0cm 0cm, clip=true, width=.13\textwidth, height=20mm]{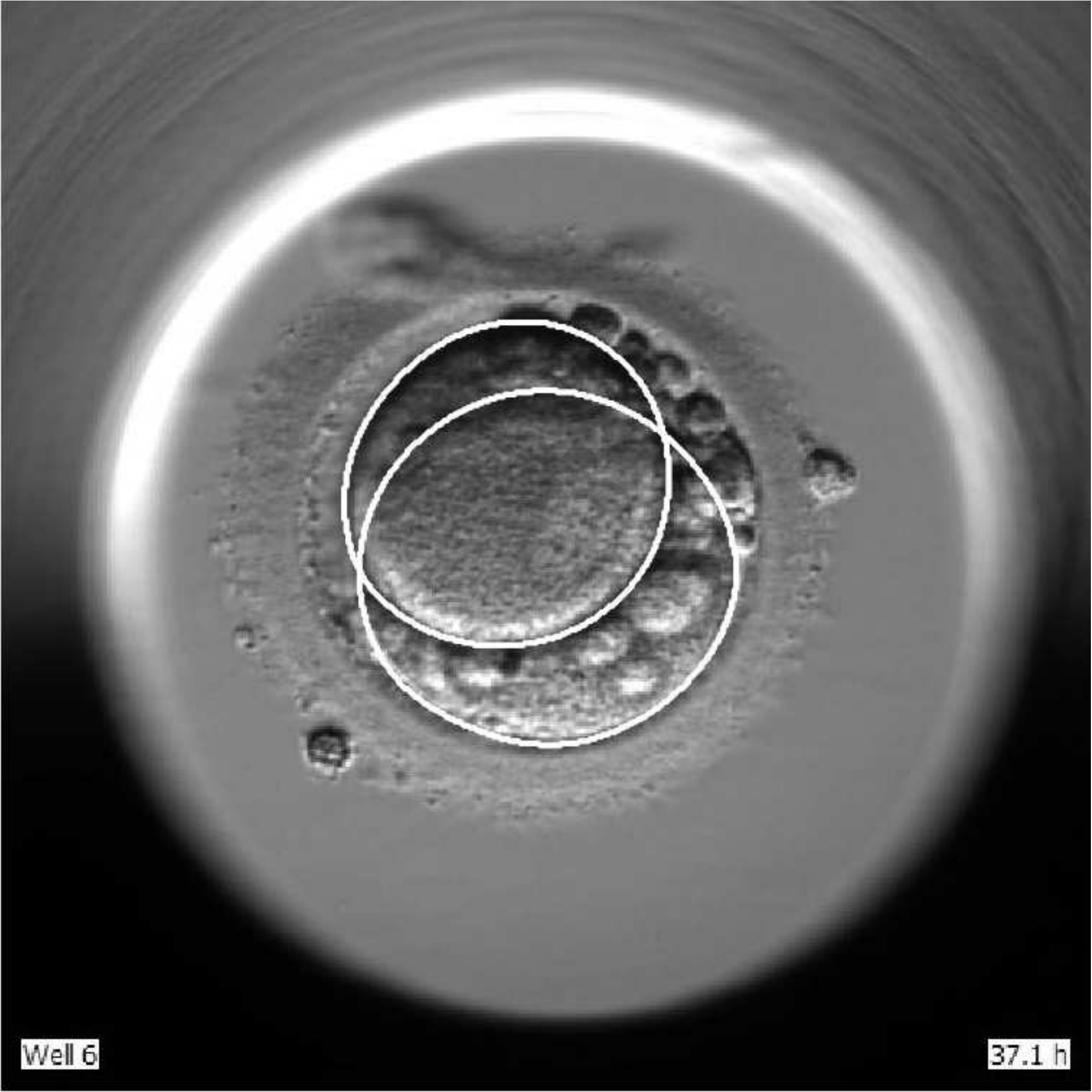} 
&  \includegraphics[trim=0cm 0cm 0cm 0cm, clip=true, width=.13\textwidth, height=20mm] {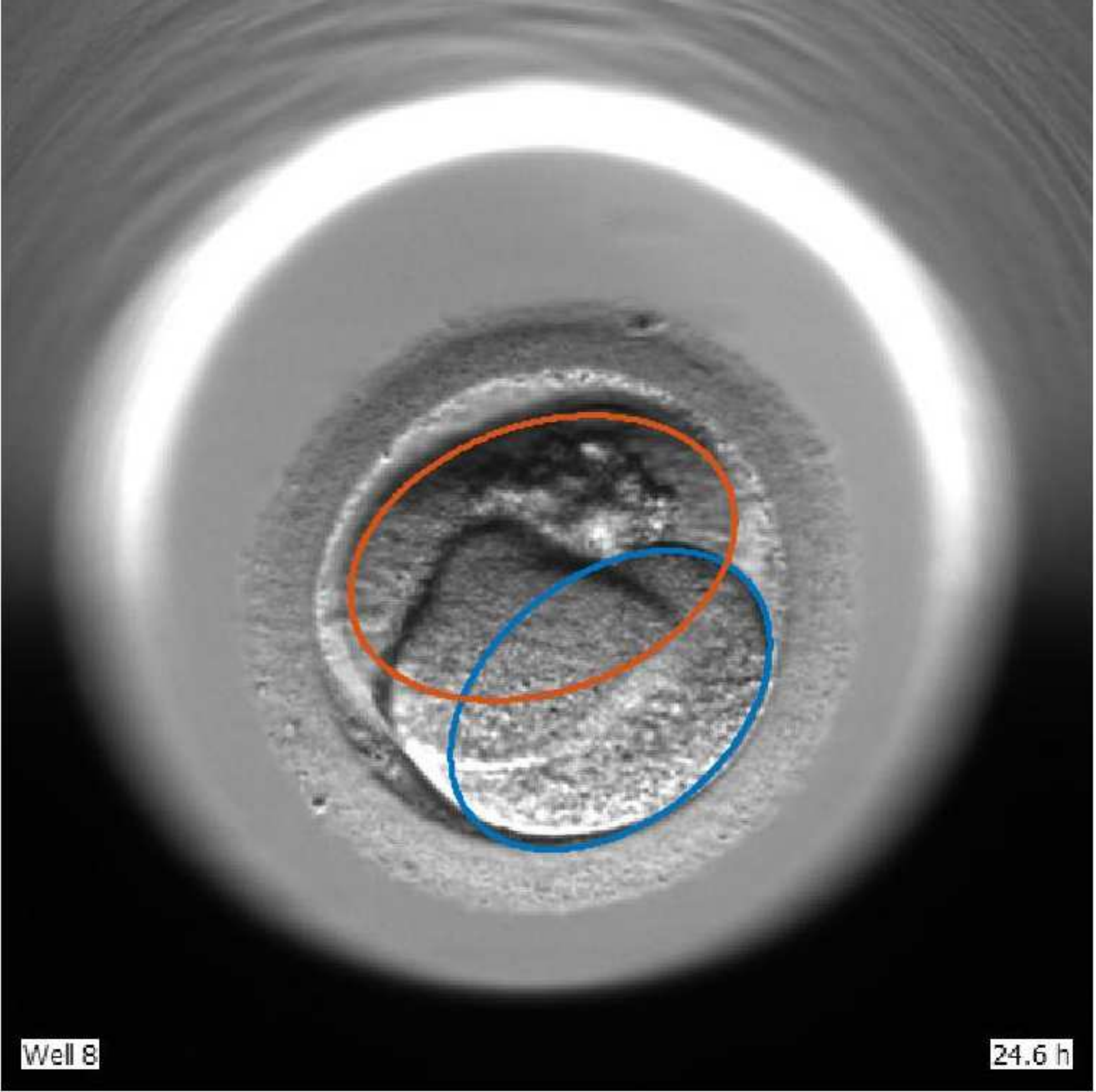}  & \includegraphics[trim=0cm 0cm 0cm 0cm, clip=true, width=.13\textwidth, height=20mm]{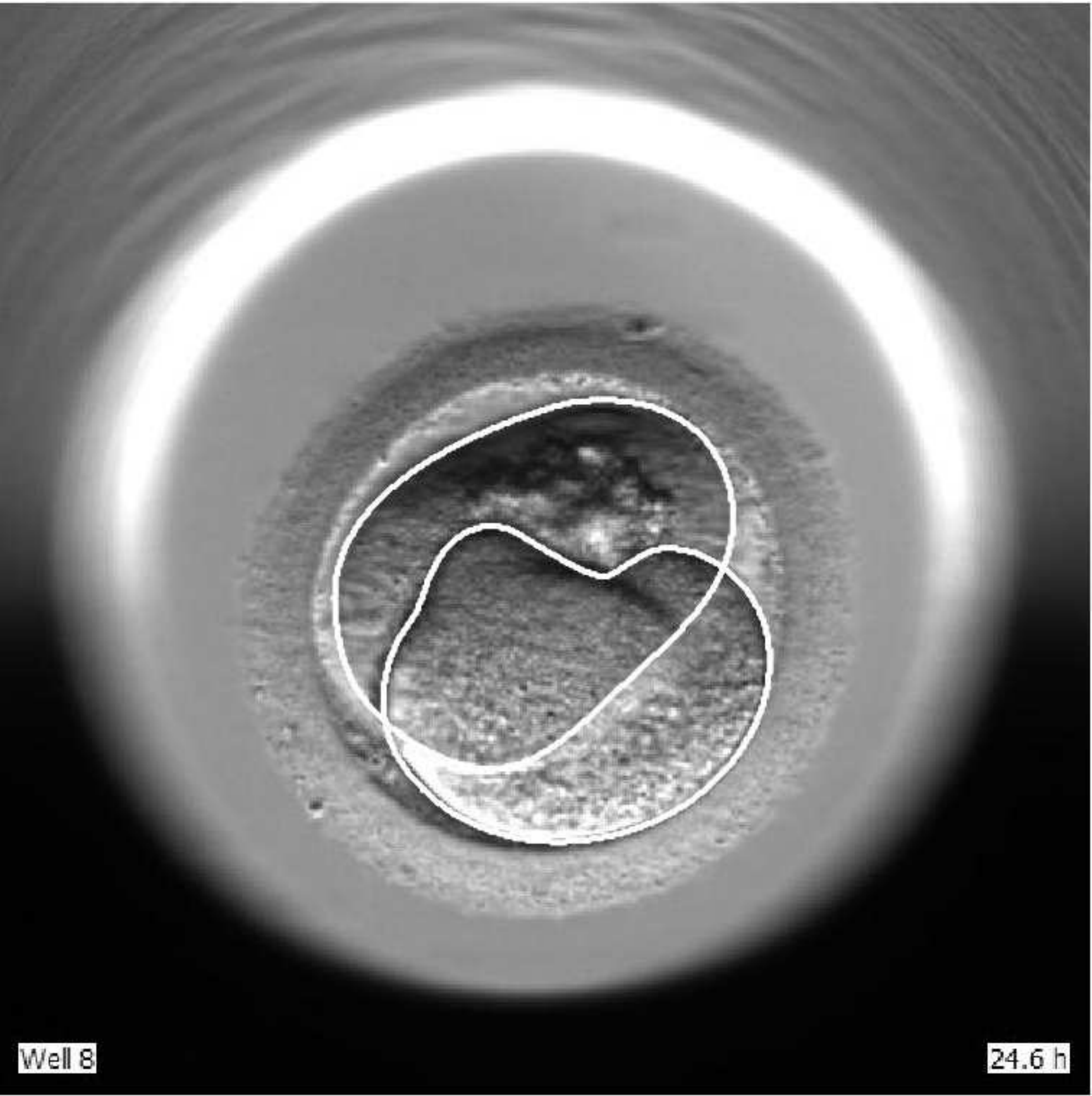}
&  \includegraphics[trim=0cm 0cm 0cm 0cm, clip=true, width=.13\textwidth, height=20mm]{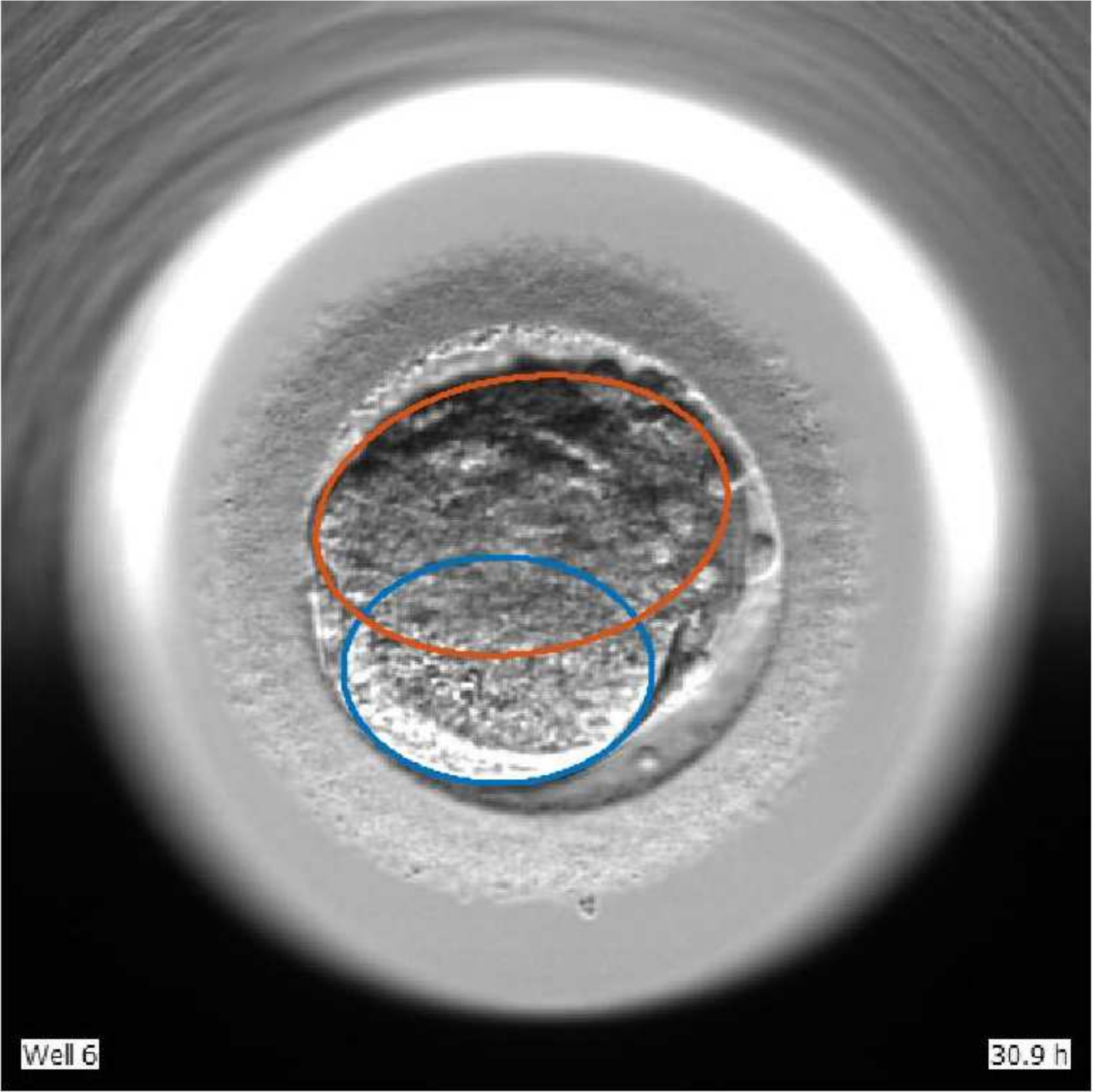} &  \includegraphics[trim=0cm 0cm 0cm 0cm, clip=true, width=.13\textwidth, height=20mm]{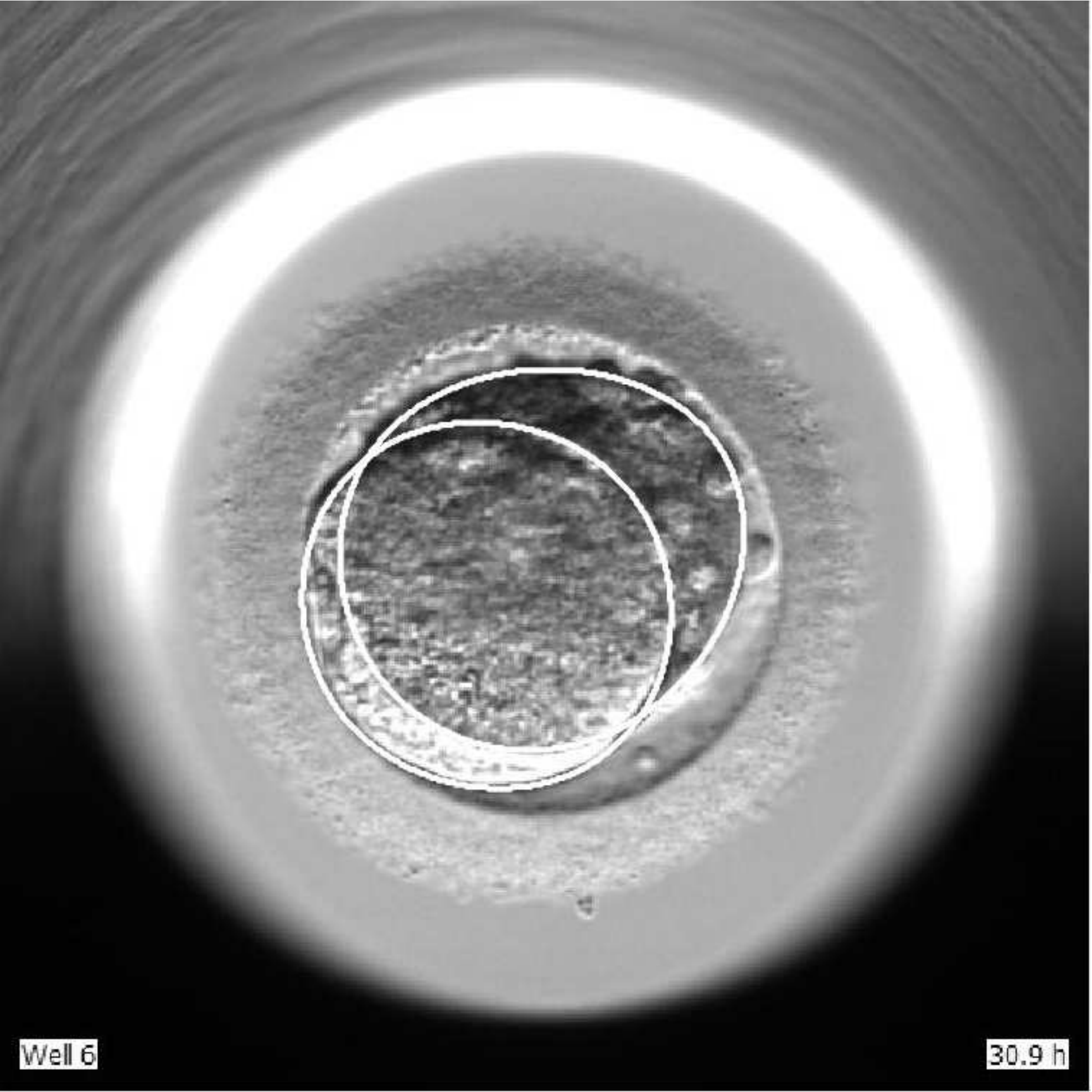} \\ \hline
{\rotatebox {90}{ 3-cell} } &  \includegraphics[trim=0cm 0cm 0cm 0cm, clip=true, width=.13\textwidth, height=20mm]{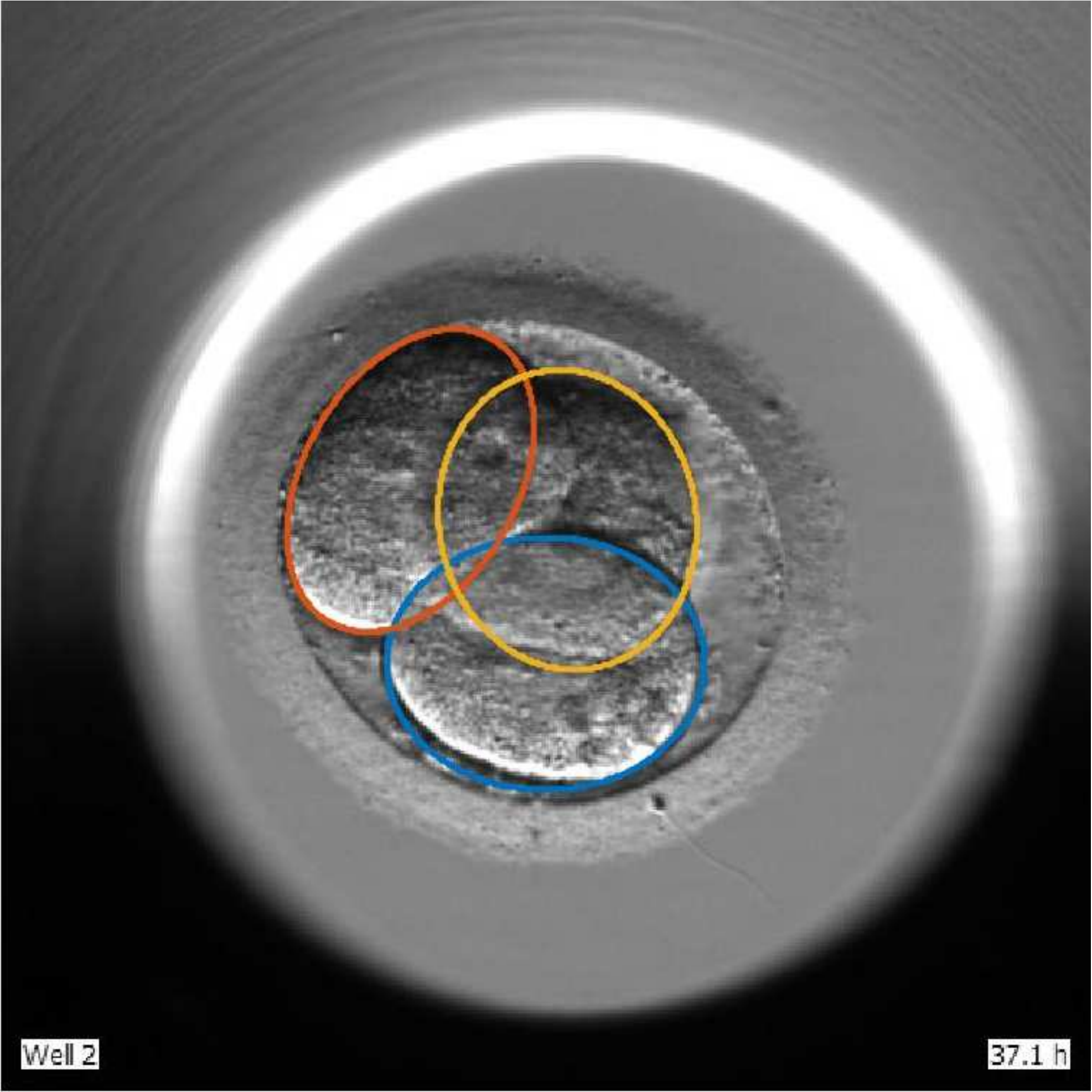} &  \includegraphics[trim=0cm 0cm 0cm 0cm, clip=true, width=.13\textwidth, height=20mm]{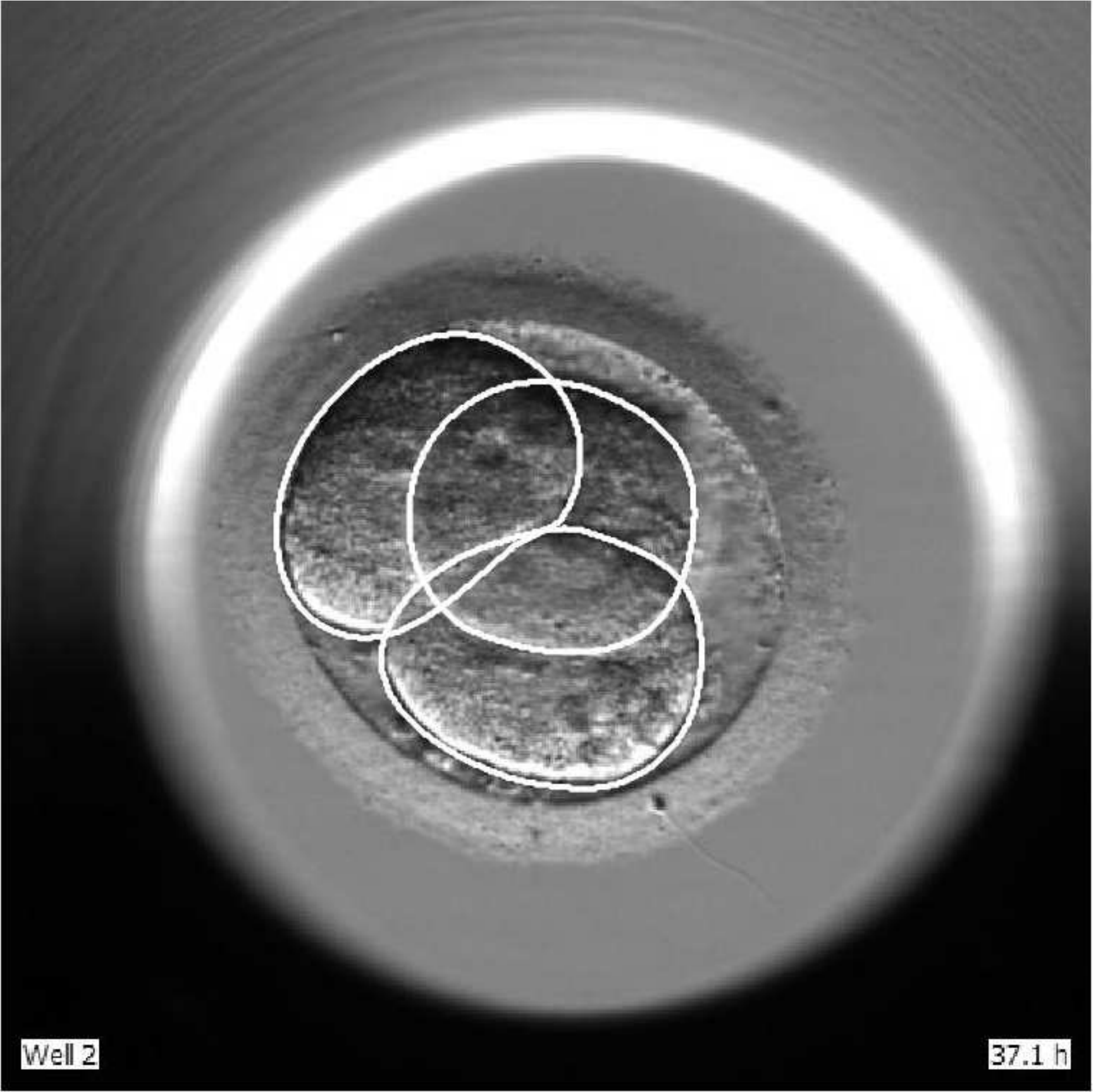} 
&  \includegraphics[trim=0cm 0cm 0cm 0cm, clip=true, width=.13\textwidth, height=20mm] {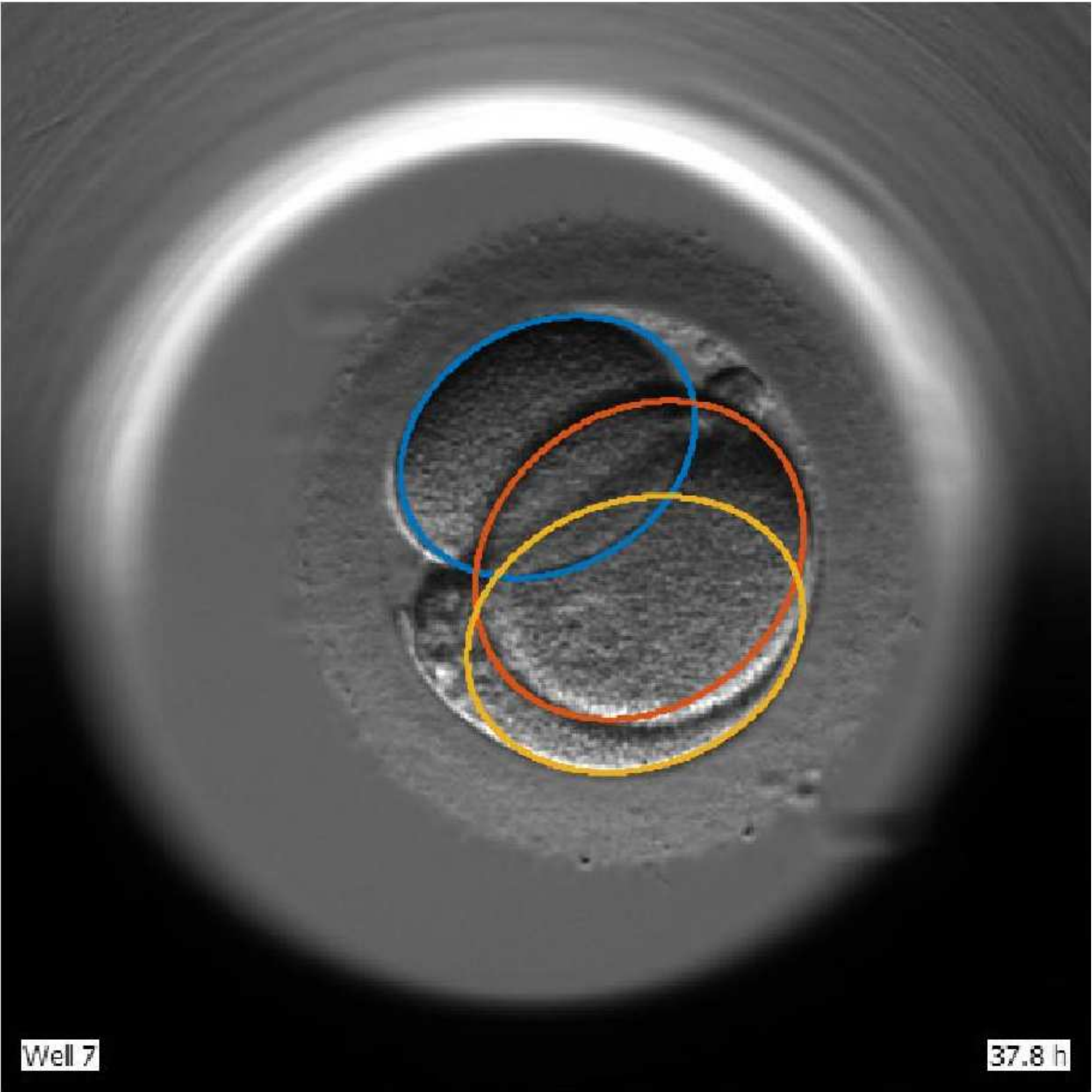}  & \includegraphics[trim=0cm 0cm 0cm 0cm, clip=true, width=.13\textwidth, height=20mm]{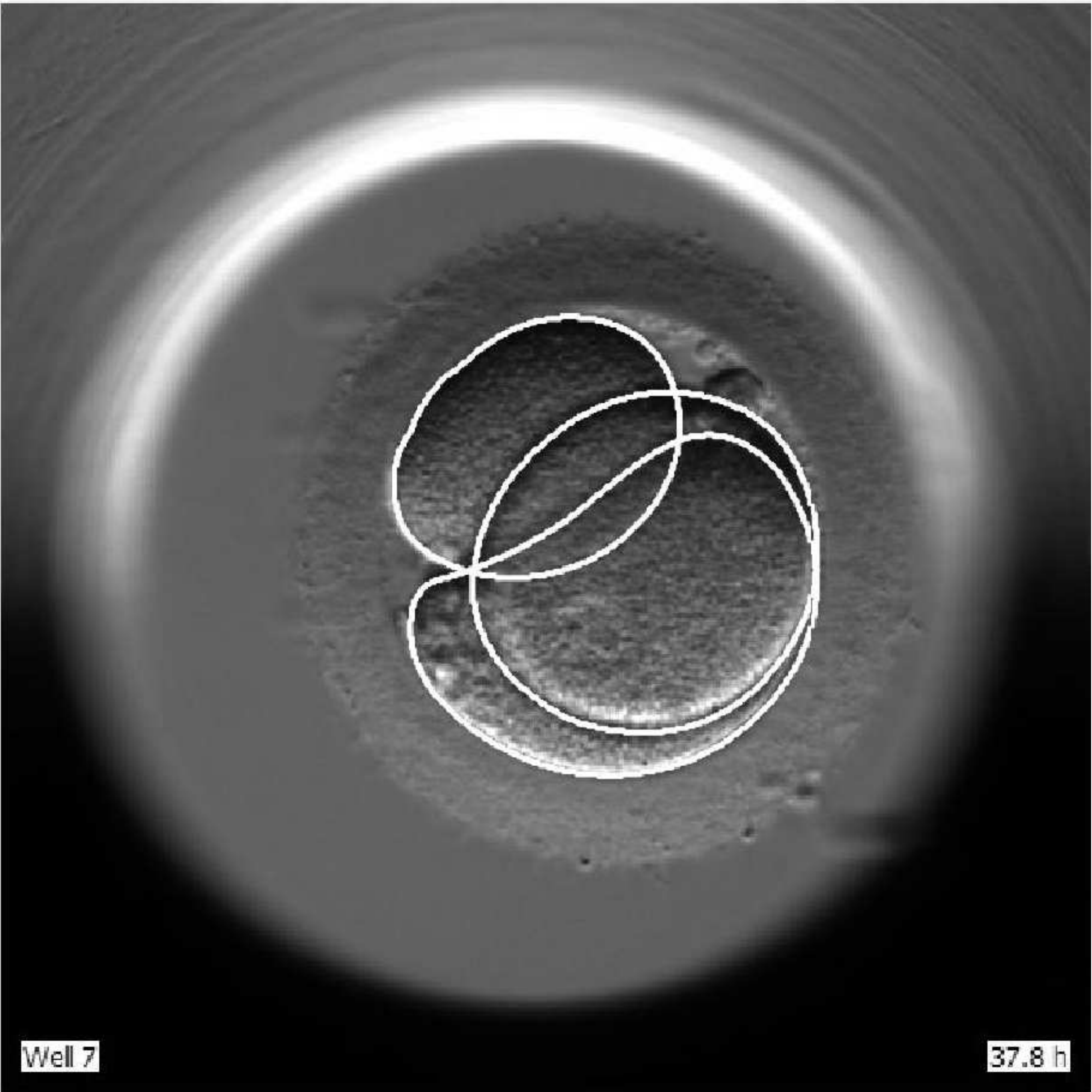}
&  \includegraphics[trim=0cm 0cm 0cm 0cm, clip=true, width=.13\textwidth, height=20mm]{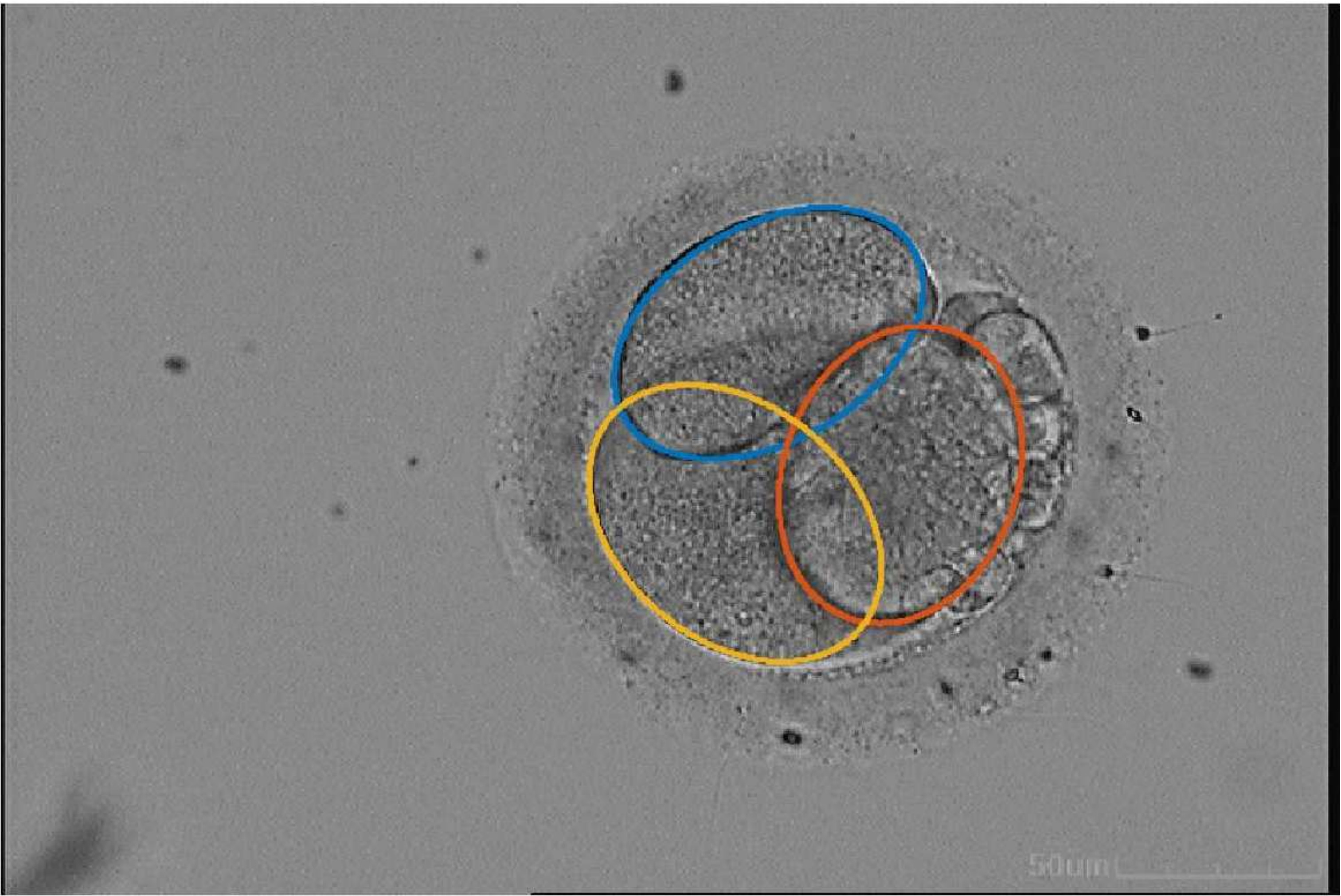} &  \includegraphics[trim=0cm 0cm 0cm 0cm, clip=true, width=.13\textwidth, height=20mm]{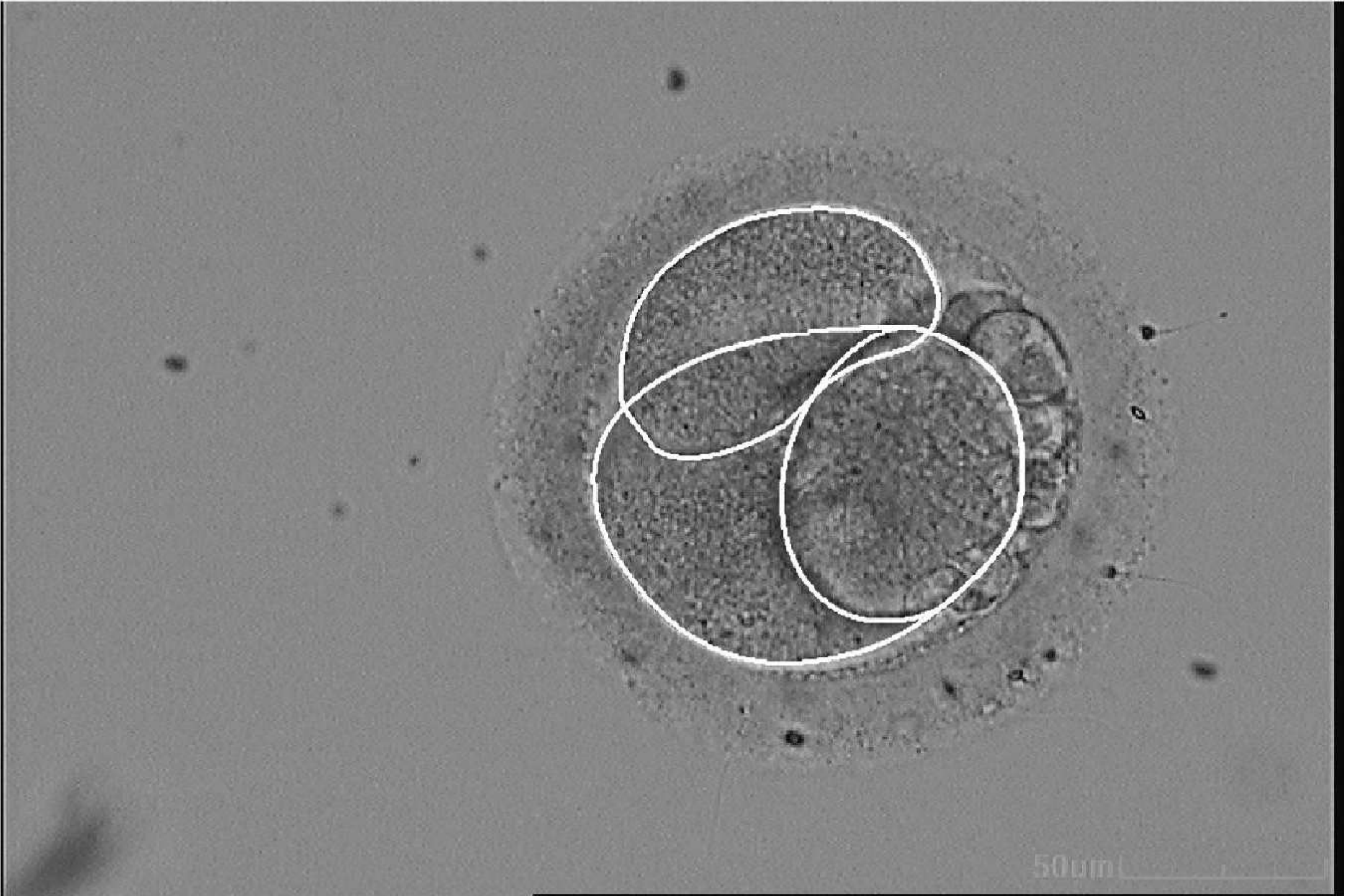} \\ \hline
{\rotatebox {90}{ 4-cell} } &  \includegraphics[trim=0cm 0cm 0cm 0cm, clip=true, width=.13\textwidth, height=20mm]{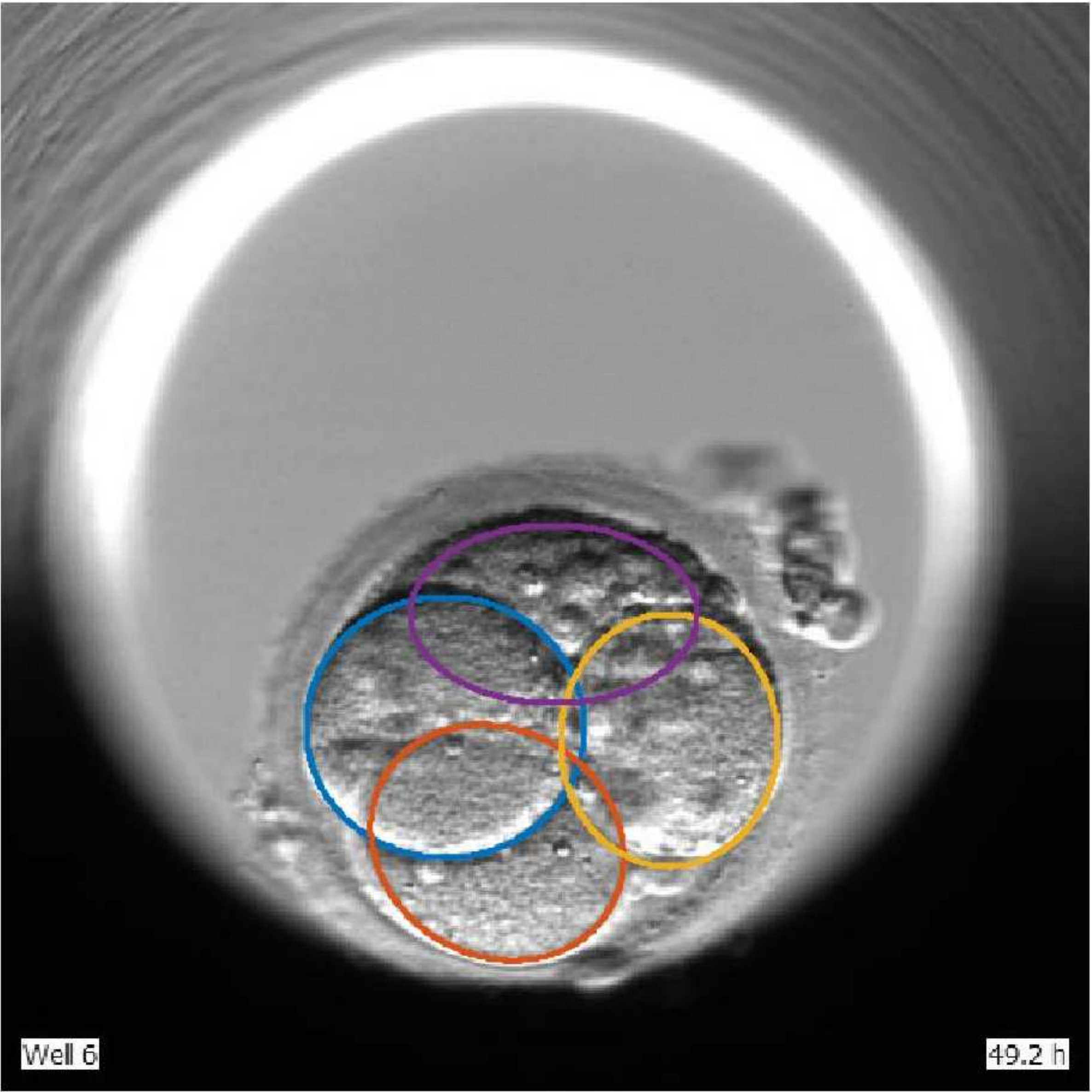} &  \includegraphics[trim=0cm 0cm 0cm 0cm, clip=true, width=.13\textwidth, height=20mm]{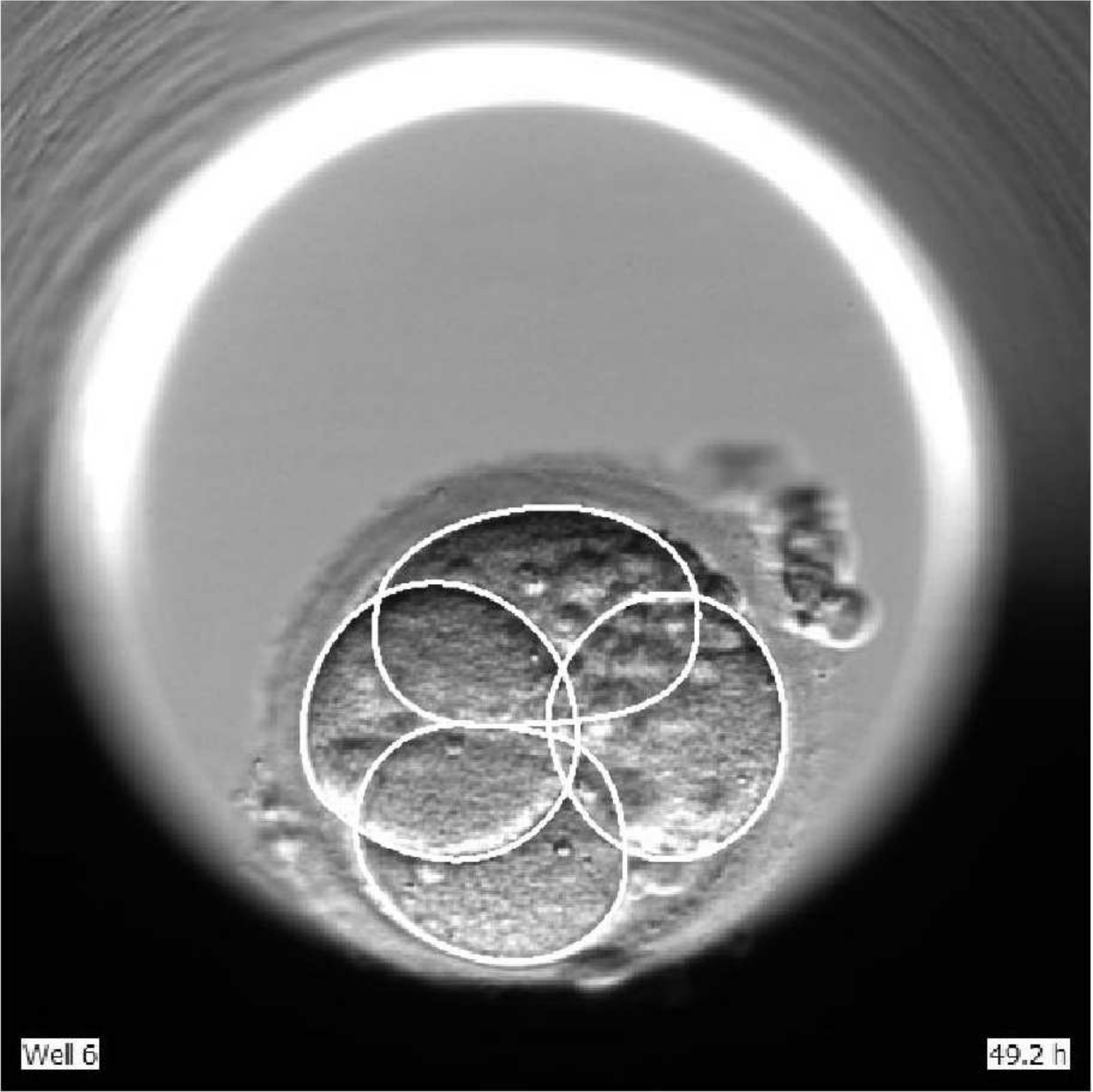} 
&  \includegraphics[trim=0cm 0cm 0cm 0cm, clip=true, width=.13\textwidth, height=20mm] {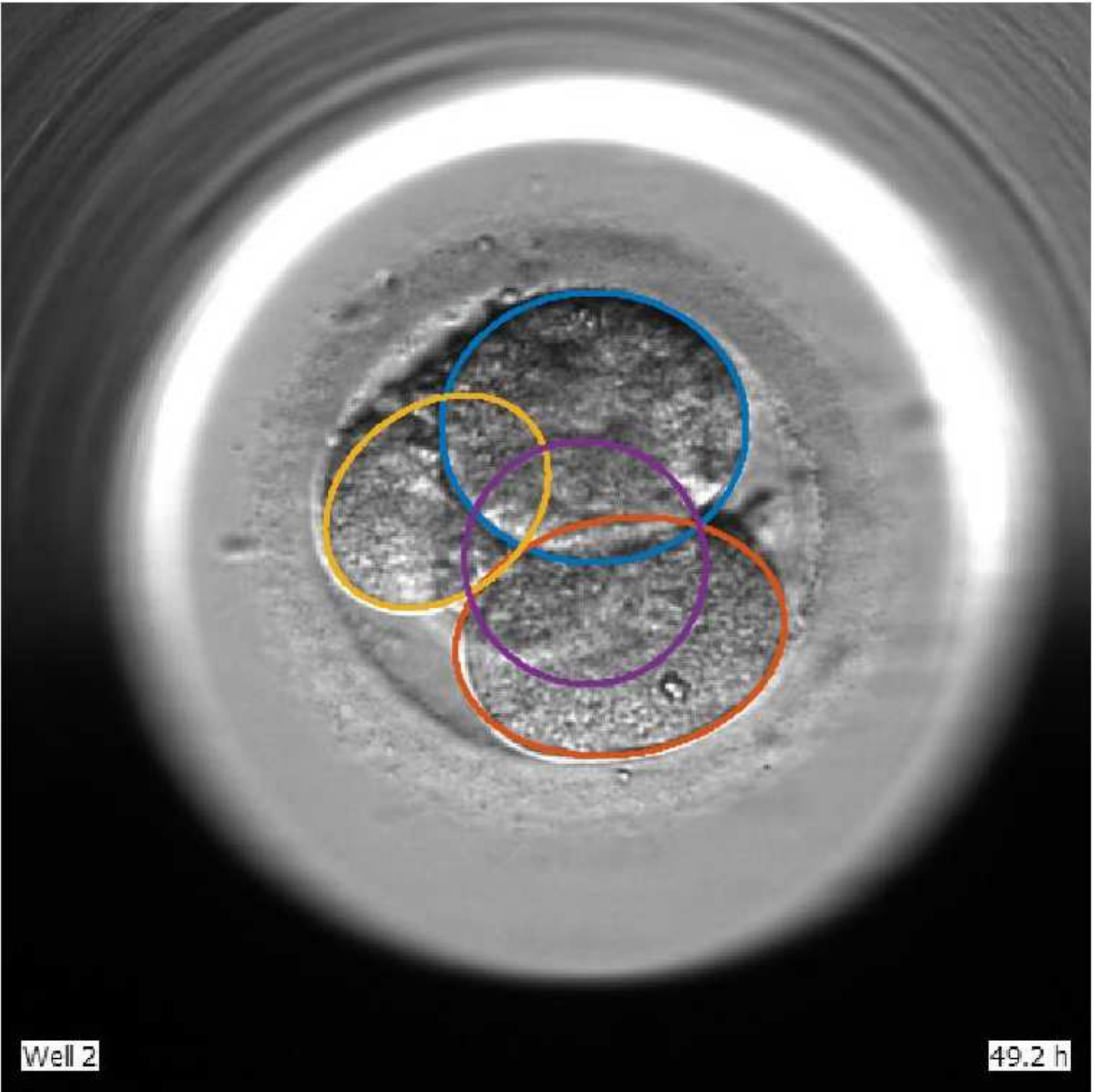}  & \includegraphics[trim=0cm 0cm 0cm 0cm, clip=true, width=.13\textwidth, height=20mm]{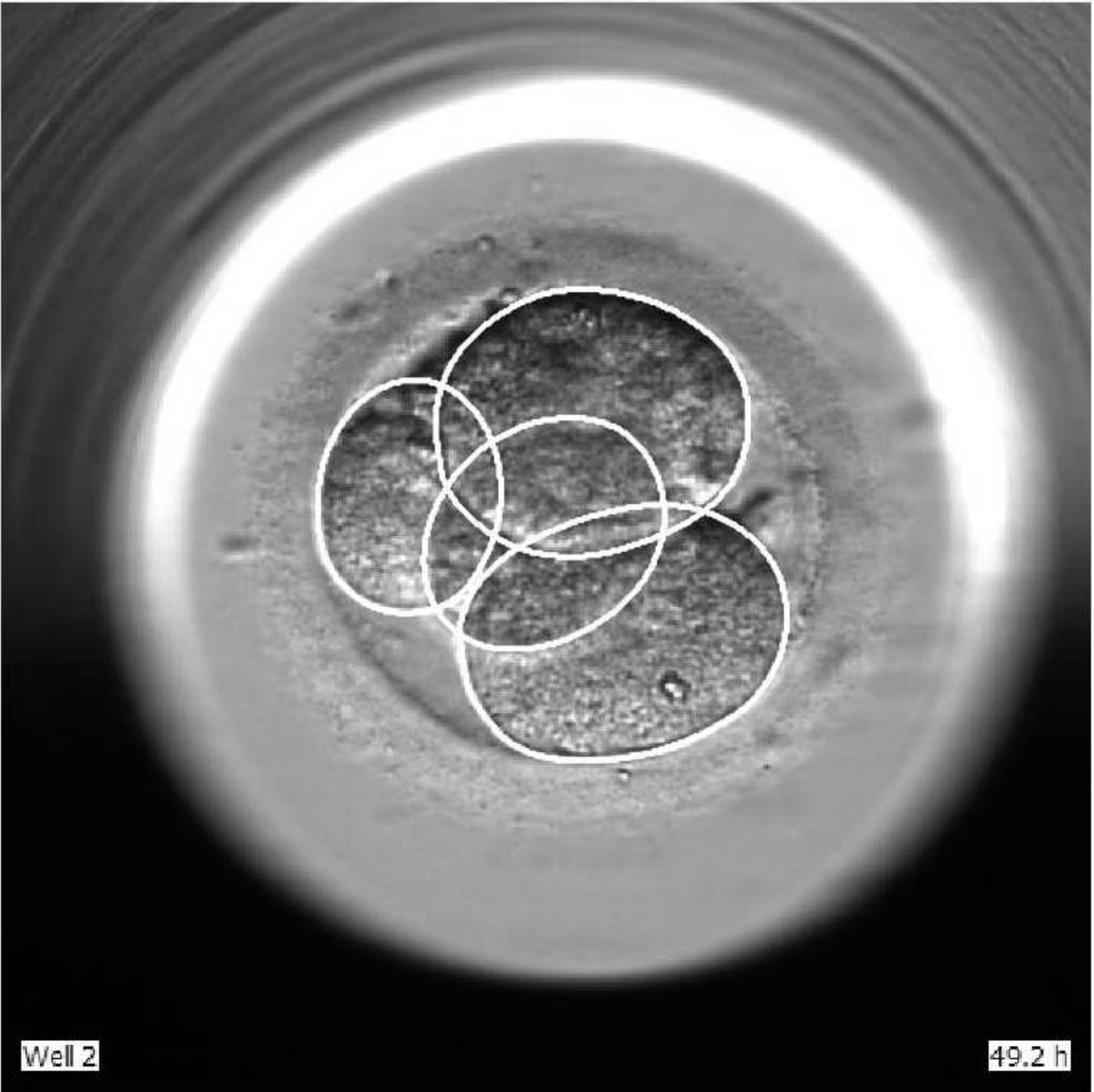}
&  \includegraphics[trim=0cm 0cm 0cm 0cm, clip=true, width=.13\textwidth, height=20mm]{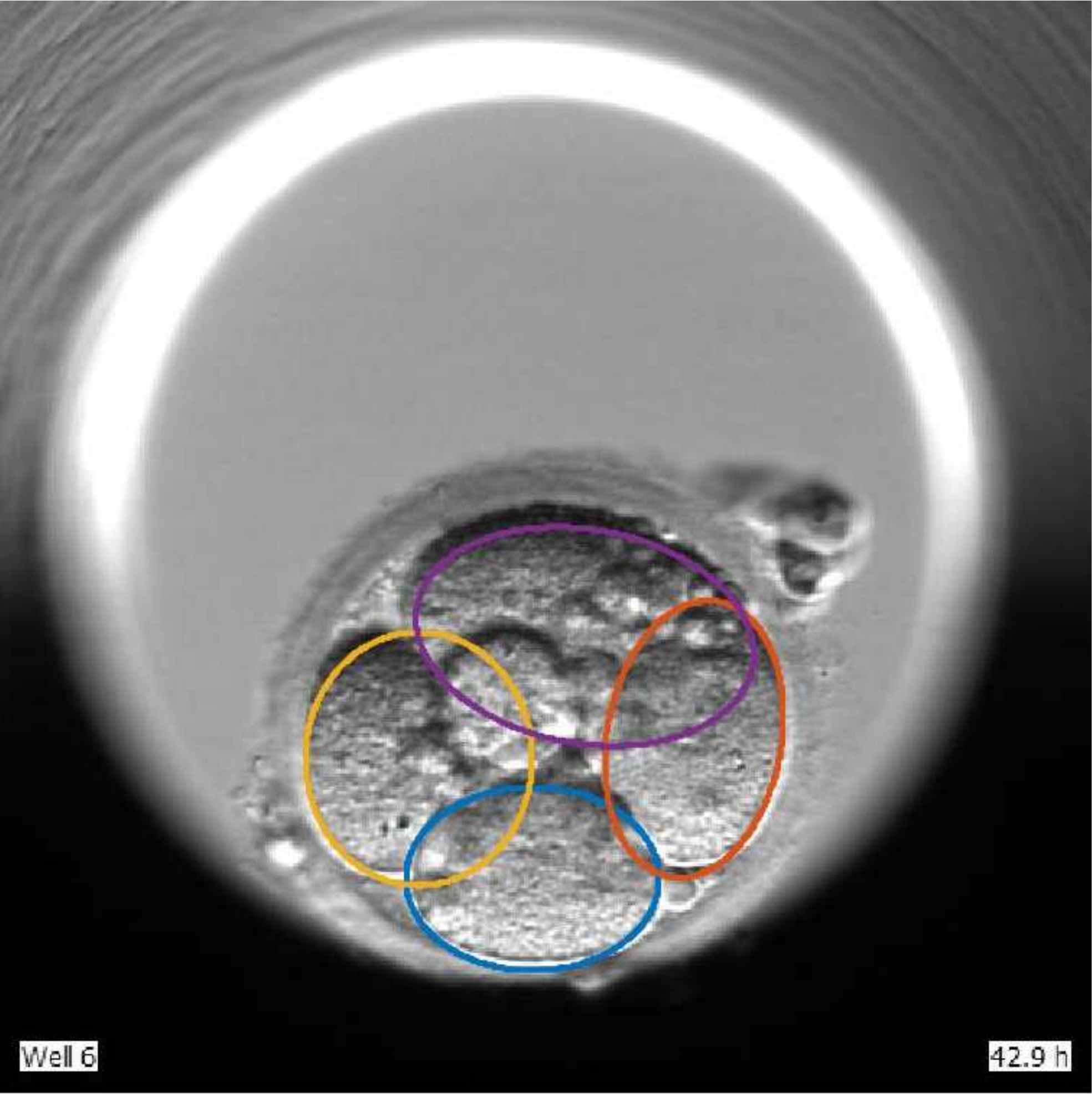} &  \includegraphics[trim=0cm 0cm 0cm 0cm, clip=true, width=.13\textwidth, height=20mm]{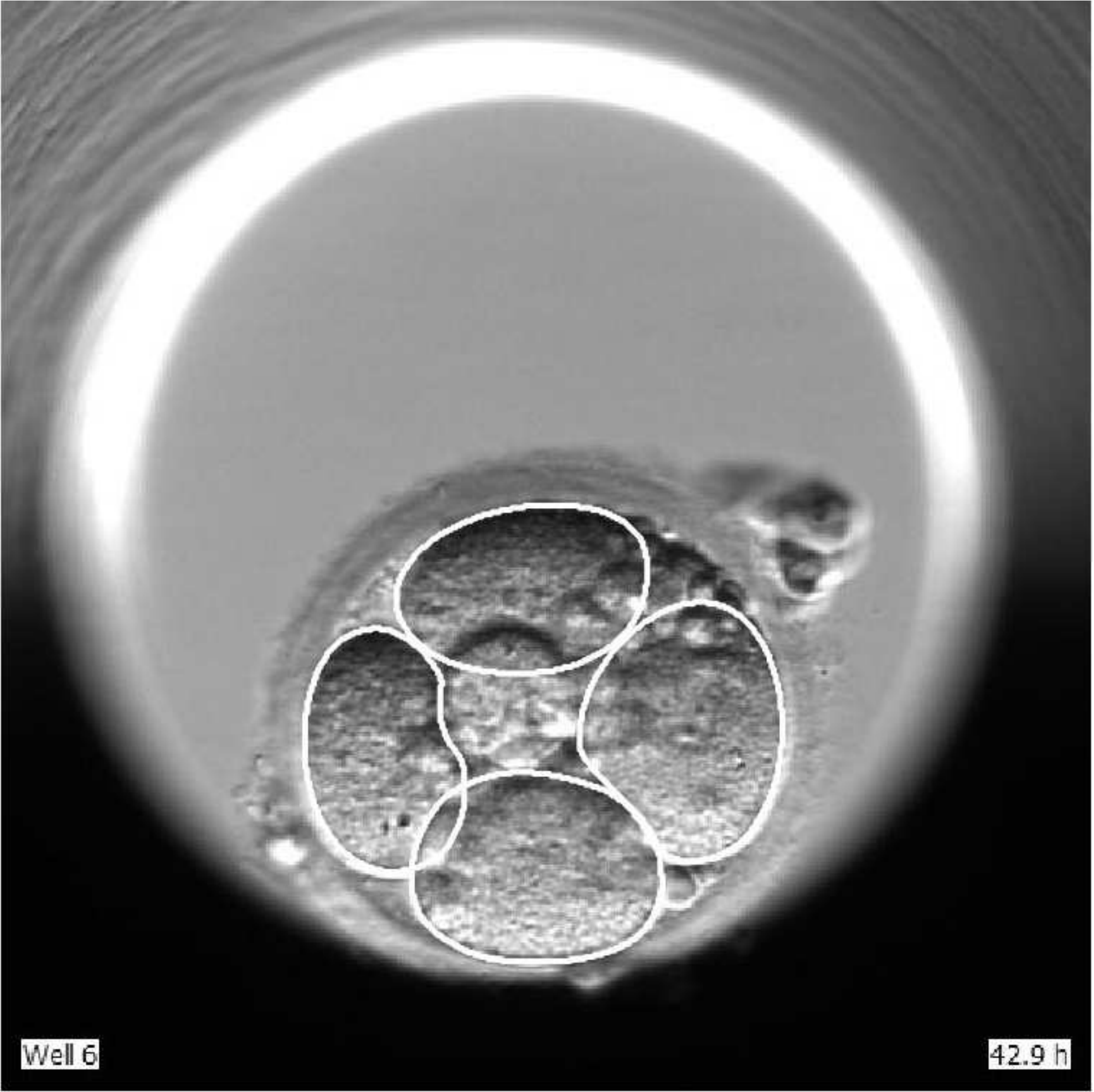} \\ \hline
{\rotatebox {90}{ 5-cell} } &  \includegraphics[trim=0cm 0cm 0cm 0cm, clip=true, width=.13\textwidth, height=20mm]{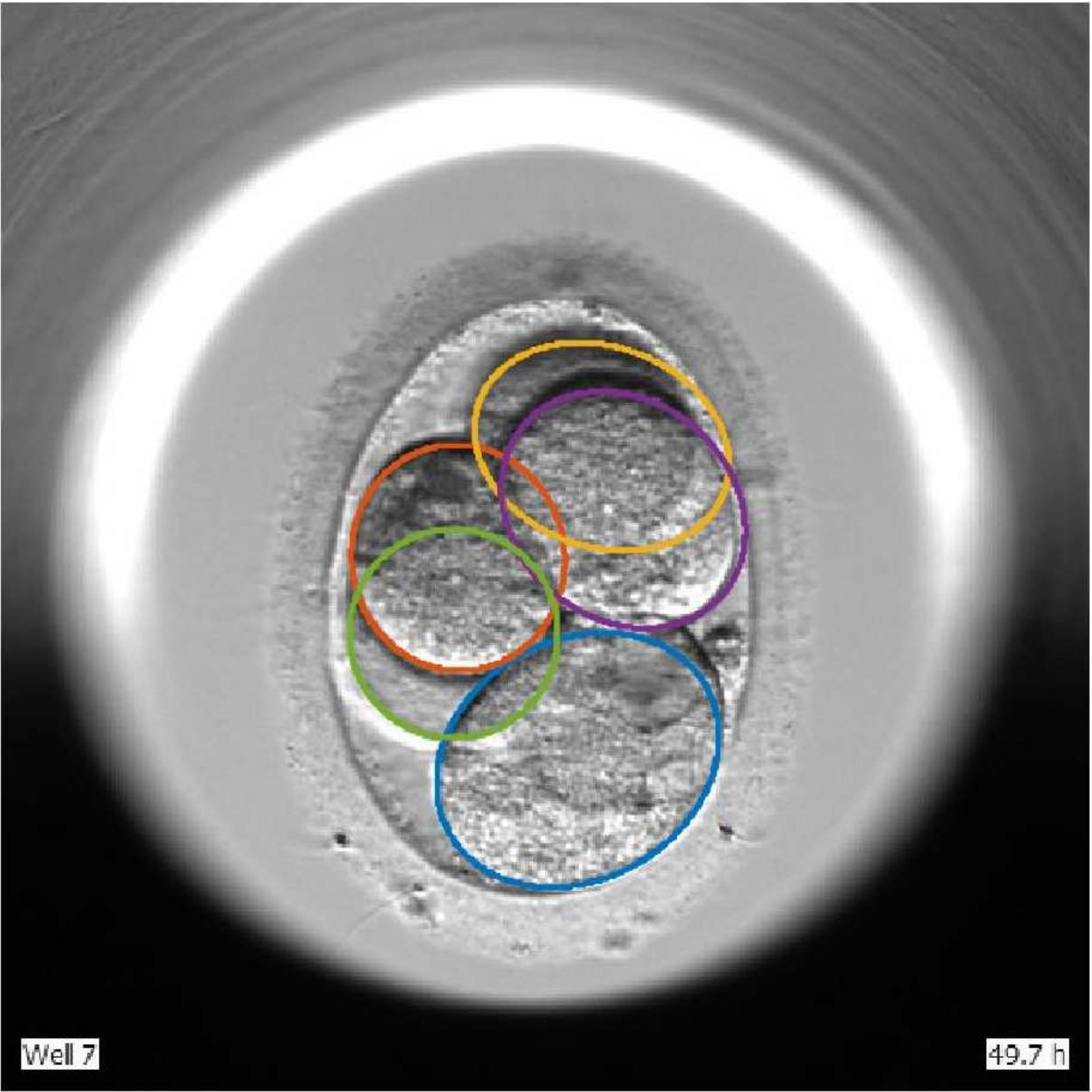} &  \includegraphics[trim=0cm 0cm 0cm 0cm, clip=true, width=.13\textwidth, height=20mm]{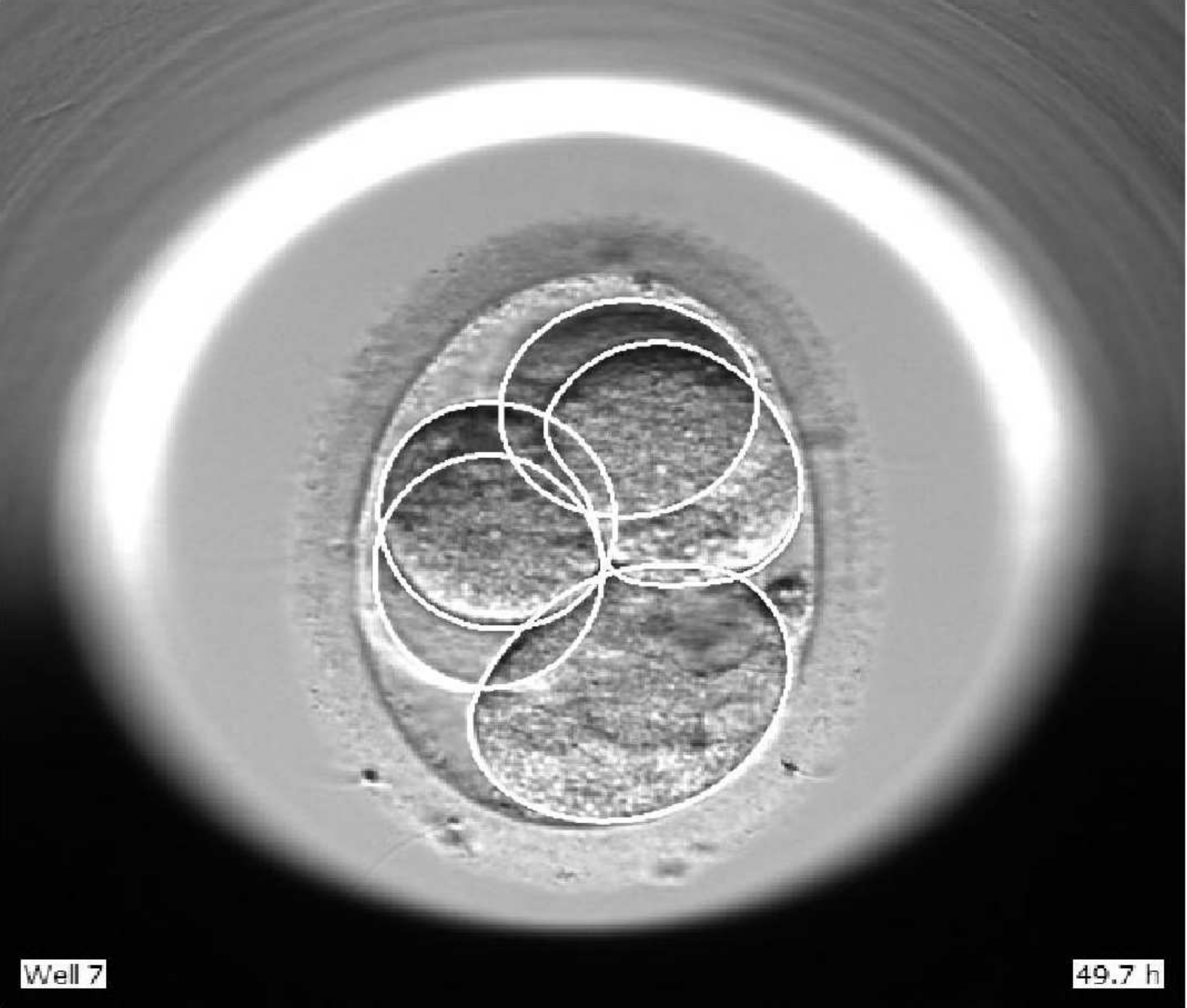} 
&  \includegraphics[trim=0cm 0cm 0cm 0cm, clip=true, width=.13\textwidth, height=20mm] {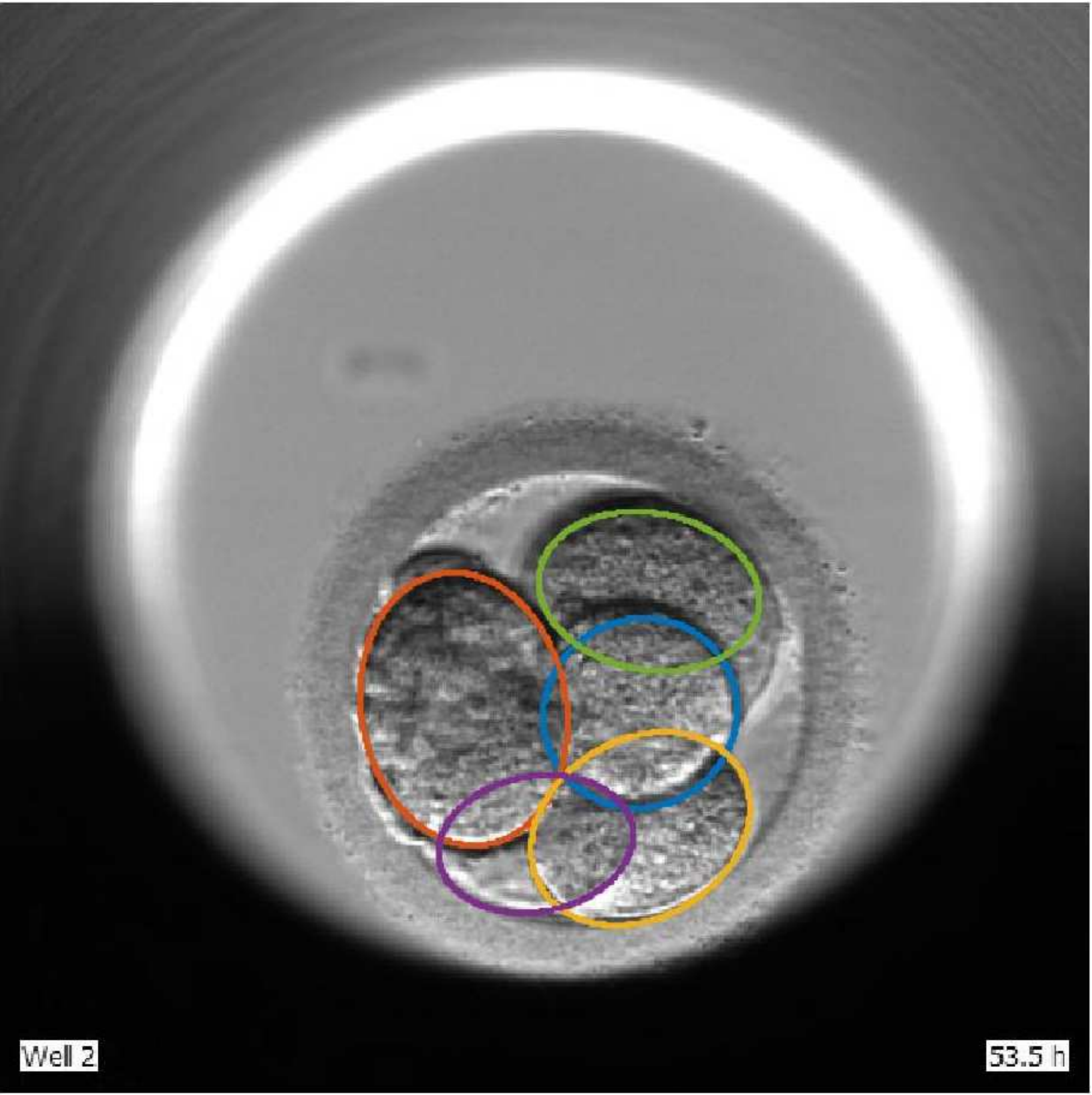}  & \includegraphics[trim=0cm 0cm 0cm 0cm, clip=true, width=.13\textwidth, height=20mm]{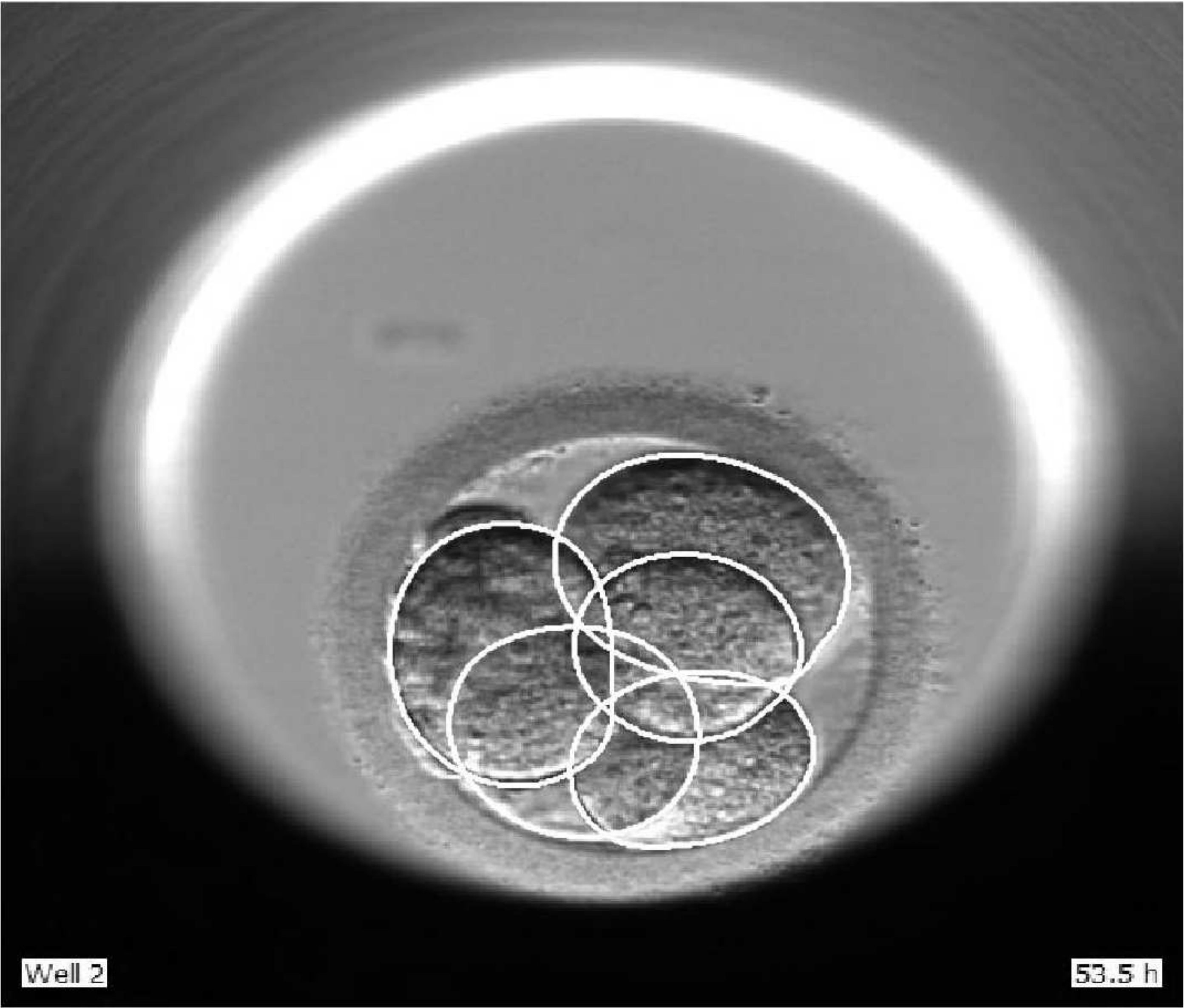}
&  \includegraphics[trim=0cm 0cm 0cm 0cm, clip=true, width=.13\textwidth, height=20mm]{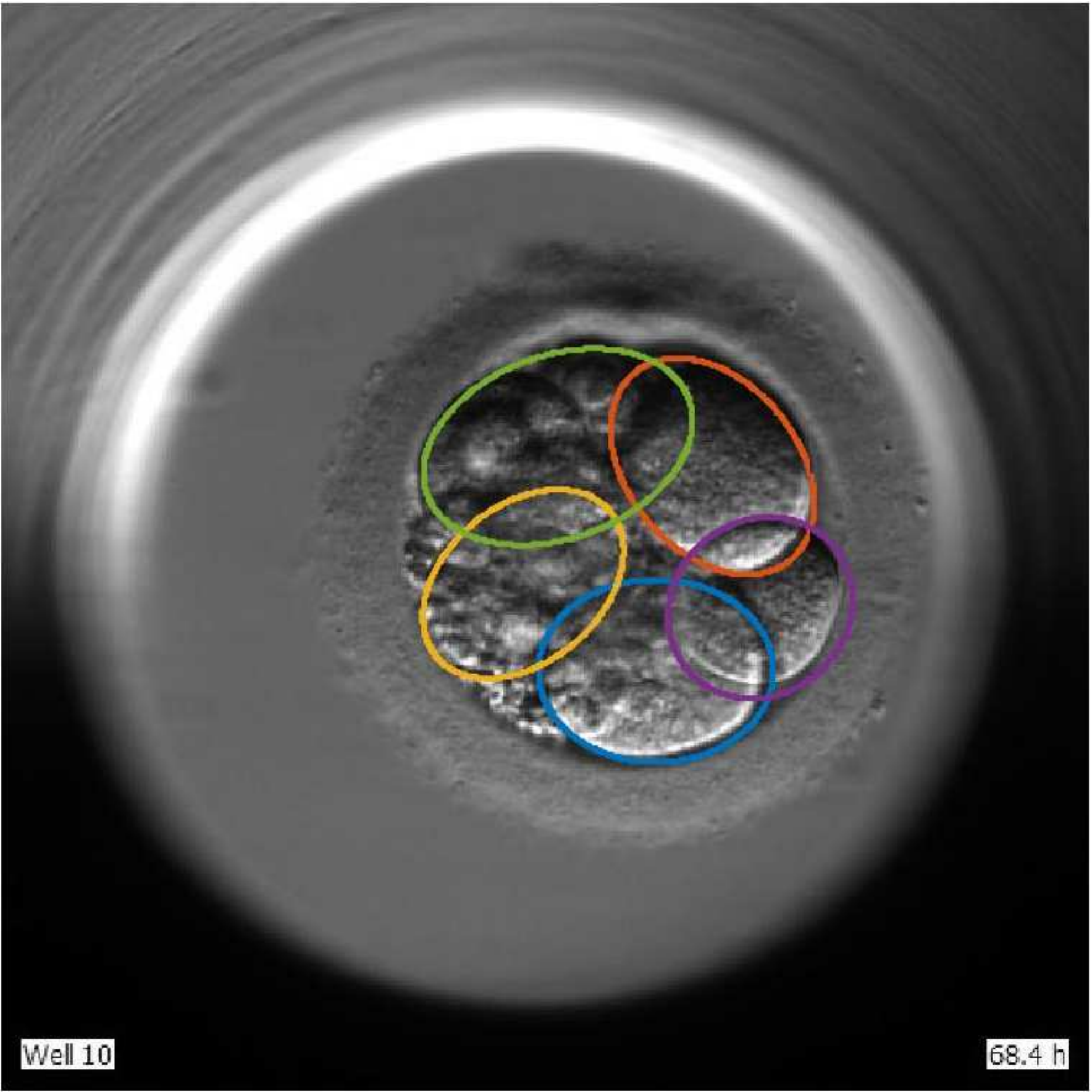} &  \includegraphics[trim=0cm 0cm 0cm 0cm, clip=true, width=.13\textwidth, height=20mm]{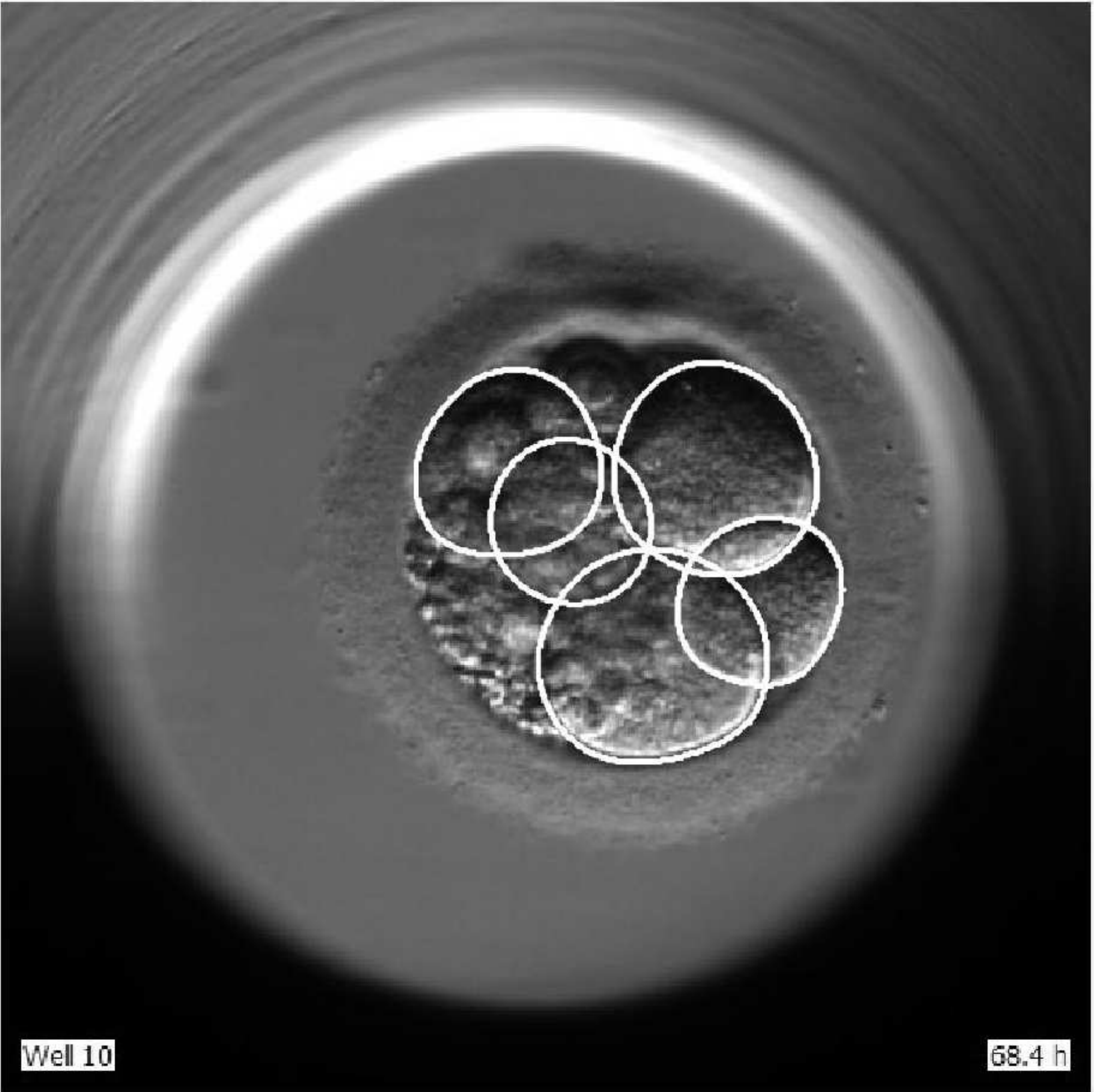} \\ \hline
{\rotatebox {90}{ 6-cell} } &  \includegraphics[trim=0cm 0cm 0cm 0cm, clip=true, width=.13\textwidth, height=20mm]{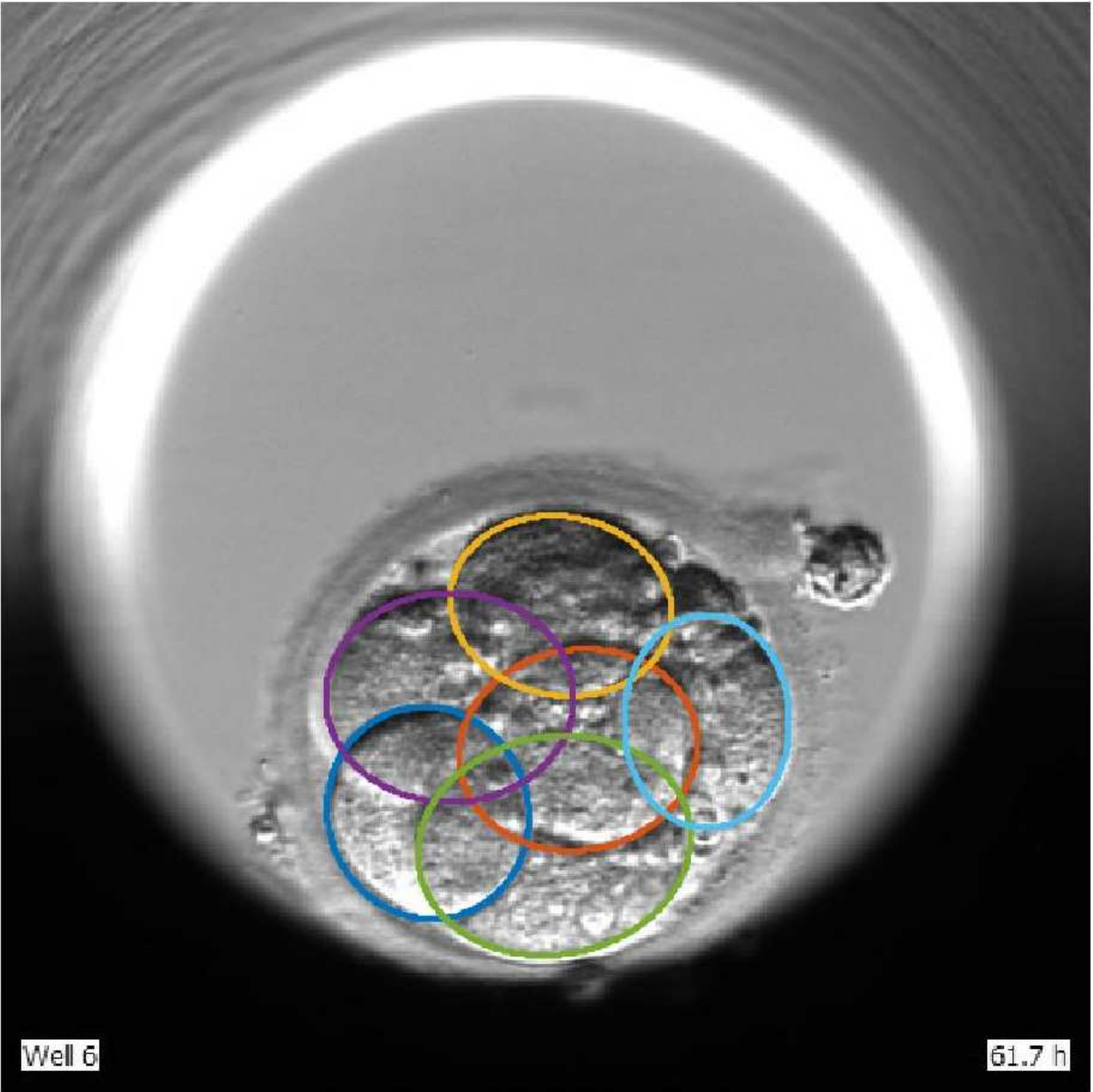} &  \includegraphics[trim=0cm 0cm 0cm 0cm, clip=true, width=.13\textwidth, height=20mm]{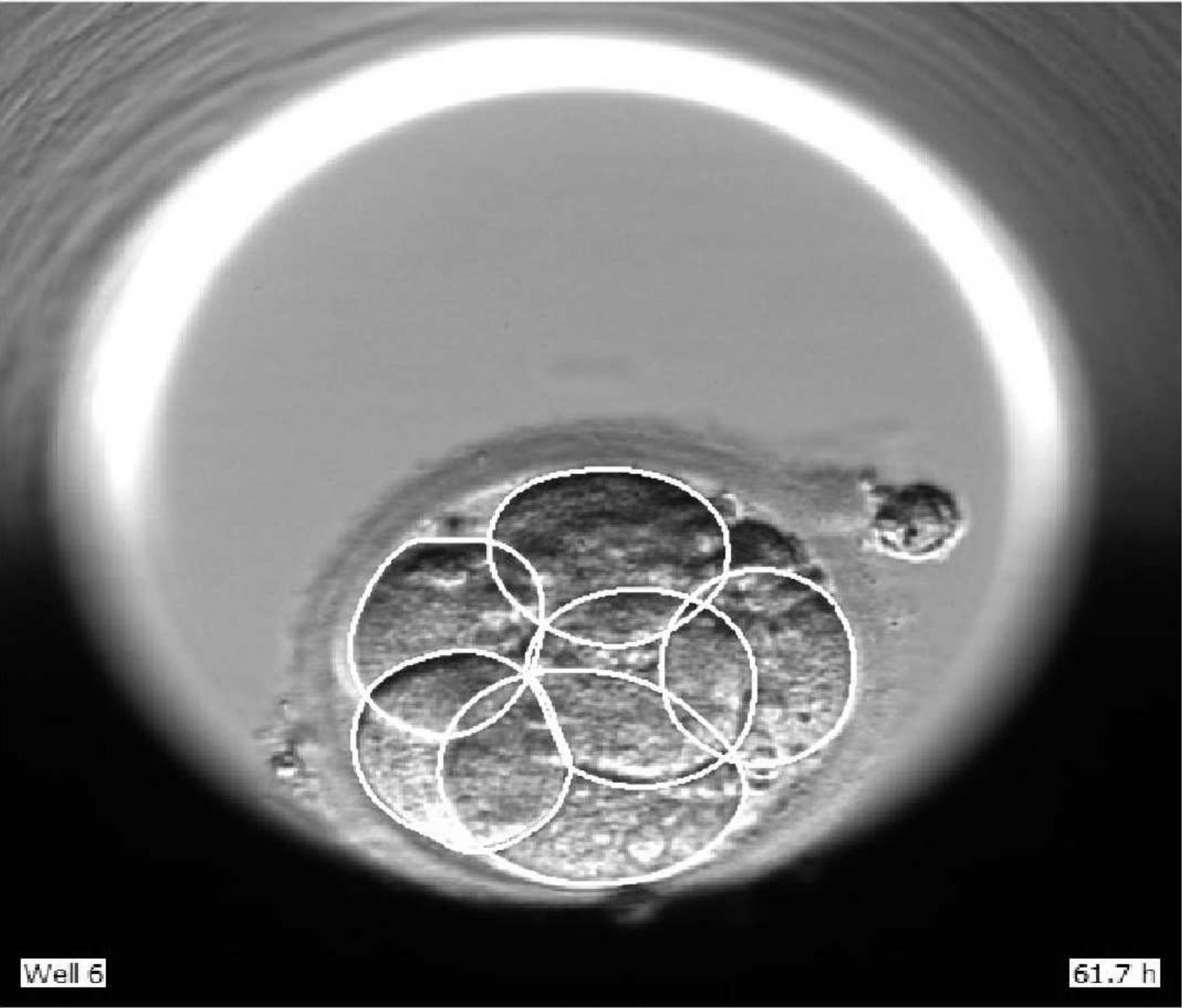} 
&  \includegraphics[trim=0cm 0cm 0cm 0cm, clip=true, width=.13\textwidth, height=20mm] {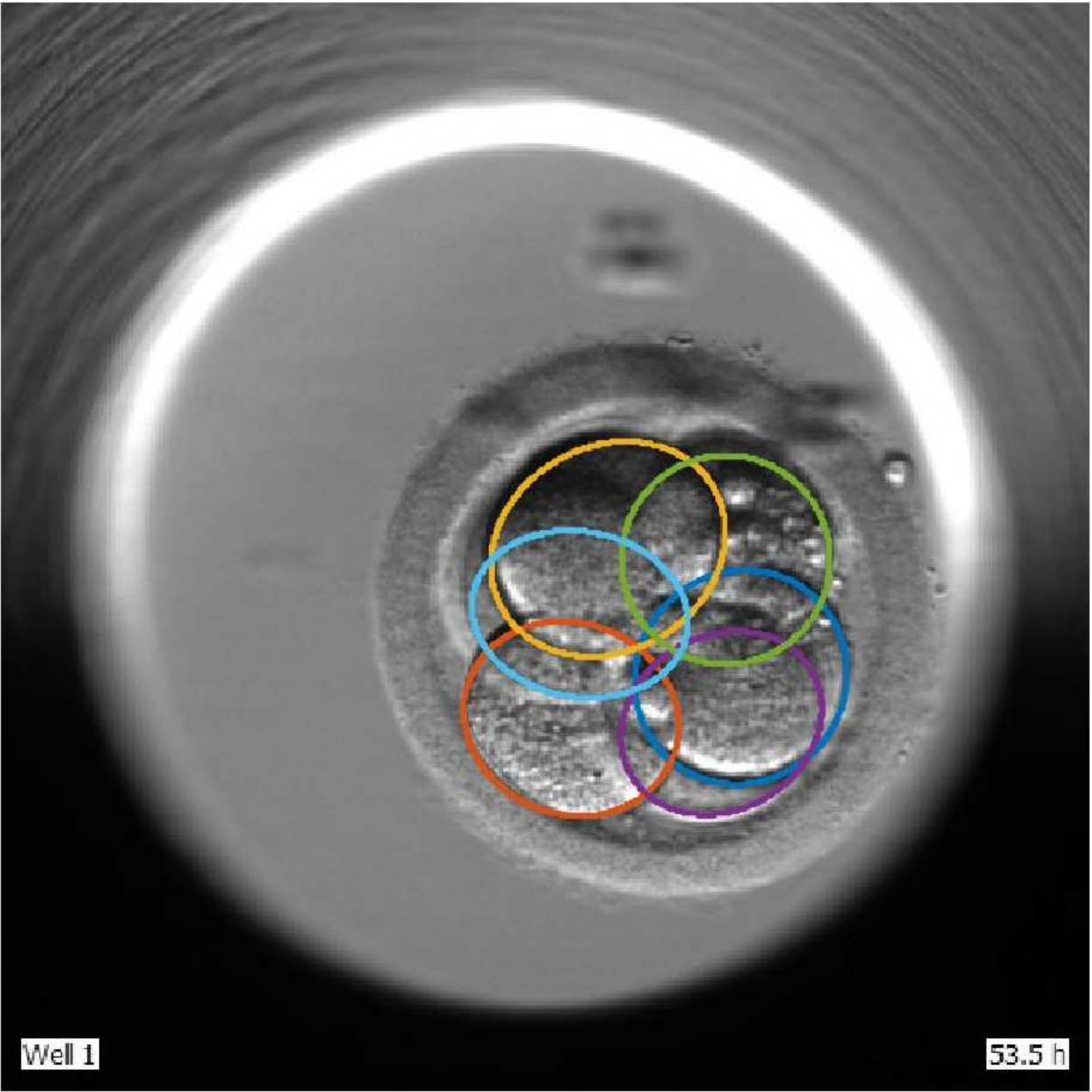}  & \includegraphics[trim=0cm 0cm 0cm 0cm, clip=true, width=.13\textwidth, height=20mm]{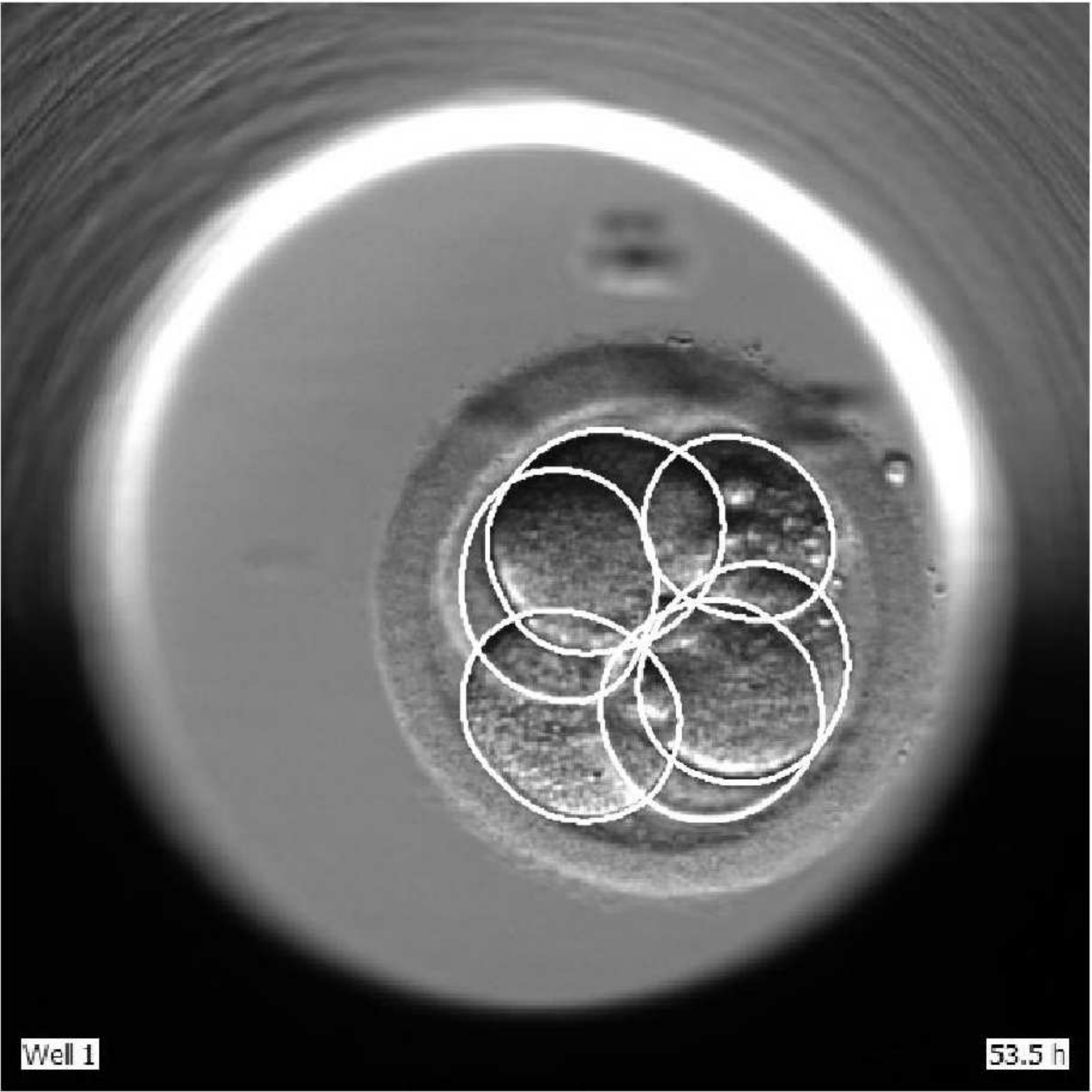}
&  \includegraphics[trim=0cm 0cm 0cm 0cm, clip=true, width=.13\textwidth, height=20mm]{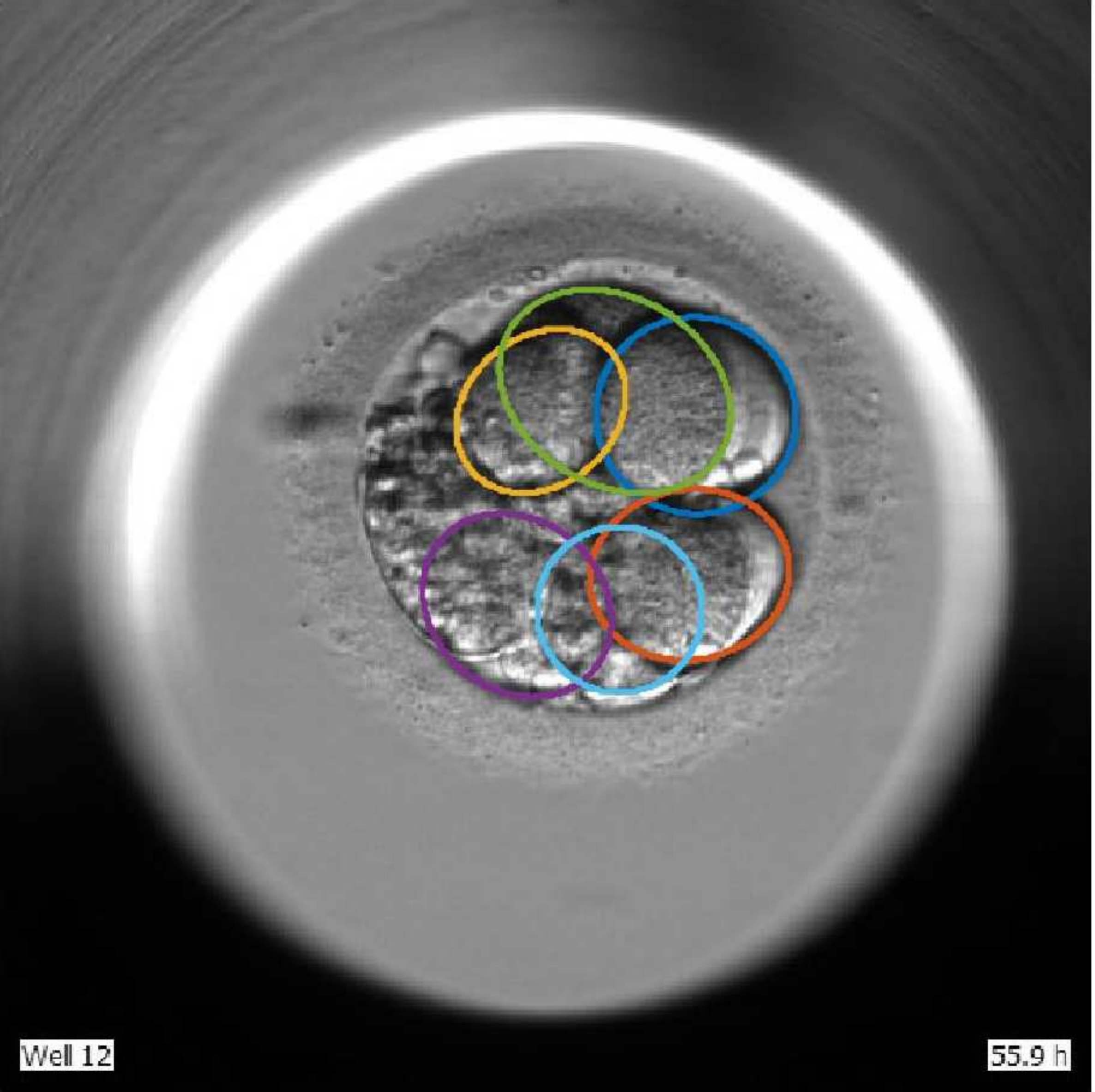} &  \includegraphics[trim=0cm 0cm 0cm 0cm, clip=true, width=.13\textwidth, height=20mm]{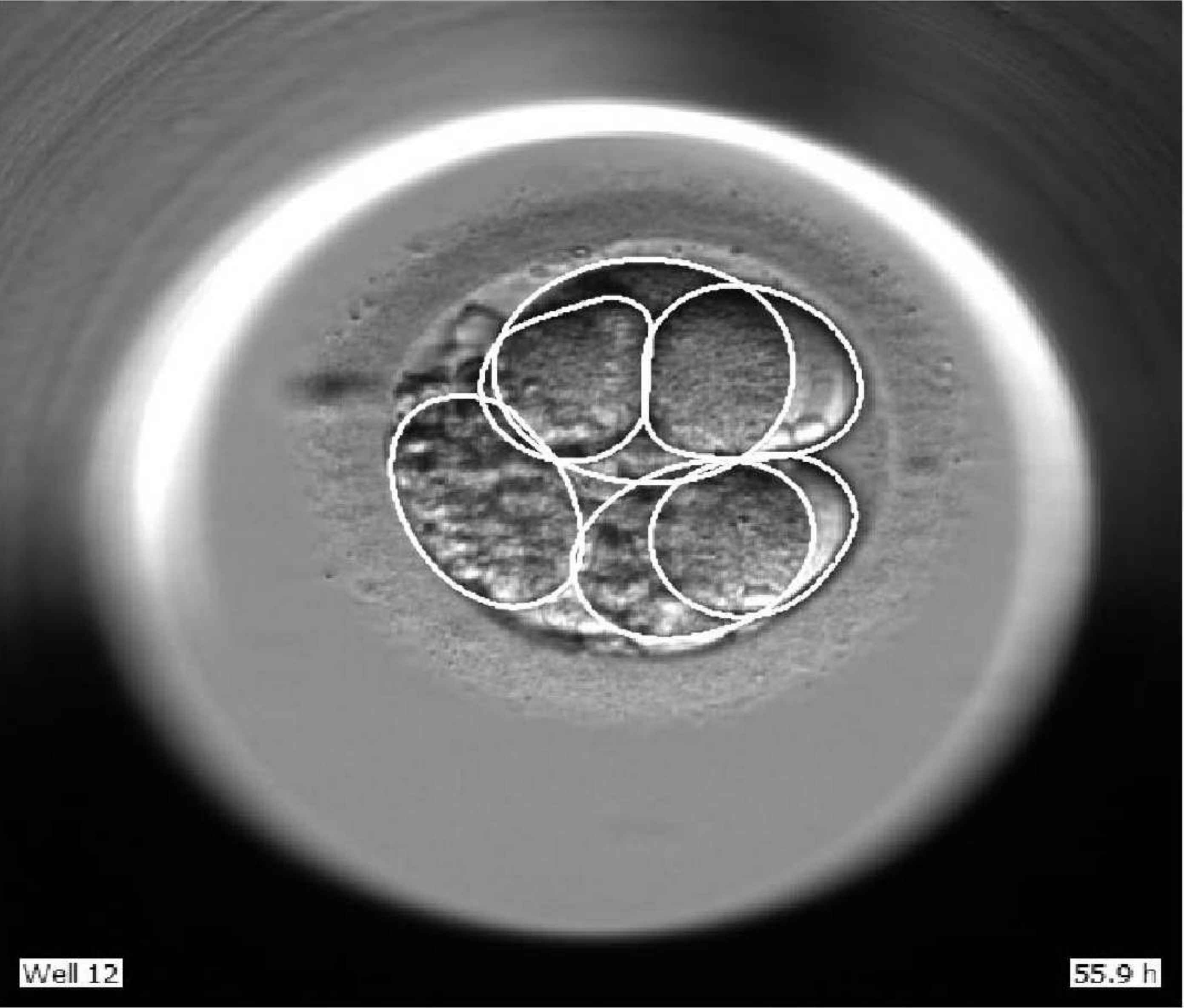} \\ \hline
{\rotatebox {90}{ 7-cell} } &  \includegraphics[trim=0cm 0cm 0cm 0cm, clip=true, width=.13\textwidth, height=20mm]{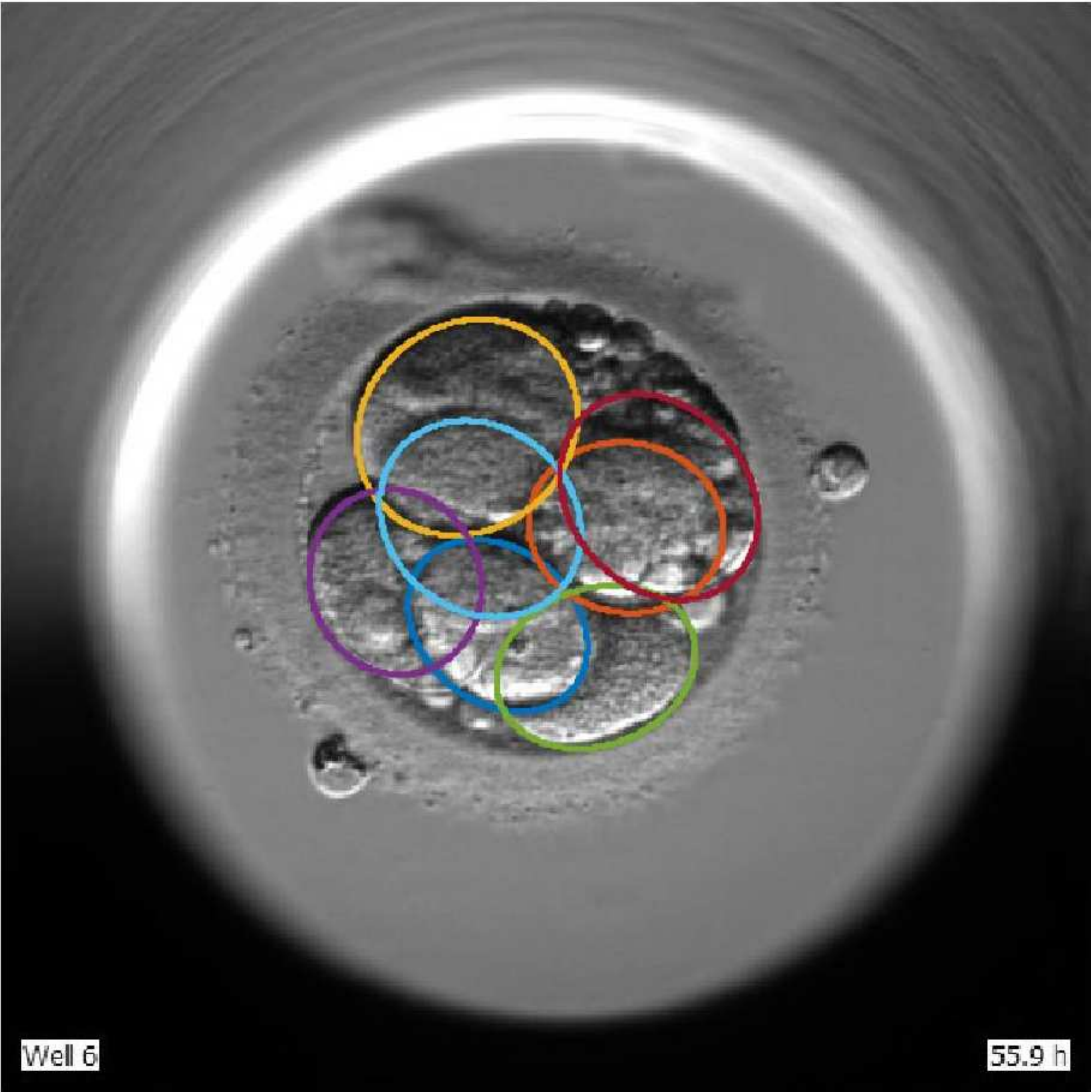} &  \includegraphics[trim=0cm 0cm 0cm 0cm, clip=true, width=.13\textwidth, height=20mm]{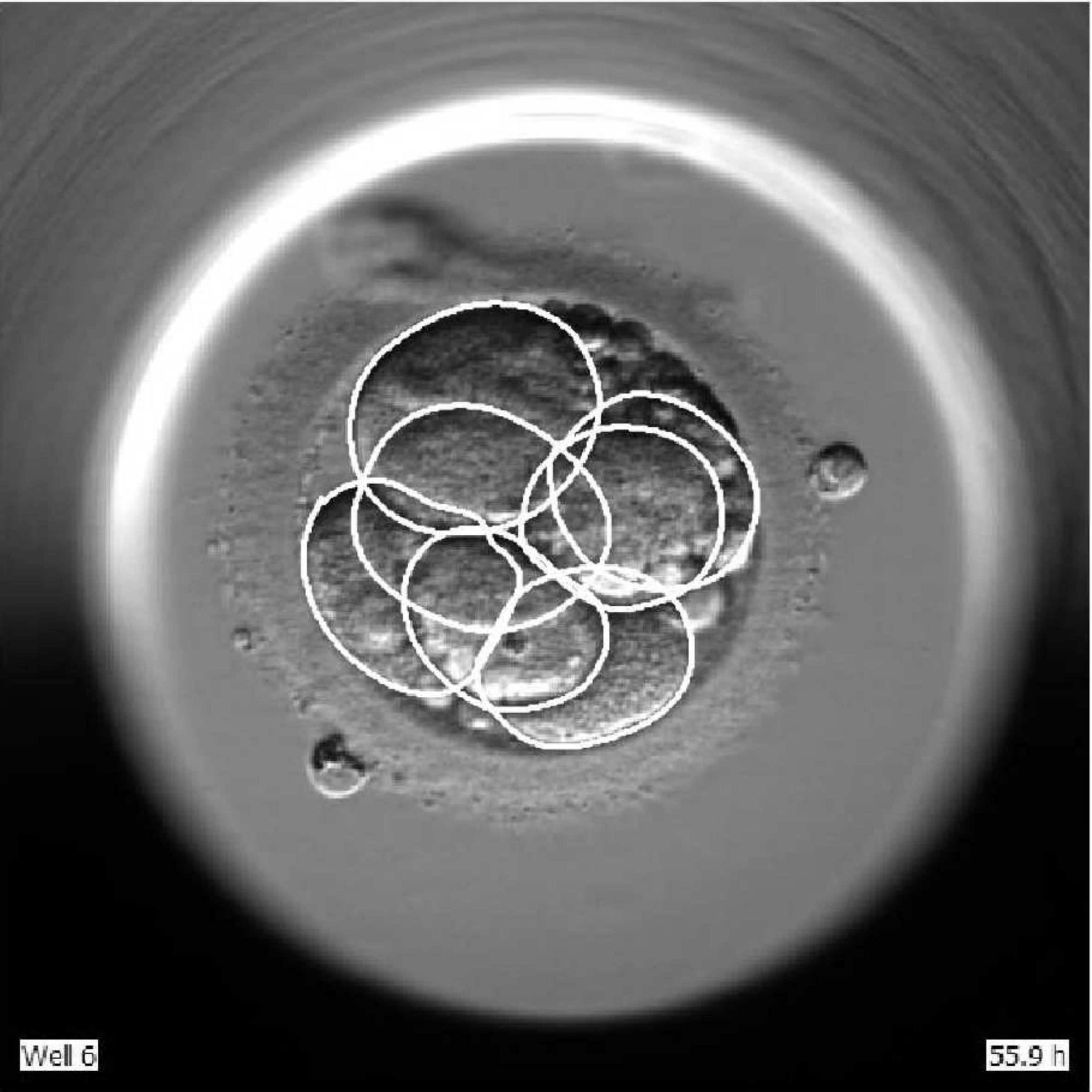} 
&  \includegraphics[trim=0cm 0cm 0cm 0cm, clip=true, width=.13\textwidth, height=20mm] {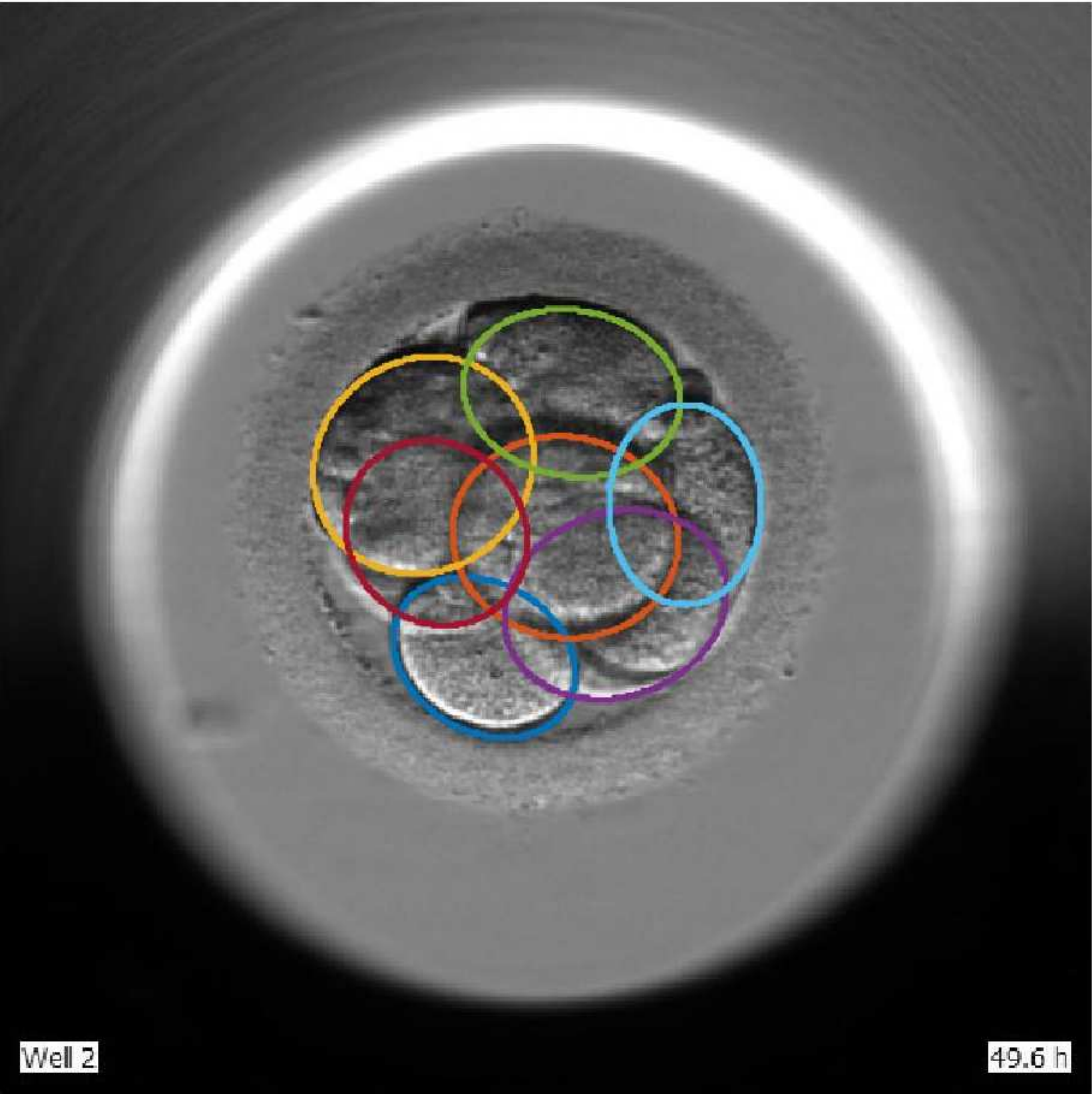}  & \includegraphics[trim=0cm 0cm 0cm 0cm, clip=true, width=.13\textwidth, height=20mm]{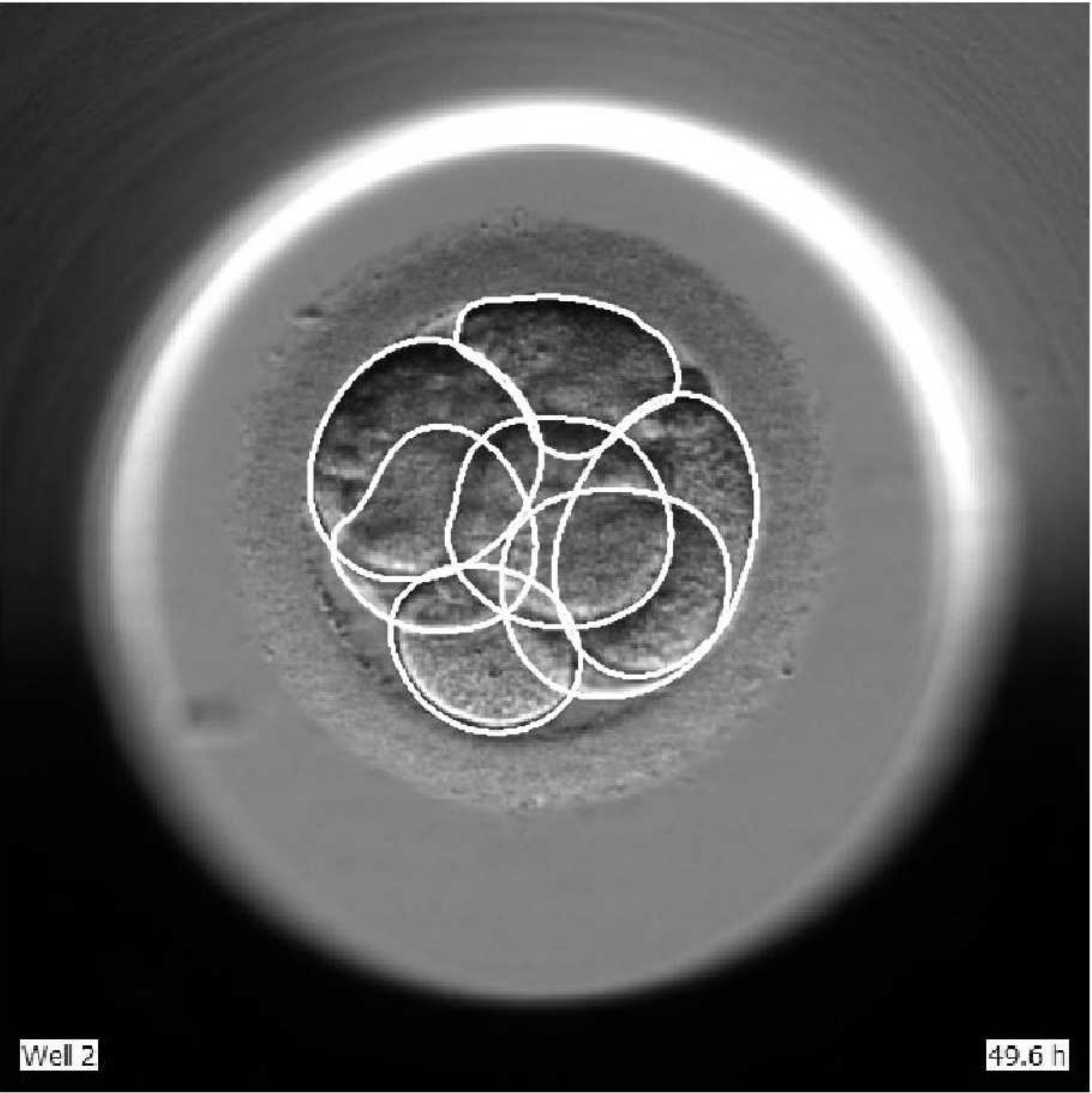}
&  \includegraphics[trim=0cm 0cm 0cm 0cm, clip=true, width=.13\textwidth, height=20mm]{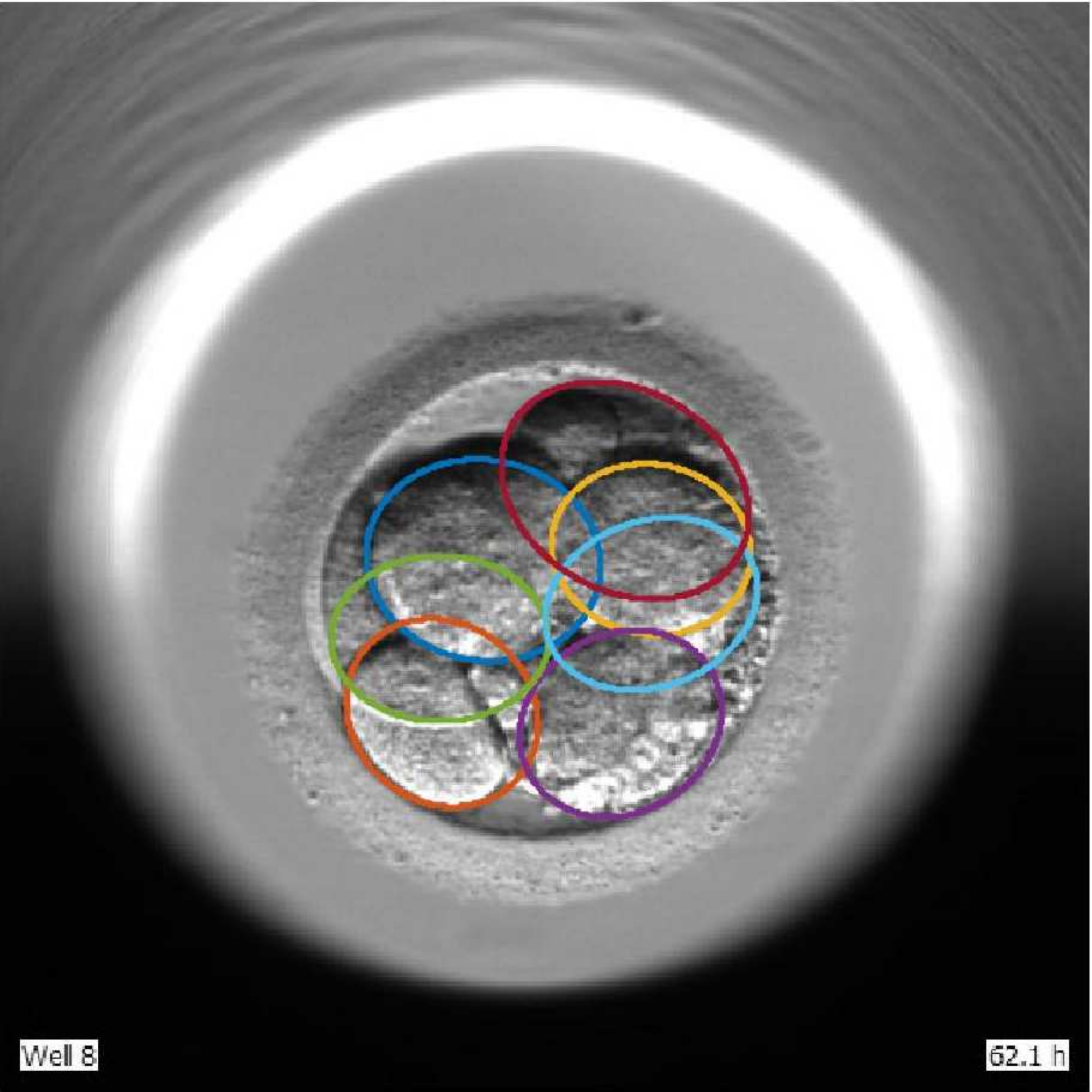} &  \includegraphics[trim=0cm 0cm 0cm 0cm, clip=true, width=.13\textwidth, height=20mm]{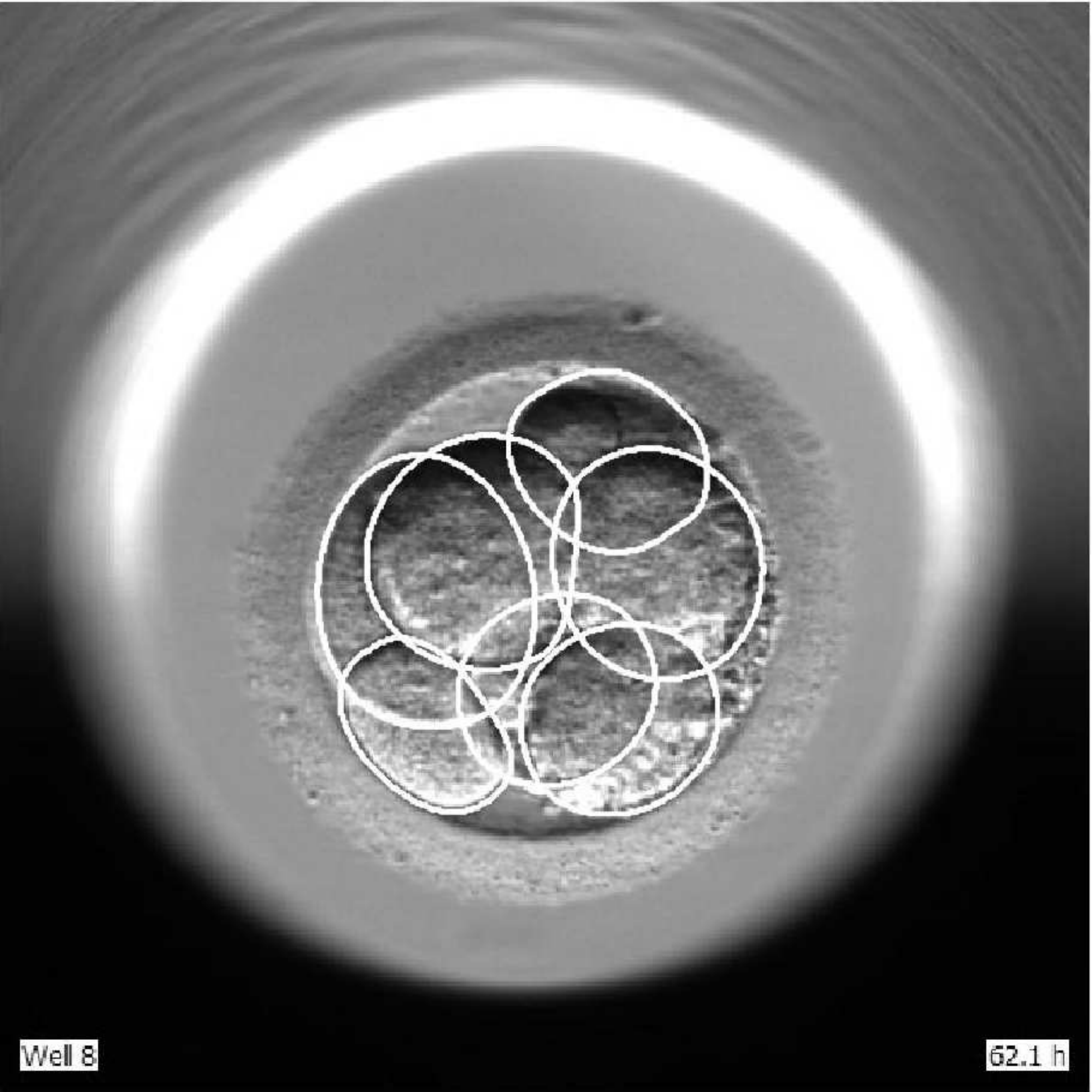} \\ \hline
{\rotatebox {90}{ 8-cell} } &  \includegraphics[trim=0cm 0cm 0cm 0cm, clip=true, width=.13\textwidth, height=20mm]{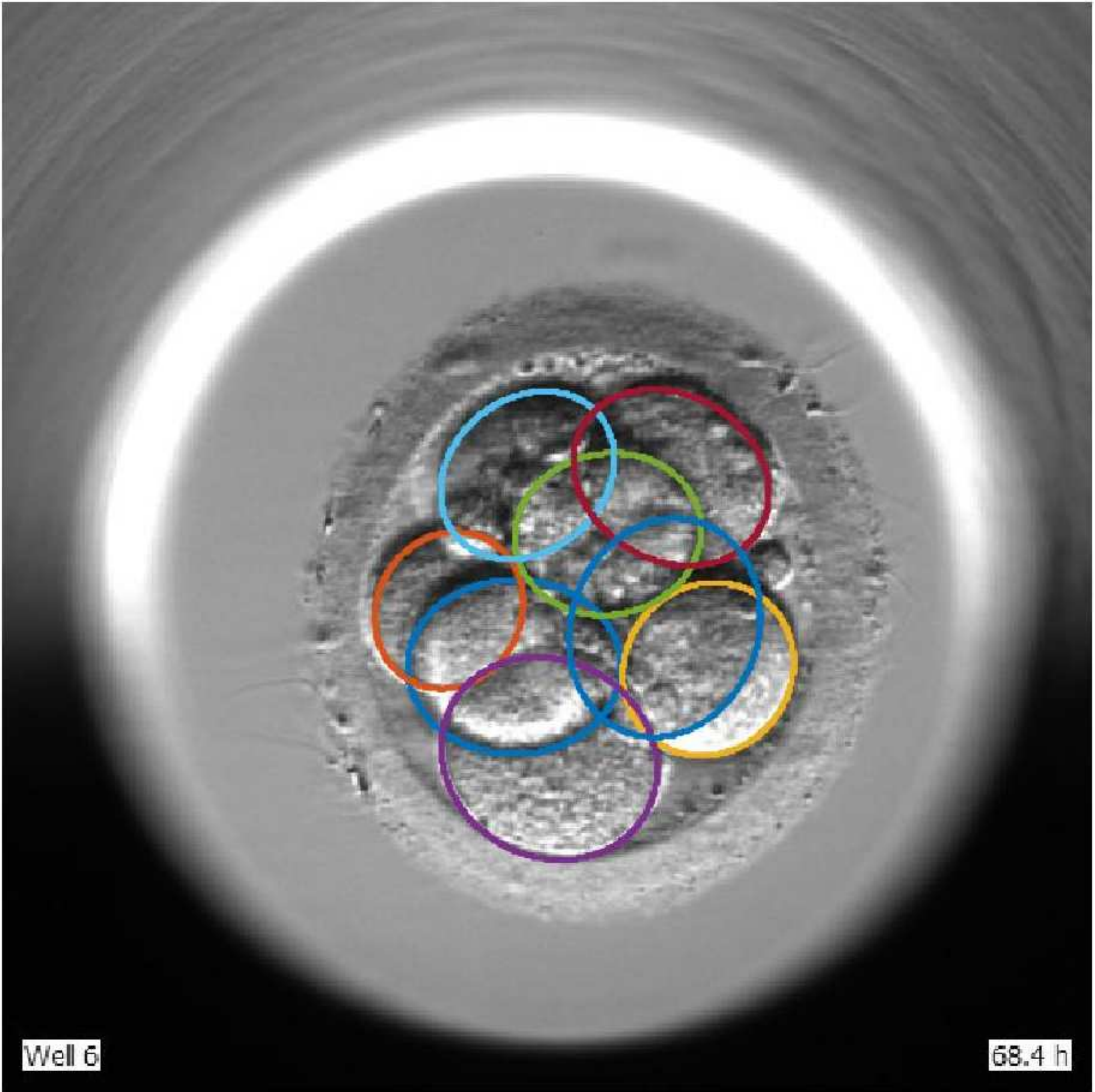} &  \includegraphics[trim=0cm 0cm 0cm 0cm, clip=true, width=.13\textwidth, height=20mm]{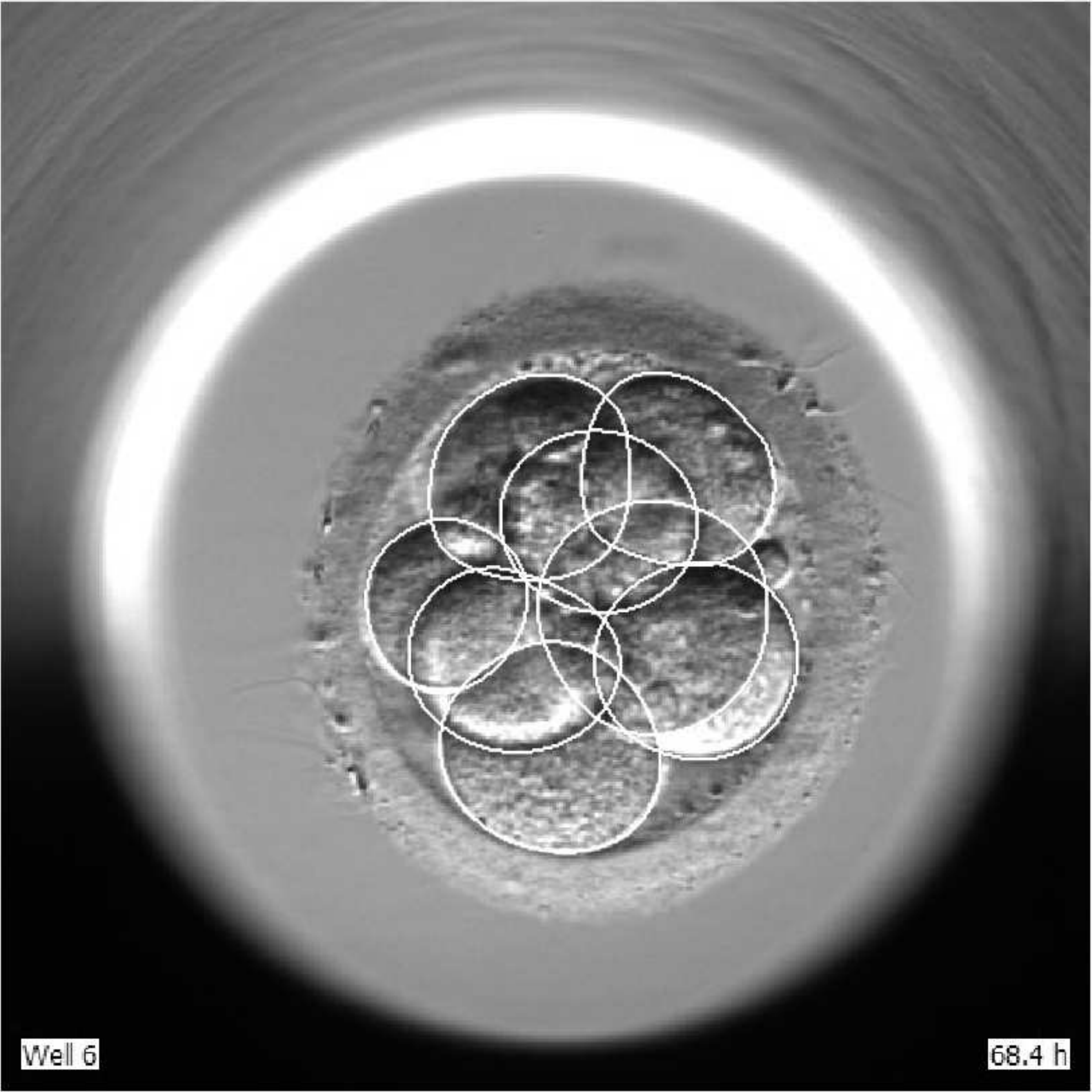} 
&  \includegraphics[trim=0cm 0cm 0cm 0cm, clip=true, width=.13\textwidth, height=20mm] {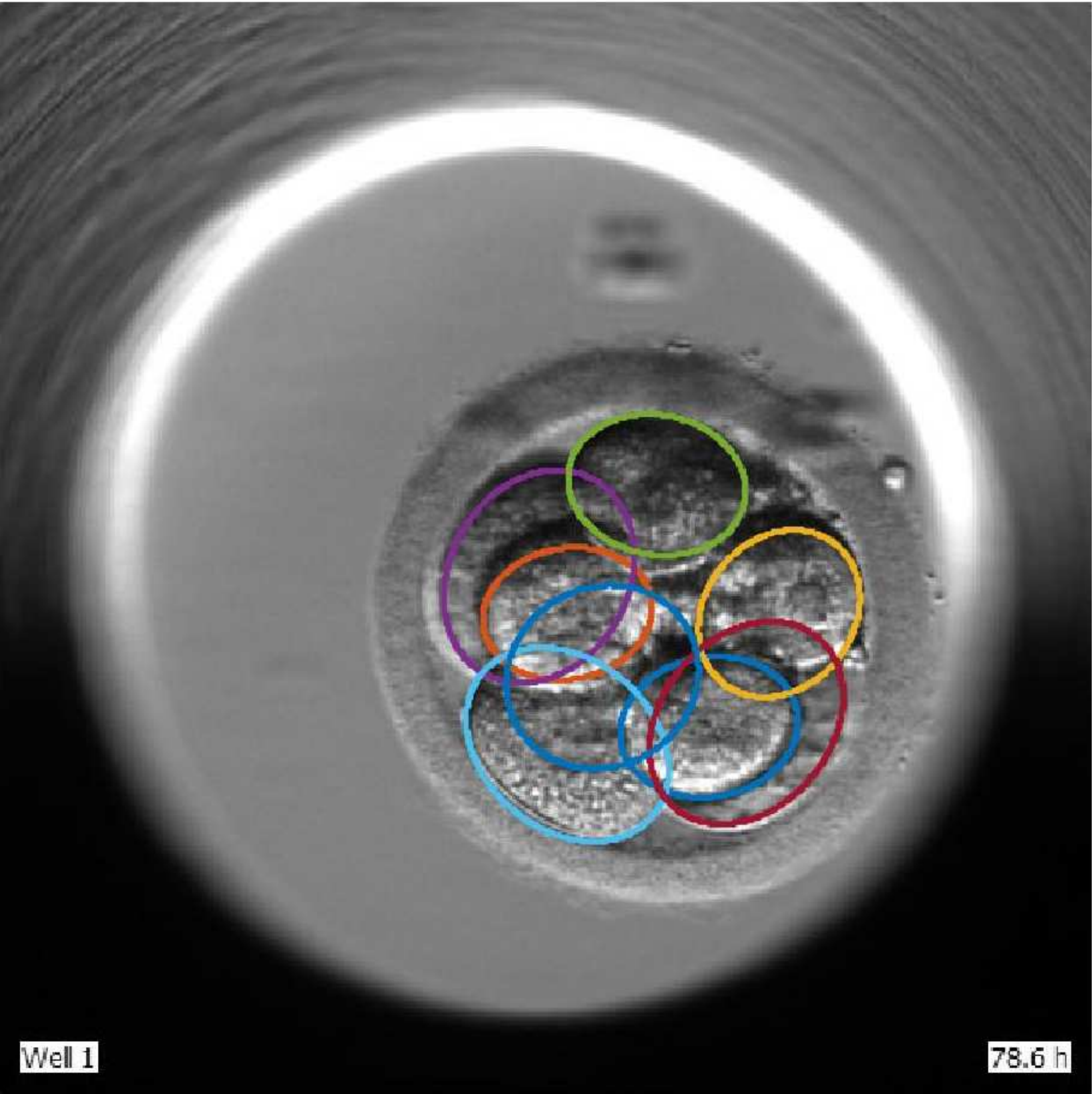}  & \includegraphics[trim=0cm 0cm 0cm 0cm, clip=true, width=.13\textwidth, height=20mm]{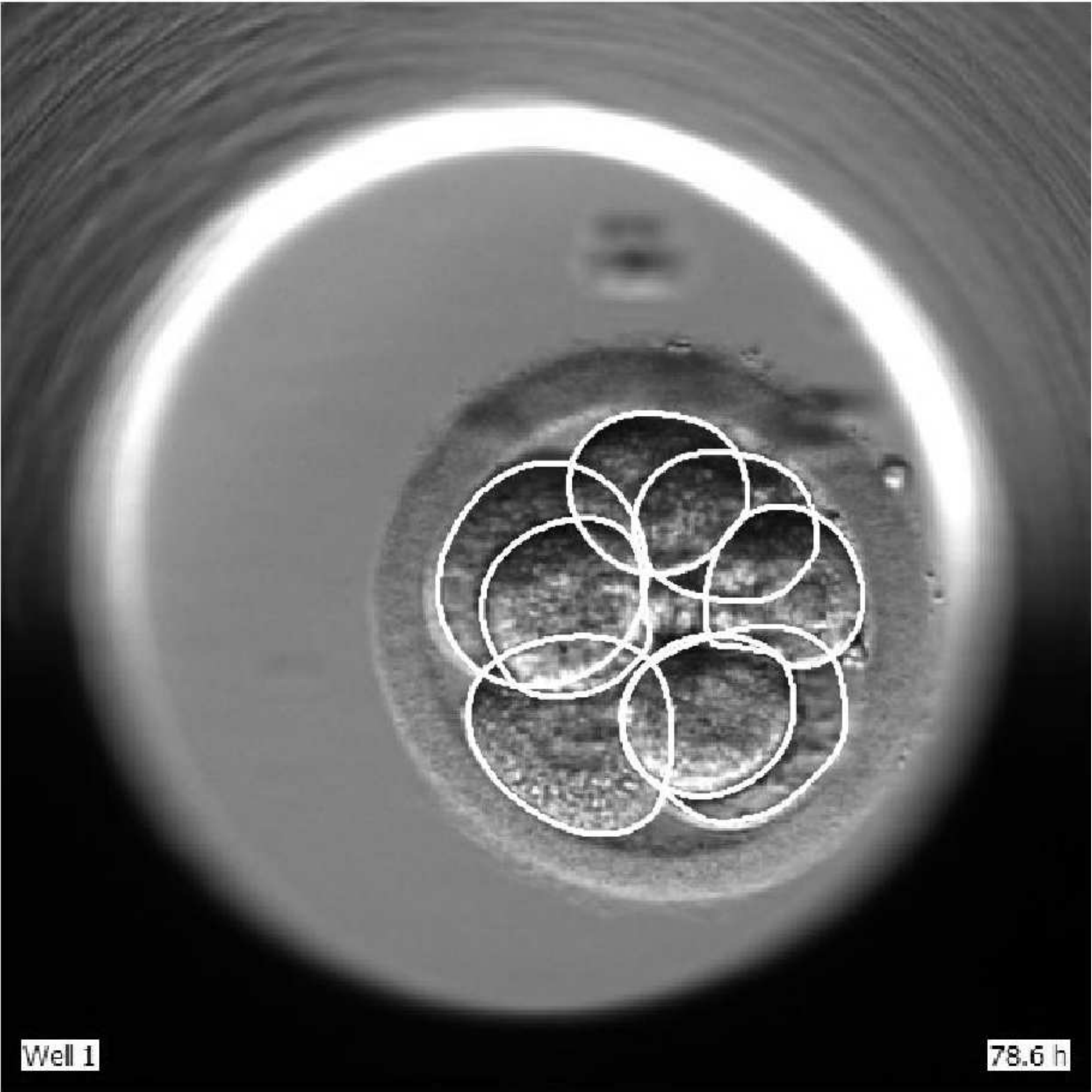}
&  \includegraphics[trim=0cm 0cm 0cm 0cm, clip=true, width=.13\textwidth, height=20mm]{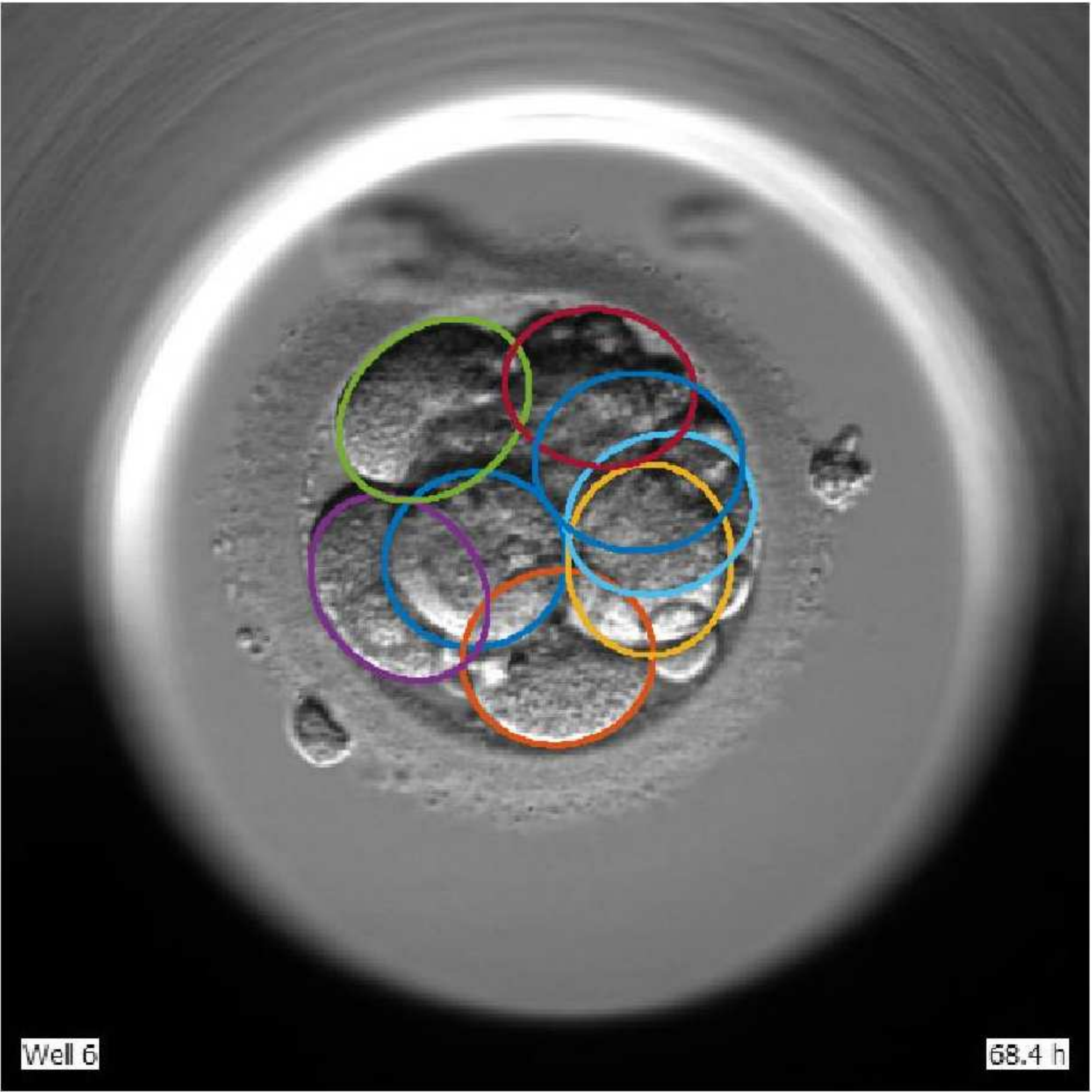} &  \includegraphics[trim=0cm 0cm 0cm 0cm, clip=true, width=.13\textwidth, height=20mm]{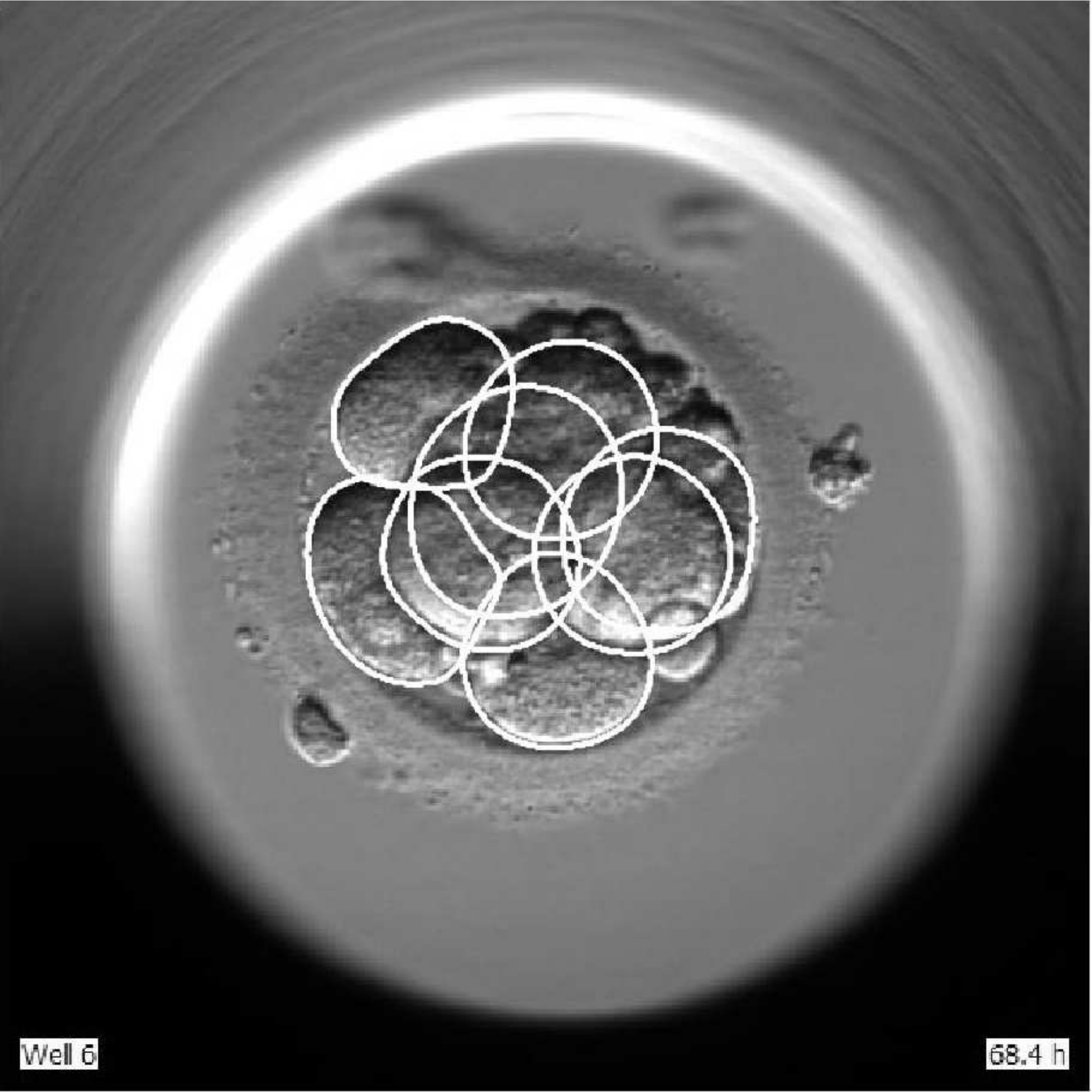} \\ \hline
\end{tabular}
\end{table*}

\section{Conclusion}
In this work, a new method for identifying blastomeres' boundaries in day 1 to day 3 of microscopic human embryo images is proposed. With the assumption of ellipsoidal models for the shape of blastomeres, local image properties (edges and normal vectors) are utilized to conform with ellipsoidal candidate models. In an iterative process, the candidates with the highest conformity with models are detected one by one. We have tested the proposed algorithm on a large dataset comprised of 468 embryo images, acquired from different sources. The results indicate overall \textit{Precision}, \textit{Sensitivity} and \textit{OQ} of 92\%, 88\% and 83\%, respectively. With cell size and symmetry as one of the important factors in a high-quality embryo, these results confirm that the proposed method can effectively identify the boundary of multiple blastomeres inside crowded embryos under occlusion. 
\bibliographystyle{elsarticle-num}
\bibliography{bd_sh}

\begin{thebibliography}{10}
\expandafter\ifx\csname url\endcsname\relax
  \def\url#1{\texttt{#1}}\fi
\expandafter\ifx\csname urlprefix\endcsname\relax\def\urlprefix{URL }\fi
\expandafter\ifx\csname href\endcsname\relax
  \def\href#1#2{#2} \def\path#1{#1}\fi

\bibitem{gardner2017handbook}
D.~K. Gardner, C.~Sim{\'o}n, Handbook of in vitro fertilization, CRC press,
  2017.

\bibitem{ElderKay}
K.~Elder, Human embryo preimplantation selection, Informa UK Ltd, 2007.

\bibitem{gardner2016assessment}
D.~K. Gardner, B.~Balaban, Assessment of human embryo development using
  morphological criteria in an era of time-lapse, algorithms and 'omics': is
  looking good still important?, MHR: Basic science of reproductive medicine
  22~(10) (2016) 704--718.

\bibitem{singh2014automatic}
A.~Singh, J.~Buonassisi, P.~Saeedi, J.~Havelock, Automatic blastomere detection
  in day 1 to day 2 human embryo images using partitioned graphs and
  ellipsoids, in: 2014 IEEE International Conference on Image Processing
  (ICIP), 2014, pp. 917--921.

\bibitem{IT-C:564}
U.~Pedersen, O.~Olsen, N.~Olsen, A multiphase variational level set approach
  for modelling human embryos, in: IEEE workshop on Variational Geometric and
  Level Set Methods in C. V., 2003, pp. 25--32.

\bibitem{Giusti2010}
A.~Giusti, G.~Corani, L.~Gambardella, C.~Magli, L.~Gianaroli, Blastomere
  segmentation and {3D} morphology measurements of early embryos from hoffman
  modulation contrast image stacks, in: IEEE int. conference on Biomedical
  imaging: from nano to Macro, 2010, pp. 1261--1264.

\bibitem{giusti2009lighting}
A.~Giusti, G.~Corani, L.~Gambardella, C.~Magli, L.~Gianaroli, Lighting-aware
  segmentation of microscopy images for in vitro fertilization, in:
  International Symposium on Visual Computing, Springer, 2009, pp. 576--585.

\bibitem{khan2016segmentation}
A.~Khan, S.~Gould, M.~Salzmann, Segmentation of developing human embryo in
  time-lapse microscopy, in: Biomedical Imaging (ISBI), 2016 IEEE 13th
  International Symposium on, IEEE, 2016, pp. 930--934.

\bibitem{Wong2010}
C.~Wong, et~al., Non-invasive imaging of human embryos before embryonic genome
  activation predicts development to the blastocyst stage, Nature
  Biotechnology~(28) (2010) 1115–--1121.

\bibitem{khan2015linear}
A.~Khan, S.~Gould, M.~Salzmann, A linear chain markov model for detection and
  localization of cells in early stage embryo development, in: 2015 IEEE Winter
  Conference on Applications of Computer Vision, 2015, pp. 526--533.

\bibitem{moussavi2014unified}
F.~Moussavi, Y.~Wang, P.~Lorenzen, J.~Oakley, D.~Russakoff, S.~Gould, A unified
  graphical models framework for automated mitosis detection in human embryos,
  IEEE transactions on medical imaging 33~(7) (2014) 1551--1562.

\bibitem{khan2015automated}
A.~Khan, S.~Gould, M.~Salzmann, Automated monitoring of human embryonic cells
  up to the 5-cell stage in time-lapse microscopy images, in: 2015 IEEE 12th
  International Symposium on Biomedical Imaging (ISBI), IEEE, 2015, pp.
  389--393.

\bibitem{grushnikov20183d}
A.~Grushnikov, R.~Niwayama, T.~Kanade, Y.~Yagi, 3d level set method for
  blastomere segmentation of preimplantation embryos in fluorescence microscopy
  images, Machine Vision and Applications 29~(1) (2018) 125--134.

\bibitem{irshad2014methods}
H.~Irshad, A.~Veillard, L.~Roux, D.~Racoceanu, Methods for nuclei detection,
  segmentation, and classification in digital histopathology: a review -
  current status and future potential, IEEE reviews in biomedical engineering 7
  (2014) 97--114.

\bibitem{xing2016robust}
F.~Xing, L.~Yang, Robust nucleus/cell detection and segmentation in digital
  pathology and microscopy images: a comprehensive review, IEEE reviews in
  biomedical engineering 9 (2016) 234--263.

\bibitem{rad2018hybrid}
R.~M. Rad, P.~Saeedi, J.~Au, J.~Havelock, A hybrid approach for multiple
  blastomeres identification in early human embryo images, Computers in biology
  and medicine 101 (2018) 100--111.

\bibitem{zou04}
K.~Zou, S.~Warfield, A.~Baharatha, C.~Tempany, M.~Kaus, S.~Haker, W.~Wells,
  F.~Jolesz, R.~Kikinis, {Statistical Validation of Image Segmentation Quality
  Based on a Spatial Overlap Index}, Academic Radiology 11 (2004) 178--189.

\bibitem{prasad2012edge}
D.~Prasad, M.~Leung, S.~Cho, Edge curvature and convexity based ellipse
  detection method, Pattern Recognition 45~(9) (2012) 3204--3221.

\bibitem{hahn2008new}
K.~Hahn, S.~Jung, Y.~Han, H.~Hahn, A new algorithm for ellipse detection by
  curve segments, Pattern Recognition Letters 29~(13) (2008) 1836--1841.

\bibitem{chia2011split}
A.~Chia, S.~Rahardja, D.~Rajan, M.~Leung, A split and merge based ellipse
  detector with self-correcting capability, IEEE Transactions on Image
  Processing 20~(7) (2011) 1991--2006.

\bibitem{Frangi98multiscalevessel}
A.~Frangi, W.~Niessen, K.~Vincken, M.~Viergever, Multiscale vessel enhancement
  filtering, IEEE Medical image computing and computer-assisted
  intervention--MICCAI (1998) 130--137.

\bibitem{Canny:1986}
J.~Canny, A computational approach to edge detection, IEEE Trans. Pattern Anal.
  Mach. Intell. 8 (1986) 679--698.

\bibitem{Dunham86}
J.~G. Dunham, Optimum uniform piecewise linear approximation of planar curves,
  IEEE Trans. Pattern Anal. Mach. Intell. 8~(1) (1986) 67--75.

\bibitem{DYee2013}
D.~Yee, P.~Saeedi, J.~Havelock, An automatic model-based approach for measuring
  the zona pellucida thickness in day five human blastocysts, in: Int. Conf.
  Image Process., Comput. Vis., Pattern Recog., 2013, pp. 877–--880.

\bibitem{UnisenseRef}
U.~Fertilitech, {Embryoscope},
  https://www.vitrolife.com/products/time-lapse-systems/ (2009).

\bibitem{fragment}
C.~Racowsky, M.~Vernon, J.~Mayer, G.~Ball, B.~Behr, K.~Pomeroy, D.~Wininger,
  W.~Gibbons, J.~Conaghan, J.~Stern, Standardization of grading embryo
  morphology, Fertility and sterility 94~(3) (2010) 1152--1153.

\bibitem{stone2005embryo}
B.~A. Stone, J.~Greene, J.~M. Vargyas, G.~E. Ringler, R.~P. Marrs, Embryo
  fragmentation as a determinant of blastocyst development in vitro and
  pregnancy outcomes following embryo transfer, American journal of obstetrics
  and gynecology 192~(6) (2005) 2014--2019.

\end{thebibliography}
\end{document}